\documentclass[11pt]{article}
\usepackage[letterpaper,top=0.92in,bottom=0.95in,left=1in,right=1in,headheight=15pt,headsep=18pt,footskip=30pt]{geometry}
\usepackage{iftex}
\usepackage{amsmath}
\ifPDFTeX
  \usepackage[T1]{fontenc}
  \usepackage[utf8]{inputenc}
\fi
\usepackage[scaled=0.98]{XCharter}
\usepackage[xcharter,vvarbb]{newtxmath}
\usepackage[scaled=0.95]{sourcesanspro}
\usepackage[scaled=0.86]{sourcecodepro}
\ifPDFTeX\else
  \AtBeginDocument{\DeclareSymbolFont{operators}{T1}{XCharter-TLF}{m}{n}%
    \SetSymbolFont{operators}{bold}{T1}{XCharter-TLF}{b}{n}%
    \DeclareMathAlphabet{\mathbf}{T1}{XCharter-TLF}{b}{n}%
    \DeclareMathAlphabet{\mathrm}{T1}{XCharter-TLF}{m}{n}}
\fi
\usepackage{microtype}
\usepackage[table]{xcolor}
\usepackage{graphicx,booktabs,array,multirow}
\usepackage{tikz}
\usepackage[most]{tcolorbox}
\usepackage{fontawesome5}
\usepackage{titlesec,titletoc,fancyhdr,extramarks,multicol}
\usepackage{enumitem,listings}
\usepackage[numbers,sort&compress]{natbib}
\usepackage{caption}
\usepackage{placeins,url}
\usepackage[hidelinks]{hyperref}
\definecolor{cdInk}{HTML}{172036}
\definecolor{cdTeal}{HTML}{0B7A75}
\definecolor{cdTealDark}{HTML}{0B4F55}
\definecolor{cdTealLight}{HTML}{E6F4F2}
\definecolor{cdAmber}{HTML}{E09A2C}
\definecolor{cdAmberLight}{HTML}{FCF2E0}
\definecolor{cdCoral}{HTML}{C8553D}
\definecolor{cdGray}{HTML}{6B7280}
\definecolor{cdLine}{HTML}{D5DBE3}
\definecolor{cdMist}{HTML}{F3F5F8}
\definecolor{szA}{HTML}{8CCBC3}\colorlet{szAtext}{cdInk}
\definecolor{szB}{HTML}{2A9D8F}\colorlet{szBtext}{white}
\definecolor{szC}{HTML}{0B4F55}\colorlet{szCtext}{white}
\definecolor{cdIndigo}{HTML}{4455B3}
\definecolor{cdIndigoDark}{HTML}{2A3578}
\definecolor{cdIndigoLight}{HTML}{E8EBF8}
\definecolor{idA}{HTML}{9AA6DE}\colorlet{idAtext}{cdInk}
\definecolor{idB}{HTML}{5B6FD0}\colorlet{idBtext}{white}
\definecolor{idC}{HTML}{2A3578}\colorlet{idCtext}{white}
\definecolor{hfYellow}{HTML}{FFD21E}
\definecolor{hfOrange}{HTML}{FF9D0B}
\hypersetup{colorlinks=true,linkcolor=cdTealDark,citecolor=cdTealDark,urlcolor=cdTeal,
  pdftitle={Chinese Decision: A Family of Language Models for Single-Pass Probabilistic Decision-Making},pdfauthor={Qiang Yan, Jiafeng Guo}}
\color{cdInk!92!black}
\newcommand{\chip}[2][cdTeal]{\tikz[baseline=(c.base)]\node[fill=#1,text=white,rounded corners=2.2pt,inner xsep=3pt,
  inner ysep=1.5pt,font=\sffamily\bfseries\footnotesize](c){#2};}
\newcommand{\minichips}{\tikz[baseline=-0.1ex]{\foreach \c [count=\i] in {szA,szB,szC}
  \fill[\c,rounded corners=0.6pt] ({(\i-1)*0.62em},0) rectangle ++(0.5em,0.5em);}}
\newcommand{\szdot}[1]{\tikz[baseline=-0.62ex]\fill[#1] (0,0) circle (0.34em);}
\newcommand{\szA}{\szdot{szA}}\newcommand{\szB}{\szdot{szB}}\newcommand{\szC}{\szdot{szC}}
\newcommand{\uniA}{\szdot{idA}}\newcommand{\uniB}{\szdot{idB}}\newcommand{\uniC}{\szdot{idC}}
\newcommand{\runA}{\szdot{cdCoral}}\newcommand{\runB}{\szdot{cdAmber}}
\newcommand{\best}[1]{\textbf{#1}}
\newcommand{\tablenote}[1]{{\footnotesize\color{cdGray}#1\par}}
\newcommand{\sigmark}{\textcolor{cdTeal}{\scriptsize\faCheckCircle}\,}
\newcommand{\nsmark}{\textcolor{cdGray}{\scriptsize\faCircle[regular]}\,}
\newcommand{\better}{\textcolor{cdTeal}{$\searrow$}}
\newcommand{\worse}{\textcolor{cdCoral}{$\nearrow$}}
\newcommand{\hficon}{\tikz[baseline=-0.78ex,x=1.05em,y=1.05em]{%
  \fill[hfYellow] (0,0) circle (0.5);
  \fill[cdInk] (-0.17,0.1) ellipse (0.055 and 0.085); \fill[cdInk] (0.17,0.1) ellipse (0.055 and 0.085);
  \fill[cdInk!80] (-0.21,-0.07) .. controls (-0.13,-0.31) and (0.13,-0.31) .. (0.21,-0.07) -- cycle;
  \fill[hfOrange] (-0.42,-0.24) ellipse (0.14 and 0.11); \fill[hfOrange] (0.42,-0.24) ellipse (0.14 and 0.11);}}
\newcommand{\linkitem}[3]{{\color{cdInk}#1}\hspace{0.35em}\textbf{#2}\hspace{0.3em}#3}
\newcommand{\linksep}{\hspace{0.9em}{\color{cdLine}\rule[-0.2ex]{0.6pt}{1.9ex}}\hspace{0.9em}}
\newcommand{\sectionchip}[2][cdTeal]{\tikz[baseline=(c.base)]\node[fill=#1,text=white,rounded corners=3pt,
  minimum width=1.6em,minimum height=1.45em,inner sep=2pt,font=\sffamily\bfseries\large](c){#2};}
\newcommand{\sectioncolor}{cdTeal}
\titleformat{\section}{\sffamily\bfseries\Large\color{cdInk}}{\sectionchip[\sectioncolor]{\thesection}}{0.55em}{}
\titlespacing*{\section}{0pt}{2.6ex plus .8ex minus .3ex}{1.2ex plus .2ex}
\titleformat{\subsection}{\sffamily\bfseries\large\color{cdInk}}{\textcolor{\sectioncolor}{\thesubsection}}{0.65em}{}
\titlespacing*{\subsection}{0pt}{2.0ex plus .6ex minus .3ex}{0.8ex plus .2ex}
\titleformat{\paragraph}[runin]{\sffamily\bfseries\color{cdTealDark}}{}{0pt}{}
\titlespacing*{\paragraph}{0pt}{1.3ex plus .4ex minus .2ex}{0.55em}
\titlecontents{section}[1.9em]{\sffamily\footnotesize\color{cdInk}\addvspace{2.2pt}}
  {\contentslabel[\textcolor{cdTeal}{\textbf{\thecontentslabel}}]{1.9em}}{\hspace*{-1.9em}}
  {\,{\color{cdLine}\titlerule*[0.5pc]{.}}\,\textcolor{cdGray}{\contentspage}}
\makeatletter
\newcommand{\covercontents}{\noindent{\sffamily\bfseries\footnotesize\color{cdTeal}CONTENTS}\par\vspace{2pt}
  {\color{cdLine}\hrule height 0.5pt}\vspace{3pt}
  \setlength{\columnsep}{18pt}\begin{multicols}{3}\@starttoc{toc}\end{multicols}}
\makeatother
\fancypagestyle{first}{\fancyhf{}\fancyfoot[C]{\sffamily\footnotesize\color{cdGray}\thepage}}
\DeclareCaptionLabelSeparator{bar}{\hspace{0.5em}{\color{cdLine}\rule[-0.25ex]{0.7pt}{2ex}}\hspace{0.5em}}
\setlist[itemize]{leftmargin=1.4em,itemsep=2pt,topsep=4pt,label={\textcolor{cdTeal}{\scriptsize\faCaretRight}}}
\setlist[enumerate]{leftmargin=1.6em,itemsep=2pt,topsep=4pt}
\newtcolorbox{finding}[1][Finding]{enhanced,breakable,colback=cdTealLight,colframe=cdTealLight,boxrule=0pt,arc=3pt,
  left=10pt,right=10pt,top=6pt,bottom=6pt,borderline west={3pt}{0pt}{cdTeal},fontupper=\small,before skip=8pt,after skip=10pt,
  before upper={\textcolor{cdTealDark}{\sffamily\bfseries\faLightbulb[regular]\hspace{0.45em}#1.}\hspace{0.35em}}}
\newtcolorbox{ufinding}[1][Finding]{enhanced,breakable,colback=cdIndigoLight,colframe=cdIndigoLight,boxrule=0pt,arc=3pt,
  left=10pt,right=10pt,top=6pt,bottom=6pt,borderline west={3pt}{0pt}{cdIndigo},fontupper=\small,before skip=8pt,after skip=10pt,
  before upper={\textcolor{cdIndigoDark}{\sffamily\bfseries\faLayerGroup\hspace{0.45em}#1.}\hspace{0.35em}}}
\newtcolorbox{caveat}[1][Caveat]{enhanced,breakable,colback=cdAmberLight,colframe=cdAmberLight,boxrule=0pt,arc=3pt,
  left=10pt,right=10pt,top=6pt,bottom=6pt,borderline west={3pt}{0pt}{cdAmber},fontupper=\small,before skip=8pt,after skip=10pt,
  before upper={\textcolor{cdAmber!80!black}{\sffamily\bfseries\faExclamationTriangle\hspace{0.45em}#1.}\hspace{0.35em}}}
\newtcolorbox{keyeq}{enhanced,colback=cdMist,colframe=cdMist,boxrule=0pt,arc=3pt,left=6pt,right=6pt,top=0pt,bottom=2pt,
  before skip=6pt,after skip=8pt}
\newtcolorbox{glance}{enhanced,breakable,colback=white,colframe=cdLine,boxrule=0.6pt,arc=4pt,left=11pt,right=11pt,top=7pt,bottom=7pt,
  title={\faCompass[regular]\hspace{0.45em}At a glance},fonttitle=\sffamily\bfseries\small,coltitle=cdTealDark,colbacktitle=cdTealLight,
  titlerule=0pt,toptitle=3pt,bottomtitle=3pt,fontupper=\small,before skip=6pt,after skip=12pt}
\newcommand{\kpi}[4]{\begin{tcolorbox}[enhanced,colback=white,colframe=cdLine,boxrule=0.6pt,arc=4pt,left=7pt,right=6pt,top=6pt,bottom=6pt,
  borderline north={2.6pt}{0pt}{cdTeal},equal height group=kpi,halign=flush left]
  {\color{cdTeal}\rule{0pt}{1.05em}\smash{#1}}\par\vspace{3pt}
  {\sffamily\bfseries\fontsize{19}{21}\selectfont\color{cdInk}#2}\par\vspace{2pt}
  {\sffamily\footnotesize\color{cdInk}#3}\par
  {\sffamily\scriptsize\color{cdGray}#4}\end{tcolorbox}}
\newcommand{\familycard}[5]{\begin{tcolorbox}[enhanced,colback=white,colframe=cdLine,boxrule=0.6pt,arc=4pt,left=8pt,right=8pt,
  top=7pt,bottom=7pt,borderline north={3pt}{0pt}{#1},fontupper=\sffamily\footnotesize]
  \tikz[baseline=(c.base)]\node[fill=#1,text=#1text,rounded corners=2.5pt,inner xsep=4pt,inner ysep=2.5pt,font=\sffamily\bfseries\large](c){#2};%
  \hspace{0.45em}{\sffamily\bfseries\tiny\color{cdGray}#4}\hfill{\ttfamily\scriptsize\color{cdGray}#3}\par\vspace{7pt}#5\end{tcolorbox}}

\newcommand{\speckey}[2]{\noindent\makebox[5.6em][l]{\color{cdGray}#1}{\color{cdInk}#2}\par}
\newcommand{\statcell}[2]{\begin{tabular}[t]{@{}l@{}}{\bfseries\large\color{cdInk}#1}\\[-1pt]{\scriptsize\color{cdGray}#2}\end{tabular}}
\newcommand{\specstats}[6]{\vspace{6pt}{\color{cdLine}\hrule height 0.5pt}\vspace{5pt}\noindent\statcell{#1}{#2}\hfill\statcell{#3}{#4}\hfill\statcell{#5}{#6}\par}
\newcommand{\specspeed}[4]{\vspace{5pt}\noindent{\color{cdTeal}\faBolt}\hspace{0.3em}#1\hspace{0.25em}{\color{cdGray}#2}\hfill
  {\color{cdTeal}\faMicrochip}\hspace{0.3em}#3\hspace{0.25em}{\color{cdGray}#4}\par}
\newcommand{\model}{\textsc{Chinese Decision}}
\newcommand{\noul}{\texttt{noul}}
\newcommand{\choice}{\texttt{choice}}
\newcommand{\score}{\texttt{score}}
\newcommand{\CE}{\mathcal{L}_{\mathrm{CE}}}
\newcommand{\R}{\mathbb{R}}
\newcommand{\ind}{\mathbf{1}}

\begin{document}
\thispagestyle{first}
% ---------------------------------------------------------------- title block
\begingroup
\setlength{\parindent}{0pt}
{\sffamily\small\color{cdTeal}\minichips\hspace{0.6em}\textbf{TECHNICAL REPORT}\hspace{0.5em}\textperiodcentered\hspace{0.5em}VERSION 6\hspace{0.5em}\textperiodcentered\hspace{0.5em}NOVEMBER 2026\par}
\vspace{12pt}
{\sffamily\bfseries\fontsize{31}{35}\selectfont\color{cdInk}Chinese Decision\par}
\vspace{5pt}
{\sffamily\fontsize{15}{19}\selectfont\color{cdTealDark}A Family of Language Models for Single-Pass\\Probabilistic Decision-Making\par}
\vspace{11pt}
\tikz{\foreach \c/\w [remember=\x as \lx (initially 0)] in {szA/0.14,szB/0.22,szC/0.30,idB/0.22,cdAmber/0.12}{
  \pgfmathsetmacro{\x}{\lx+\w}
  \fill[\c] (\lx\linewidth+0.6pt,0) rectangle (\x\linewidth-0.6pt,3.2pt);}}
\par\vspace{11pt}
{\sffamily\large\bfseries\color{cdInk}Qiang Yan\hspace{0.6em}\textperiodcentered\hspace{0.6em}Jiafeng Guo\par}
\vspace{2pt}
{\sffamily\small\color{cdGray}Institute of Computing Technology, Chinese Academy of Sciences\hspace{0.5em}\textperiodcentered\hspace{0.5em}\href{mailto:yanqiang@ict.ac.cn}{\texttt{yanqiang@ict.ac.cn}}, \href{mailto:guojiafeng@ict.ac.cn}{\texttt{guojiafeng@ict.ac.cn}}\par}
\vspace{8pt}
{\sffamily\small
\linkitem{\hficon}{Models:}{\href{https://huggingface.co/quincyqiang}{Hugging Face}}\linksep
\linkitem{\faCubes}{Mirror:}{\href{https://modelscope.cn/models/quincyqiang/chinese-decision-unified-2b}{ModelScope}}\linksep
\linkitem{\faGithub}{Code:}{\href{https://github.com/yanqiangmiffy/chinese-decision}{GitHub}}\linksep
\linkitem{\faDesktop}{Studio:}{\href{https://github.com/yanqiangmiffy/chinese-decision-studio}{Decision Studio}}\par}
\vspace{9pt}
\endgroup
% ---------------------------------------------------------------- abstract
\begin{tcolorbox}[enhanced,colback=cdMist,colframe=cdMist,boxrule=0pt,arc=4pt,left=13pt,right=13pt,top=10pt,bottom=10pt]
{\sffamily\bfseries\footnotesize\color{cdTeal}ABSTRACT}\par\vspace{4pt}
\small
Many applications need either a probability distribution over a few decisions or a set of typed spans from a text, and both are usually obtained by generating text. We present \model{}, a family of Chinese models that answers both with a single forward pass and no generated token: candidate-letter logits give probabilities for Boolean, categorical, and ordinal decisions, and a small span head reading an echoed copy of the input gives typed character spans. Both readouts share one backbone and one merged LoRA, so a unified model is one weight file. We release decision-only and unified models at 0.8B, 2B, and 4B (Qwen3.5), trained for one epoch on 13.05 million decisions and, for the unified models, on CJ-IE, a 1.2-million-row Chinese extraction mixture. On the full 307,900-decision CJ-Bench, the unified 4B model reaches 86.86\% accuracy, above every reported baseline and the decision-only 4B model (85.97\%), and extracts spans on CJ-IE with micro F1 of 68.9. Retention has a price: against a 0.8B model continued on CJ-IE alone, which reaches 73.9 F1 but forgets CJ-Bench (71.02\%), keeping decision ability costs 5.0 to 10.7 F1 points, shrinking with size. Gains of the unified models over decision-only ones are small, and all comparisons rest on single training runs with paired clustered bootstrap intervals.
\end{tcolorbox}
\vspace{4pt}
% ---------------------------------------------------------------- key numbers
\begin{tcbraster}[raster columns=4,raster equal height=rows,raster column skip=7pt]
\kpi{\faBullseye}{86.86\%}{CJ-Bench accuracy}{unified 4B model}
\kpi{\faHighlighter}{68.9 F1}{typed spans, unified 4B}{CJ-IE test, 35 sources}
\kpi{\faObjectGroup}{6 models}{3 decision-only, 3 unified}{0.8B, 2B, and 4B each}
\kpi{\faBolt}{0 tokens}{generated per request}{decisions and spans alike}
\end{tcbraster}
\vfill
\covercontents
\clearpage

\section{Introduction}
\label{sec:intro}
\begin{figure}[t]
\centering
\includegraphics[width=\textwidth]{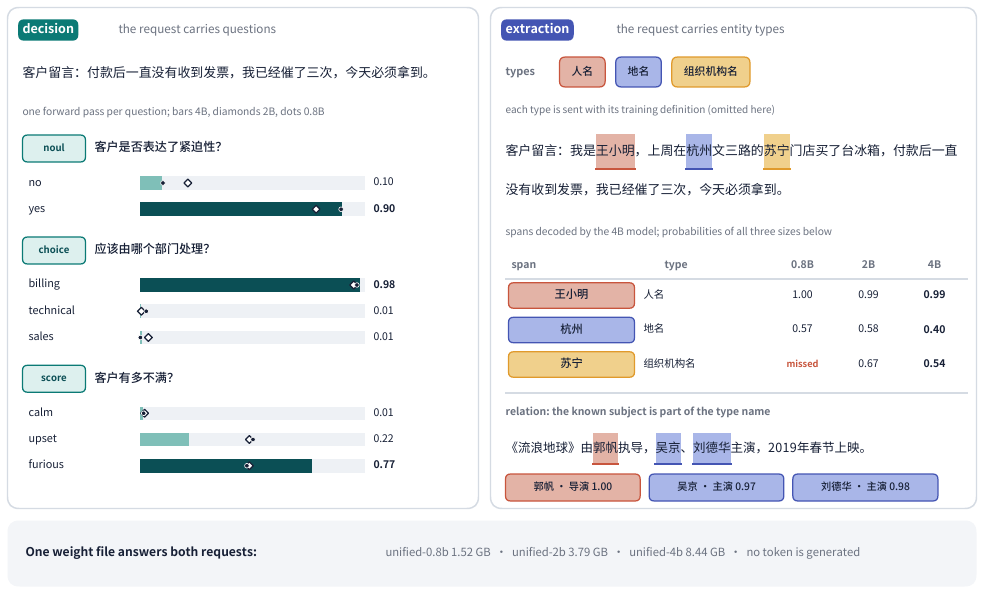}
\caption{\textbf{One weight file, two request shapes.} The released unified models (BF16, GPU) answer a customer message in two ways. Left: three questions about the message; bars are the 4B probabilities, diamonds the 2B ones, and dots the 0.8B ones. Right: a request that carries entity types instead of questions returns character spans with probabilities, and a relation request puts the known subject into the type name. The 0.8B model misses the organization that the 2B and 4B models find, with probabilities 0.67 and 0.54. Every number is a real model output; the types are sent with their training definitions, and no token is generated.}
\label{fig:teaser}
\end{figure}

A customer-support system may need to select a department, decide whether a message is urgent, and also find the customer, the store, and the street named in it. The first two are bounded decisions whose answer should be a probability. The third is typed span extraction. Both are usually obtained by generating text and parsing it, and an application that depends on uncertainty then has to recover probabilities that the generation interface does not expose.

\model{} exposes both problems directly. For a decision, it accepts a context, a question, and candidate descriptions, and returns a normalized probability distribution that supports a Boolean judgment, a categorical selection, or the expected value of an ordered rating. For extraction, it accepts a text and a set of entity types, each with a definition, and returns typed character spans with probabilities. Each request is processed once. A decision reads existing vocabulary logits at the final input position. An extraction reads a small head over the hidden states of an echoed copy of the text (Figure~\ref{fig:teaser}). No answer token is sampled, no free-form answer is parsed, and no chain of thought is generated. Here, \emph{single-pass} refers to one forward evaluation per compiled request; it does not imply constant work independent of input length or that all requests in an API call share one forward evaluation.

The project builds on the option-letter decision workflow in \texttt{hmm}~\citep{hmm2026}, Qwen3.5 language backbones~\citep{qwen08,qwen2b,qwen4b}, and the CJ-Dataset and CJ-Bench resources~\citep{wang2026chinesejev}. The span head follows the typed span decisions of CalOn~\citep{wang2026calon} on a causal backbone. Our focus is adaptation, joint-domain and joint-task training, empirical characterization, and deployment of a reusable Chinese model. We do not introduce the candidate-letter readout, the span-matching idea, the underlying language architecture, or the upstream benchmarks as new methods.

The family now has six released members (Table~\ref{tab:family}). The three decision-only models (0.8B, 2B, 4B) each adapt a Qwen3.5 text backbone with low-rank adaptation~\citep{hu2021lora} for one epoch over the same 13,048,000 decisions in four domains. The three unified models (0.8B, 2B, 4B) are trained from the raw backbone on the same decisions plus CJ-IE, and each is a single safetensors file whose merged backbone also carries a 3.4M-, 6.6M-, or 8.1M-parameter span head. All six are released with a validation-fitted temperature, and the decision-only models also with Q8\_0 GGUF conversions.

\begin{table}[t]
\caption{\textbf{The released family.} Top: decision-only models (CJ-Bench accuracy and ECE of the BF16 release at its fitted temperature; latency is the batch-one median over 25 inputs on a shared server). Bottom: unified models, which add the span readout (CJ-Bench accuracy; micro F1 on the CJ-IE test split; accuracy on the extraction-related decisions of that split).}
\label{tab:family}
\begin{tcbraster}[raster columns=3, raster equal height, raster column skip=8pt]
\familycard{szA}{0.8B}{chinese-decision-0.8b}{\textsc{decision}}{%
\speckey{Backbone}{Qwen3.5-0.8B}
\speckey{Objective}{CE}
\speckey{BF16 / Q8\_0}{1.50 / 0.81 GB}
\speckey{Fitted $T$}{1.108}
\specstats{83.11}{overall \%}{76.90}{prof.\ avg \%}{0.27}{ECE \%}
\specspeed{44\,ms}{GPU}{0.43\,s}{CPU}}
\familycard{szB}{2B}{chinese-decision-2b}{\textsc{decision}}{%
\speckey{Backbone}{Qwen3.5-2B}
\speckey{Objective}{CE + RLCD}
\speckey{BF16 / Q8\_0}{3.76 / 2.01 GB}
\speckey{Fitted $T$}{1.075}
\specstats{84.11}{overall \%}{78.25}{prof.\ avg \%}{0.64}{ECE \%}
\specspeed{45\,ms}{GPU}{0.82\,s}{CPU}}
\familycard{szC}{4B}{chinese-decision-4b}{\textsc{decision}}{%
\speckey{Backbone}{Qwen3.5-4B}
\speckey{Objective}{CE + RLCD}
\speckey{BF16 / Q8\_0}{8.41 / 4.48 GB}
\speckey{Fitted $T$}{1.025}
\specstats{85.97}{overall \%}{84.00}{prof.\ avg \%}{1.00}{ECE \%}
\specspeed{60\,ms}{GPU}{2.00\,s}{CPU}}
\end{tcbraster}
\par\vspace{7pt}
\begin{tcbraster}[raster columns=3, raster equal height, raster column skip=8pt]
\familycard{idA}{0.8B}{chinese-decision-unified-0.8b}{\textsc{unified}}{%
\speckey{Backbone}{Qwen3.5-0.8B}
\speckey{Objective}{CE + RLCD, both readouts}
\speckey{Weights}{1.52 GB, one file}
\speckey{Span head}{3.4M parameters}
\speckey{Fitted $T$}{1.133}
\specstats{83.46}{CJ-Bench \%}{63.1}{span F1}{90.6}{IE decisions \%}}
\familycard{idB}{2B}{chinese-decision-unified-2b}{\textsc{unified}}{%
\speckey{Backbone}{Qwen3.5-2B}
\speckey{Objective}{CE + RLCD, both readouts}
\speckey{Weights}{3.79 GB, one file}
\speckey{Span head}{6.6M parameters}
\speckey{Fitted $T$}{1.062}
\specstats{85.00}{CJ-Bench \%}{66.0}{span F1}{91.3}{IE decisions \%}}
\familycard{idC}{4B}{chinese-decision-unified-4b}{\textsc{unified}}{%
\speckey{Backbone}{Qwen3.5-4B}
\speckey{Objective}{CE + RLCD, both readouts}
\speckey{Weights}{8.44 GB, one file}
\speckey{Span head}{8.1M parameters}
\speckey{Fitted $T$}{1.099}
\specstats{86.86}{CJ-Bench \%}{68.9}{span F1}{91.9}{IE decisions \%}}
\end{tcbraster}

\end{table}

\begin{glance}
\begin{enumerate}[label={\chip{\arabic*}},leftmargin=2.1em,itemsep=3pt,topsep=0pt,partopsep=0pt]
\item \textbf{One checkpoint, four domains.} The 4B decision-only model reaches 75.69\% general and 84.00\% professional-average accuracy, and the unified 4B model 77.30\% and 86.03\%, above every reported baseline in each domain (Sections~\ref{sec:main} and~\ref{sec:unified-decision}).
\item \textbf{Scale helps most where data are scarce.} From 2B to 4B, law gains 8.5 points (decision-only) or 9.0 (unified) and finance 7.1 or 6.4, with clustered intervals excluding zero; the 2B gain over 0.8B in law is not resolved for decision-only models (Section~\ref{sec:uncertainty}).
\item \textbf{One weight file keeps both abilities.} The unified 0.8B, 2B, and 4B models reach 83.46\%, 85.00\%, and 86.86\% on CJ-Bench and 63.1, 66.0, and 68.9 extraction F1 (64.7, 67.3, and 70.6 with the release's decision threshold). Their decision gains over decision-only models are small: $+0.35$, $+0.89$, and $+0.87$ points, and 0.8B gains nothing once one source is set aside (Sections~\ref{sec:unified-decision} and~\ref{sec:extraction}).
\item \textbf{Retention has a price.} Continuing on CJ-IE alone gives 73.9 F1 and forgets CJ-Bench (71.02\%); keeping decisions costs 10.7, 7.9, and 5.0 F1 at 0.8B, 2B, and 4B against that reference, mostly as spans never proposed, and decoding recovers less than two points (Sections~\ref{sec:retention-cost} and~\ref{sec:decoding}).
\item \textbf{Aggregates mislead.} A global temperature lowers pooled ECE but raises it in some domains, and merging or quantizing flips 0.4--1.1\% of top-1 decisions at unchanged accuracy (Sections~\ref{sec:calibration} and~\ref{sec:quant}).
\end{enumerate}
\end{glance}

The first versions of this report covered decision-only models; version 4 added the unified 0.8B and 2B models, the CJ-IE data they use, and an analysis of what joint training costs; version 5 added the unified 4B model, the decision threshold that the release now uses for spans, and requests that are answered by both readouts in one pass. This version corrects which runs use RLCD and changes no result. Our contributions are:
\begin{itemize}
\item A single-pass interface with two readouts on one causal backbone, and a unified training and export recipe that yields one weight file; the span readout reads an echoed, character-aligned copy of the text.
\item CJ-IE, a leakage-controlled mixture of 573,294 span rows and 611,354 extraction-related decisions assembled from public Chinese NER, relation, and event corpora, with standardized type definitions.
\item A full CJ-Bench study of six released models and all 14 evaluated decision-only checkpoints, and a full CJ-IE evaluation of five joint-training runs, with imported baselines, local measurements, and confounders kept apart.
\item Analyses computed from stored per-decision and per-span predictions: parent-clustered paired bootstrap intervals, a temperature-only calibration ablation, selective prediction, source-level tests, a decoding sweep with a dev-selected operating point and the release default, and an error taxonomy.
\item An open deployment implementation with Transformers and GGUF backends for decisions and a Transformers server for both readouts, including one-pass combined requests, with one latency protocol across sizes and backends.
\end{itemize}
Code, weights for all released files, and the application workbench are linked on the first page. Appendix~\ref{app:changes} lists the changes between versions.

\section{Related Work}
\label{sec:related}
\paragraph{Bounded decisions with language models.}
Candidate descriptions allow one model to express multiple classification tasks without a separate application-specific output vocabulary. The \texttt{hmm} implementation~\citep{hmm2026} uses answer-letter logits and LoRA for structured decision prediction. We adapt that design to Chinese prompts and distribution-valued supervision. The readout is a constrained classification distribution derived from a language-model head, rather than a distribution over generated answer strings. StartLux-Decision~\citep{zhuang2026startlux} is concurrent work on decision models from 0.8B to 27B parameters that adds confidence-guided reasoning for uncertain decisions; we do not add reasoning.

\paragraph{Chinese decision resources.}
Chinese-Jev~\citep{wang2026chinesejev} provides the data and benchmark used here, and studies an encoder-based decision system with general training followed by specialist adaptation. Our implementation instead jointly adapts a causal language backbone across domains. Differences in backbone, objective, and training allocation prevent attributing performance gaps to a single architectural choice.

\paragraph{Typed span extraction.}
GLiNER~\citep{zaratiana2024gliner} matches spans with entity-type embeddings from a bidirectional encoder to handle open-type NER, and GLiNER2~\citep{zaratiana2025gliner2} serves several information-extraction tasks, including classification, from one model. CalOn~\citep{wang2026calon} frames entity, relation, and event extraction as typed decisions conditioned on type definitions, with a none class and confidence calibration. Our span head uses the same ingredients on a causal backbone: span and type vectors, a scaled dot product, and a learned none key. We do not copy that implementation, and we do not claim zero-shot generalization to new types, which none of our experiments measure systematically. Type names and definitions come from the standardized taxonomy of B2NERD~\citep{yang2025b2nerd}.

\paragraph{Reading from repeated text.}
A causal model cannot see text to the right of a token. Repeating the input and reading representations from the second copy lets each position see the whole text~\citep{springer2025repetition}. Our echo is character-aligned: the text appears once with natural tokenization and once with one unit per character, so every character boundary is a token boundary, and spans are read from the second copy.

\paragraph{Adaptation, calibration, and retention.}
LoRA~\citep{hu2021lora} restricts parameter updates to low-rank matrices that can be merged into the backbone, and it forgets less than full fine-tuning at the price of learning less~\citep{biderman2024lora}. Experience replay~\citep{rolnick2019replay} and distillation toward the previous model~\citep{li2018lwf} are standard remedies for forgetting; we use both in one recipe and report what each configuration retains. Temperature scaling~\citep{guo2017calibration} adjusts predictive sharpness using a scalar fitted on held-out data. We distinguish the raw checkpoint from the temperature-scaled deployment artifact. A low aggregate calibration error is an empirical property of a specified evaluation distribution, not a guarantee for arbitrary application inputs.

\paragraph{Uncertainty in evaluation and selective prediction.}
Bootstrap resampling~\citep{efron1993bootstrap} estimates sampling variability without distributional assumptions; when observations share a source unit, the unit rather than the observation must be resampled~\citep{field2007bootstrapping}. Benchmark decisions derived from one parent record are such clusters. Selective classification~\citep{geifman2017selective} evaluates whether confidence can rank predictions so that abstaining on the least confident ones reduces error; the excess area under the risk--coverage curve separates ranking quality from accuracy~\citep{geifman2019biasreduced}. We apply these tools to stored predictions; they do not require retraining.

\section{Method}
\label{sec:method}
\subsection{Decision Representation}
A decision example consists of a context $s$, an instruction $u$, an ordered list of candidates $\mathcal{C}=\{(k_i,d_i)\}_{i=1}^{K}$, and a target probability vector $\mathbf{y}\in\Delta^{K-1}$. Here $k_i$ is an application-facing key, $d_i$ its natural-language description, and $2\leq K\leq26$ for one compiled question. The released training and benchmark data contain 2--16 candidates per decision, so letters beyond P are never supervised as answers. Hard labels are one-hot vectors; soft labels retain the upstream distribution. Input validation requires finite, nonnegative targets with total mass one up to a numerical tolerance.

The interface has three question types. \noul{} is a proposition judgment with candidates ordered as false and true; its scalar result is $p(\mathrm{true})$. \choice{} returns the key with the largest probability. \score{} takes rating levels in increasing order and returns
\begin{equation}
\hat r=\sum_{i=1}^{K}(i-1)p_i.
\label{eq:score}
\end{equation}
The full distribution is retained in all cases. This expectation treats adjacent level indices as equally spaced; it does not establish equal semantic distance between arbitrary rating descriptions.

\subsection{Shared Prompt Compilation}
\begin{figure}[t]
\centering
\includegraphics[width=\textwidth]{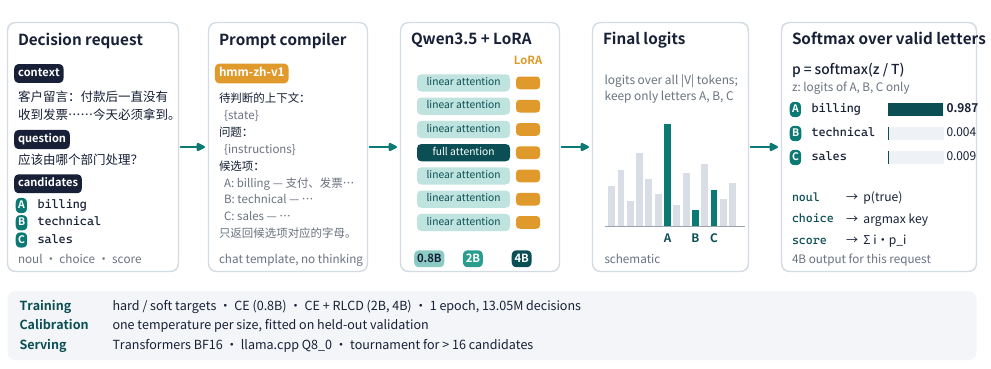}
\caption{\textbf{The shared decision path}, identical for all three sizes. A request is compiled by \texttt{hmm-zh-v1}, the Qwen3.5 backbone with merged LoRA runs once, and only the logits of the valid candidate letters are normalized. The vocabulary bars are a schematic; the three candidate probabilities are the 4B model's actual output for this question. The same distribution yields a Boolean probability, a categorical key, or an expected ordinal level.}
\label{fig:pipeline}
\end{figure}

The compiler associates candidates with the letters A through Z, renders the context, question, candidate keys and descriptions, and applies the backbone's chat template with thinking disabled. Training, benchmarking, and serving call the same compiler, identified as \texttt{hmm-zh-v1}, for all three model sizes; Figure~\ref{fig:pipeline} summarizes the shared path. Appendix~\ref{app:prompt} gives an English translation of its Chinese user-message body.

The implementation accepts at most 16,384 input tokens. Training explicitly counts and filters overlength examples; evaluation fails on overflow and serving rejects overlength requests. The full benchmark runs reported here have zero filtered or truncated decisions. This implementation limit is distinct from the backbone's native context capacity.

\subsection{Single-Pass Candidate Probabilities}
Let $x=\phi(s,u,\mathcal{C})$ be the tokenized prompt of length $L$. A causal language model produces vocabulary logits $\mathbf{v}_{L}\in\R^{|V|}$ at the final position. If $t_i$ is the token ID for candidate letter $i$, we select $z_i=v_{L,t_i}$ and compute
\begin{keyeq}
\begin{equation}
p_i(\mathbf{z};T)=\frac{\exp(z_i/T)}{\sum_{j=1}^{K}\exp(z_j/T)},\qquad T>0.
\label{eq:softmax}
\end{equation}
\end{keyeq}
The tokenizer is checked to ensure that every candidate letter is a single token. Invalid letters are masked before normalization. The output is conditioned on the provided candidate set and excludes vocabulary mass assigned to other tokens. These are decision probabilities learned under candidate supervision, rather than unrestricted next-token probabilities or probabilities of open-ended answer correctness.

Only final-position language-model logits are requested from the backbone. The input sequence still passes through its full text network. The method removes iterative answer generation; it does not remove prompt processing or guarantee lower latency than a compact encoder.

\subsection{Distribution Supervision and LoRA}
All members minimize distribution cross-entropy:
\begin{equation}
\CE(\theta)=-\frac{1}{N}\sum_{n=1}^{N}\sum_{i=1}^{K_n}y_{ni}\log p_{\theta,i}(x_n;1).
\label{eq:ce}
\end{equation}
Because $\mathbf{y}$ may be soft, this objective retains information discarded by replacing every target with its argmax. For fixed targets it differs from $D_{\mathrm{KL}}(\mathbf{y}\Vert\mathbf{p})$ only by target entropy. It does not explicitly penalize ordinal distance between rating levels.

For a frozen backbone matrix $W_0$, LoRA learns $W=W_0+(\alpha/r)BA$. All three sizes use rank $r=32$, $\alpha=64$, and dropout 0.05. Adapters target full-attention, linear-attention, and feed-forward projections. Each release merges its selected adapter into the backbone; Appendix~\ref{app:reproduce} lists the target modules.

During training, \choice{} candidates and their target probabilities are jointly permuted. \noul{} retains false/true order and \score{} retains level order. Permutation augmentation is intended to reduce dependence on categorical candidate position, but the experiments below do not measure permutation invariance directly.

\subsection{RLCD Objective}
\label{sec:rlcd}
All released models except the first-trained decision-only 0.8B model use an RLCD term adapted from the upstream implementation. For logits $\mathbf{z}$, it samples $M=4$ Gaussian perturbations, centers each perturbation over valid candidates, and forms
\begin{equation}
\mathbf{z}^{(m)}=\operatorname{sg}(\mathbf{z})+\tilde{\boldsymbol\epsilon}^{(m)},
\quad \mathbf{q}^{(m)}=\operatorname{softmax}(\mathbf{z}^{(m)}),
\end{equation}
where $\operatorname{sg}$ stops gradients. The reward combines a clipped log score, a spherical score, and a ranked probability score for ordinal questions:
\begin{align}
R(\mathbf{q},\mathbf{y})={}&\sum_i y_i\max(\log q_i,-9.21)\nonumber\\
&+0.75\frac{\mathbf{y}^{\top}\mathbf{q}}{\|\mathbf{q}\|_2}
-\ind_{\score}\operatorname{RPS}(\mathbf{q},\mathbf{y}),\label{eq:reward}\\
\operatorname{RPS}(\mathbf{q},\mathbf{y})={}&\frac{1}{K-1}\sum_{k=1}^{K}
\left(\sum_{i=1}^{k}(q_i-y_i)\right)^2.\label{eq:rps}
\end{align}
The final cumulative term is zero for normalized distributions. The implementation uses small numerical floors in logarithms and norms. Because the logarithmic reward is clipped, we make no blanket claim that the implemented composite reward is strictly proper over the entire probability simplex.

Let $A_m$ be the reward after subtracting the per-example perturbation-group mean. Advantages are divided by their standard deviation over the perturbation-by-microbatch tensor plus $10^{-6}$ and detached. The surrogate is
\begin{align}
\mathcal{L}_{\mathrm{RLCD}}&=-\frac1M\sum_m\operatorname{sg}(A_m)\ell_m(\mathbf{z}),\\
\ell_m(\mathbf{z})&=-\frac{\|\mathbf{z}^{(m)}-\mathbf{z}\|_2^2}{2\sigma^2},\label{eq:rlcd}
\end{align}
with an additional average over examples. The objective is $\CE+\mathcal{L}_{\mathrm{RLCD}}$, with unit weights. The noise scale decreases linearly from 0.4 to 0.1 with the fraction of the epoch consumed. The decision-only 0.8B member, trained first, uses cross-entropy alone; all later released models, including the unified 0.8B model, use RLCD (the CJ-IE-only ablation in Section~\ref{sec:retention-cost} is also CE only). Thus decision-only 0.8B--2B comparisons change both size and objective, while 2B--4B comparisons keep the objective fixed. No CE-only run exists at 2B or 4B, so the contribution of RLCD itself is not measured.

\subsection{Held-Out Temperature and Output Semantics}
\label{sec:temperature}
The official validation split is partitioned by a deterministic hash of parent-record identity. Four of five hash buckets are eligible for model selection; the remaining bucket is reserved for temperature fitting. Reservoir sampling selects up to 100 decisions per domain/type stratum for selection, yielding 1,200 examples. Checkpoints are chosen by distribution negative log-likelihood (NLL), not test accuracy or ECE.

For each release, a scalar temperature is fitted on 1,993 hard-labeled decisions from the reserved bucket by minimizing cross-entropy. The fitted values are $T=1.108133$ for 0.8B, $1.074872$ for 2B, and $1.024758$ for 4B. No test labels are used for fitting. Released-model results use these temperatures; checkpoint trajectories use $T=1$ on the unmerged adapters.

The API also returns a normalized sharpness statistic
\begin{equation}
c=\frac{K\max_i p_i-1}{K-1}.\label{eq:confidence}
\end{equation}
It is zero for a uniform distribution and one for a point mass. It is not an empirical correctness probability. Benchmark ECE uses $\max_i p_i$, not $c$.

\subsection{Candidate Tournaments}
\label{sec:tournament}
A \choice{} question with more candidates than one prompt holds is answered by elimination rounds that reuse the unchanged compiler. A field of $n$ candidates larger than the group size $g$ (default 16, the largest trained candidate count) is shuffled with a request seed and dealt into $\lceil n/g\rceil$ balanced groups. Each group is asked as an ordinary \choice{} question; its $k$ most probable candidates advance (default $k=2$, with $k\leq g/2$, so each round at least halves the field). The process repeats until the remaining field fits into one question, whose probability distribution is the answer. A 41-intent example thus uses three groups of 14, 14, and 13 candidates and a six-candidate final, for four forward passes in total.

\begin{figure}[t]
\centering
\includegraphics[width=\textwidth]{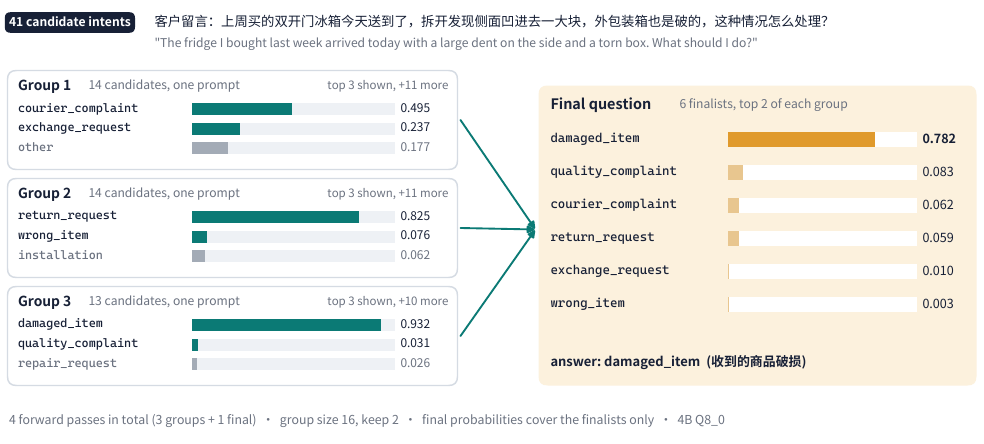}
\caption{\textbf{A candidate tournament} for the repository's 41-intent customer-service example, run through the released 4B model (Q8\_0, CPU). Each group is one ordinary \choice{} question of at most 16 candidates; the two most probable candidates of each group (teal) advance to a six-candidate final. Four forward passes in total select \texttt{damaged\_item} with probability 0.78 among the finalists.}
\label{fig:tournament}
\end{figure}

Figure~\ref{fig:tournament} traces that example through the released 4B model. The final probabilities cover only the finalists; they are not a distribution over all $n$ candidates, and an eliminated candidate receives no probability. Because CJ-Bench contains at most 16 candidates, every benchmark decision fits into a single question, and no result in this report measures tournament accuracy.

\renewcommand{\sectioncolor}{cdIndigo}
\subsection{A Second Readout: Typed Spans}
\label{sec:span}
\begin{figure}[t]
\centering
\includegraphics[width=\textwidth]{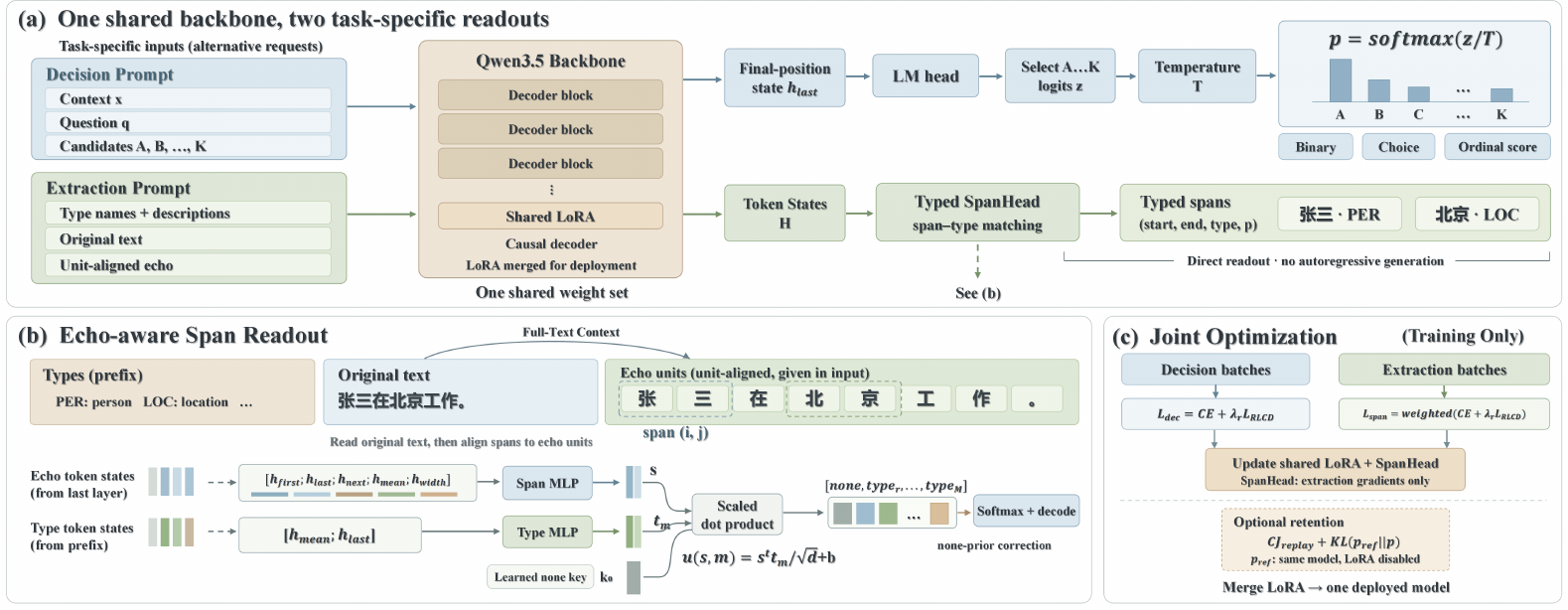}
\caption{\textbf{The unified model.} (a) A decision request and an extraction request are alternative inputs to one Qwen3.5 backbone with one merged LoRA. Decisions read the logits of the candidate letters at the last position through the pretrained LM head; extraction reads last-layer token states through a small typed span head. (b) The extraction prompt lists the types, then the original text, then an echo with one unit per character. Spans are scored against type vectors pooled from the type tokens and against a learned none key. (c) Training mixes decision, replay, and extraction batches; only the span head sees extraction gradients. The optional KL retention term in (c) belongs to the replay-and-KL recipe of Section~\ref{sec:retention-cost}; the released unified models train on all of CJ-Dataset instead of using it.}
\label{fig:architecture}
\end{figure}
Besides the candidate-letter readout of Section~\ref{sec:method}, a unified model answers extraction requests (Figure~\ref{fig:architecture}). An extraction request carries a text $x$ and a set of types $\mathcal{T}=\{(n_m,d_m)\}_{m=1}^{M}$, each with a name $n_m$ and a definition $d_m$, and asks for the character spans of $x$ that belong to each type.

\paragraph{Echo layout.} The compiler writes an instruction, the type list (name and definition on one line each), the text with natural tokenization, and then the same text again with one \emph{unit} per position (Appendix~\ref{app:prompt}). A unit is one CJK character or symbol, one ASCII word or number, or one whitespace character, and each unit is tokenized on its own, so every unit boundary is a token boundary; with natural tokenization alone, 9--14\% of entity boundaries fall inside a token in our feasibility study. Because the model is causal, a representation in the echo has already seen the whole text and every type, which a representation in the first copy has not. Candidate spans are all $(i,j)$ with $j-i<20$ units that neither start nor end on whitespace. This width limit and the unit boundaries make some gold spans unrepresentable: 1,371 of the 232,291 test spans (0.59\%), which count as misses in every score below.

\paragraph{Span head.} Let $\mathbf{h}$ be the last-layer states of the echo tokens. For a candidate span, the head concatenates the states at its first and last tokens, the state at the first token of the following unit, the mean state over the span, and a learned 32-dimensional embedding of its width. A type is represented by the mean and the last-token state of its name-and-definition tokens in the prefix. Two one-hidden-layer MLPs (512 units, GELU, dropout 0.1) map both to 256 dimensions, giving $\mathbf{s}_{ij}$ and $\mathbf{t}_m$, and a learned key $\mathbf{k}_0$ stands for the none class:
\begin{equation}
u_{ij,m}=\frac{\mathbf{s}_{ij}^{\top}\mathbf{t}_m}{\sqrt{256}}+b,\qquad
u_{ij,0}=\frac{\mathbf{s}_{ij}^{\top}\mathbf{k}_0}{\sqrt{256}},\qquad
p(m\mid i,j)=\operatorname{softmax}_{m}\bigl(u_{ij,0},u_{ij,1},\dots,u_{ij,M}\bigr).
\label{eq:span}
\end{equation}
In Eq.~\eqref{eq:span}, type vectors come from the same backbone states as span vectors; there is no second encoder, no CRF, and no pointer network. The head has 3.43M parameters at 0.8B (hidden size 1,024), 6.57M at 2B (hidden size 2,048), and 8.15M at 4B (hidden size 2,560), 0.4\%, 0.3\%, and 0.2\% of the backbones.

\paragraph{Loss and decoding.} Each candidate's distribution is trained with the same CE+RLCD objective as decisions (Section~\ref{sec:rlcd}), on a target that is one-hot (none or the gold type) or uniform over several gold types. Candidates whose target is none are the large majority, so they are weighted by $0.2$ in the mean. At inference the none utility is raised by $s=\log(1/0.2)=1.61$ to undo that weighting, which keeps the reported probabilities calibrated. Each candidate then takes its most probable class, non-none candidates are kept (optionally above a probability threshold), and overlapping spans are resolved greedily by probability, so the output is a flat set of spans, each with its probability.

\paragraph{Decision shift.} The calibrated shift is also a decision threshold, and a high one: at $s=\log 5$ precision exceeds recall by 17 to 18 points for the unified models. Since October 31, 2026, the released configurations therefore carry a second parameter, a \emph{decision shift} $s_d=0.75$ that alone decides which candidates are entities (a candidate is kept when the utility of its best type exceeds the none utility by more than $s_d$), while the reported probability, the order of the overlap removal, and the optional probability threshold still use $s$. Without the field, decoding is unchanged. We call $s=\log 5$ the \emph{model-card operating point} (the one the training evaluations and the model cards report) and $s_d=0.75$ the \emph{release default}. The value comes from a sweep over nine shifts on the CJ-IE dev split, where F1 differs by at most 0.31 points among shifts from 0.5 to 1.0, and one value is used for all three sizes; Section~\ref{sec:decoding} compares the two operating points on the test split.

\subsection{One Weight Set: Joint Training and Export}
\label{sec:joint}
\paragraph{Streams.} Training builds length-bucketed micro-batches from three streams: CJ-Dataset decisions (32 per micro-batch), CJ-IE decisions (32), and CJ-IE span rows (16), under a padded-token budget (24,576 at 0.8B and 4B, 12,288 at 2B; the 4B run fits the larger budget with gradient checkpointing). Micro-batches are shuffled and shared among four ranks, and each optimizer update accumulates eight micro-batches per rank. Because micro-batches follow data volume (407,750 : 19,105 : 35,831 at 0.8B, and the same proportions at 2B and 4B), about 88\% carry CJ-Dataset decisions, 4\% extraction-related decisions, and 8\% span rows. An average update therefore sees about 940 decisions but only 40 span rows. LoRA gradients are averaged over all micro-batches of an update, and span-head gradients over its extraction micro-batches only.

\paragraph{Selection and temperature.} Checkpoints are scored every 1,000 updates on held-out data that never include CJ-Bench: a stratified 3,600-decision subset of CJ-Dataset validation (partitions 1--4, as in Section~\ref{sec:temperature}) and the CJ-IE dev split. The selected checkpoint maximizes the mean of CJ-dev accuracy and an extraction score (the mean of extraction-decision accuracy and macro extraction F1); for all three unified models it is the final update. For runs that continue from a decision model, selection instead keeps the checkpoints within one point of the starting CJ-dev accuracy and maximizes the extraction score among them. The decision temperature is fitted on 1,984 hard decisions of validation partition 0; extraction has no temperature, only the none shift.

\paragraph{Export.} The selected LoRA is merged into the BF16 backbone, and the FP32 span-head tensors are written with a \texttt{span\_head.} prefix into the same safetensors file. \texttt{decision\_config.json} records the temperature, prompt version, and extraction settings (head sizes, none shift, echo). The directory loads with the standard causal-LM class for decisions, which reports the span-head tensors as unused, or with the project's pipeline for both readouts. A merge probe compares adapter and merged logits (largest difference 0.125 at 0.8B and 2B and 0.0625 at 4B, BF16 rounding). The span readout is not implemented for llama.cpp, so a GGUF conversion of a unified model would answer decisions only.

\paragraph{Routing, combined requests, and cascades.} The request, not the output, selects the readout: a request with \texttt{questions} is a decision, one with \texttt{types} is an extraction, a request with both receives both answers about the same input (\texttt{text} defaults to \texttt{state} and the reverse), and a request with neither is an error. Any input can be pushed through both readouts, but the other readout's numbers mean nothing for it and cannot be compared, so routing by confidence would be unsound. A combined request that is short enough is answered in one backbone pass: the decision rows and the span row are left-padded to one width, the language-model head reads the last position of every decision row, and the span head reads the span row. This is used while the padded rows fit in 3,072 tokens; beyond that, padding costs more than the saved pass, and the questions and the span row run as separate passes. The two paths agree closely but not exactly, because padding changes BF16 rounding (Section~\ref{sec:unified-serving}). Cascades read one answer to build the next request. A relation request fixes the subject and puts it into the type name (``director of the film X''), so the extracted span is the object. An event is found by a yes/no decision per event type followed by an extraction of trigger and argument roles. Intent and slot filling choose the slot types by the decided intent. The CJ-IE relation and event scores below are conditional in this sense (objects given the subject, arguments given the event type) and are not end-to-end triple or record scores.
\renewcommand{\sectioncolor}{cdTeal}

\section{Experimental Setup}
\label{sec:setup}
\subsection{Data and Mixing}
We use the released CJ training and benchmark splits rather than constructing a new decision corpus. The training split contains 13,048,000 decisions: 10,000,000 general, 3,000,000 medical, and 24,000 each in law and finance. Law and finance together are less than 0.4\% of training. Table~\ref{tab:data} reports split sizes and hard-label counts used in evaluation. Counts refer to decisions, not unique source documents or independent real-world events.

\begin{table}[t]
\begin{minipage}[t]{0.35\textwidth}
\captionof{table}{\textbf{Data.} Training and full CJ-Bench sizes; accuracy and ECE use hard test targets only.}
\label{tab:data}
\centering
{\footnotesize
\setlength{\tabcolsep}{2.3pt}
\begin{tabular}{lrrr}
\arrayrulecolor{cdInk}\toprule
\rowcolor{cdMist}Domain & Train & Test & Hard test \\
\arrayrulecolor{cdLine}\midrule
General & 10,000,000 & 100,000 & 96,922 \\
Medical & 3,000,000 & 200,000 & 200,000 \\
Legal & 24,000 & 4,300 & 4,300 \\
Finance & 24,000 & 3,600 & 3,600 \\
\arrayrulecolor{cdLine}\midrule
\textbf{Total} & \textbf{13,048,000} & \textbf{307,900} & \textbf{304,822} \\
\arrayrulecolor{cdInk}\bottomrule
\end{tabular}
}

\end{minipage}\hfill
\begin{minipage}[t]{0.625\textwidth}
\captionof{table}{\textbf{Training configurations} of the three decision-only models.}
\label{tab:training}
\centering
{\footnotesize
\setlength{\tabcolsep}{3.5pt}
\begin{tabular}{lccc}
\arrayrulecolor{cdInk}\toprule
\rowcolor{cdMist}Setting & \szA\,0.8B & \szB\,2B & \szC\,4B \\
\arrayrulecolor{cdLine}\midrule
Backbone & Qwen3.5-0.8B & Qwen3.5-2B & Qwen3.5-4B \\
Objective & CE & CE + RLCD & CE + RLCD \\
LoRA $r$ / $\alpha$ / dropout & 32 / 64 / 0.05 & 32 / 64 / 0.05 & 32 / 64 / 0.05 \\
Peak learning rate & $10^{-4}$ & $10^{-4}$ & $10^{-4}$ \\
Min.\ decisions / update & 1,024 & 2,048 & 2,048 \\
Padded-token budget & 262,144 & 524,288 & 262,144 \\
Updates (1 epoch) & 10,132 & 4,994 & 5,446 \\
Mean decisions / update & 1,288 & 2,613 & 2,396 \\
Selected step / val.\ NLL & 10,000 / 0.600 & 4,994 / 0.586 & 5,446 / 0.458 \\
Fitted temperature $T$ & 1.1081 & 1.0749 & 1.0248 \\
RLCD perturbations / $\sigma$ & -- & 4 / $0.4\to0.1$ & 4 / $0.4\to0.1$ \\
\arrayrulecolor{cdInk}\bottomrule
\end{tabular}
\par\smallskip\begin{minipage}{\linewidth}\tablenote{All three runs: one epoch over the same 13,048,000 decisions, seed 42, four GPUs, compiler \texttt{hmm-zh-v1}. The minimum is a threshold that token-limited microbatches can overshoot; gradients are normalized by the actual count. 2B and 4B differ only in backbone and token budget.}\end{minipage}
}

\end{minipage}
\end{table}

Training data are sharded by domain, source, and question type. A streaming mixer samples shards in proportion to their remaining counts and visits each decision once per epoch. This interleaves tasks while retaining corpus proportions; it does \emph{not} equalize domain frequencies or oversample the smaller domains. Length bucketing and padded-token budgets reduce padding overhead. Validation files are not mixed into training. All decision-only models see the same decisions in the same per-epoch proportions.

A separate full-file candidate-count audit checks release checksums and finds 2--16 candidates in the training, validation, and test splits, with no decision exceeding 26 candidates. Thus the 26-letter interface does not exclude released benchmark decisions. This audit verifies format and coverage, not freedom from pretraining contamination.

\subsection{CJ-IE: Chinese Extraction Data for Both Readouts}
\label{sec:cjie}
The unified models also train on CJ-IE, which we assembled from public Chinese information-extraction corpora (Appendix~\ref{app:cjie}). It has two parts (Table~\ref{tab:cjie}). \emph{Span rows} are (text, type list, gold spans) triples for the span head: NER rows from 32 corpora, relation rows from DuIE and CMeIE, and event-argument rows from DuEE and FewFC. \emph{Extraction-related decisions} use the letter readout in the CJ format: the type of a marked mention, the relation between a given entity pair, whether a sentence describes an event type, its event category, and two document-level tasks over announcements. The same source record falls into one split in every view.

\begin{table}[t]
\caption{\textbf{CJ-IE.} Rows per split. A span row is one text with one type list; relation rows carry the subject in the type name. Test rows come from official dev or test records after leakage removal.}
\label{tab:cjie}
\centering
{\footnotesize
\setlength{\tabcolsep}{3.4pt}
\begin{tabular}{llcrrr}
\arrayrulecolor{cdInk}\toprule
\rowcolor{cdMist}Readout & Task view & Sources & Train & Dev & Test \\
\arrayrulecolor{cdLine}\midrule
span head & NER & 32 corpora & 402,770 & 9,918 & 64,796 \\
span head & relation objects & DuIE, CMeIE & 150,561 & 1,138 & 11,402 \\
span head & event arguments & DuEE, FewFC & 19,963 & 591 & 8,843 \\
\arrayrulecolor{cdLine}\midrule
letter readout & \texttt{re\_pair} & DuIE, CMeIE, FinRE, SanWen & 345,513 & 5,233 & 40,417 \\
letter readout & \texttt{ner\_typing} & 31 NER corpora & 184,660 & 4,728 & 20,312 \\
letter readout & \texttt{ee\_type} & DuEE, FewFC & 30,812 & 915 & 13,604 \\
letter readout & \texttt{ee\_category} & DuEE, FewFC & 13,339 & 404 & 5,863 \\
letter readout & \texttt{doc\_field} & ChFinAnn & 21,372 & 743 & 2,756 \\
letter readout & \texttt{doc\_event} & ChFinAnn & 15,658 & 531 & 1,942 \\
\arrayrulecolor{cdLine}\midrule
 & \textbf{span rows} &  & \textbf{573,294} & \textbf{11,647} & \textbf{85,041} \\
 & \textbf{decisions} &  & \textbf{611,354} & \textbf{12,554} & \textbf{84,894} \\
\arrayrulecolor{cdInk}\bottomrule
\end{tabular}
}

\end{table}

All NER corpora use the standardized Chinese type names and definitions of B2NERD~\citep{yang2025b2nerd}, mapped by label for the corpora we downloaded separately (over 99\% of entities agree with B2NERD on the training text), so that ``person'' has one meaning across corpora and annotation differences appear in the definitions. B2NERD gives type and name but no position, so each name is located in the text: every occurrence becomes an entity unless it overlaps a longer one already placed. Texts longer than 300 characters are split at sentence ends. Leakage control normalizes text and compares whole records and sentences of 12 or more characters: a development record that overlaps any test record, or a training record that overlaps any development or test record, is dropped. Sampling caps per source and split (50,000 / 500 / 5,000 records, with exceptions) keep large sources from dominating. CJ-IE is an \emph{internal} evaluation set: its types are standardized, its test split is sampled, and its scores are not comparable to published leaderboards.

\subsection{Training Configurations}
Table~\ref{tab:training} gives the decision-only runs. Every run completes one epoch over all 13,048,000 decisions with BF16, gradient checkpointing, AdamW, peak learning rate $10^{-4}$, gradient norm clipping at 1.0, 100-step warmup, and cosine decay toward 10\% of peak. Every run evaluates validation NLL every 1,000 updates and at the end of training. The 0.8B run reaches its minimum validation NLL at step 10,000 of 10,132. For 2B and 4B the minimum is at the final update (4,994 and 5,446).

The nominal global batch is a minimum decision count accumulated before an update, rather than a fixed exact size. Token-limited microbatches can overshoot this threshold; gradients are normalized by the actual decision count. The decision-only comparisons therefore differ as follows.
\begin{caveat}[What each comparison controls]
\begin{itemize}[topsep=2pt]
\item \emph{0.8B vs.\ 2B} (decision-only) changes backbone size, objective (CE vs.\ CE+RLCD), minimum batch (1,024 vs.\ 2,048), and token budget. It compares two complete recipes.
\item \emph{2B vs.\ 4B} keeps data, objective, RLCD schedule, minimum batch, learning rate, adapter settings, and seed, and changes the backbone and the padded-token budget (524,288 vs.\ 262,144). The smaller budget produces smaller microbatches and less overshoot, so 4B takes 5,446 updates averaging 2,396 decisions, against 4,994 updates averaging 2,613 decisions for 2B. The learning-rate and noise schedules follow consumed data, so this changes the number of updates rather than the data seen.
\end{itemize}
\end{caveat}
Each decision-only configuration is trained once with seed 42 on four GPUs.

Table~\ref{tab:recipes} lists the five joint-training runs that Section~\ref{sec:retention-cost} compares. \emph{CJ-IE only} continues the decision 0.8B model on CJ-IE with cross-entropy and no CJ data. \emph{Replay + KL} continues the same model with RLCD, a 348,000-decision replay sample of CJ-Dataset (200,000 general, 100,000 medical, and all of law and finance), and a KL term that ties the replay batches to the starting model, read with the LoRA adapter disabled so no second model is loaded. The three \emph{unified} runs start from the raw backbones and treat all of CJ-Dataset as ordinary training data, without a KL term; they share one recipe and differ in backbone size and in what that size forces (the padded-token budget of the 2B run is half that of the others, and the 4B run uses gradient checkpointing). Measured in the training logs from the start of training to the last update (periodic dev evaluation included, data indexing and final export excluded), the unified runs took 17.0 (0.8B), 26.1 (2B), and 82.6 hours (4B) on four A800 GPUs. The five runs differ in starting point, data, objective, batch composition, and update count at once, so the comparison describes recipes and does not isolate any single factor.

\begin{table}[t]
\caption{\textbf{Joint-training recipes.} The five runs that train the span head, with the settings that differ between them.}
\label{tab:recipes}
\centering
{\footnotesize
\setlength{\tabcolsep}{3.2pt}
\begin{tabular}{lccccc}
\arrayrulecolor{cdInk}\toprule
\rowcolor{cdMist}Setting & \runA\,CJ-IE only & \runB\,replay + KL & \uniA\,unified 0.8B & \uniB\,unified 2B & \uniC\,unified 4B \\
\arrayrulecolor{cdLine}\midrule
Start from & decision 0.8B & decision 0.8B & raw 0.8B & raw 2B & raw 4B \\
CJ-Dataset decisions & none & 348k replay & all 13.05M & all 13.05M & all 13.05M \\
CJ-IE data & all & all & all & all & all \\
KL to the start model & -- & 1.0 & -- & -- & -- \\
RLCD & off & on & on & on & on \\
Micro-batch (decisions / spans) & 16 / 8 & 16 / 8 & 32 / 16 & 32 / 16 & 32 / 16 \\
Micro-batches per update & 1 & 8 & 8 & 8 & 8 \\
Padded-token budget & 12,288 & 12,288 & 24,576 & 12,288 & 24,576 \\
Gradient checkpointing & off & off & off & off & on \\
LR warmup (updates) & 500 & 100 & 100 & 100 & 100 \\
Updates & 27,514 & 4,169 & 14,504 & 15,304 & 14,504 \\
Selected checkpoint & final & dev: keep CJ & dev: joint & dev: joint & dev: joint \\
\arrayrulecolor{cdInk}\bottomrule
\end{tabular}
\par\smallskip\begin{minipage}{\linewidth}\tablenote{All runs: one epoch over CJ-IE, LoRA $r=32$, learning rates $10^{-4}$ (LoRA) and $5\times10^{-4}$ (span head), seed 42, four GPUs, span-head loss with none weight 0.2. The 4B run keeps the 0.8B token budget and fits it with gradient checkpointing. Recipes differ in several factors at once; Section~\ref{sec:retention-cost} discusses what the comparison can and cannot isolate.}\end{minipage}
}

\end{table}

\subsection{Metrics and Evaluation Protocol}
Every full CJ-Bench run covers all 307,900 decisions, including 304,822 hard targets and 3,078 soft targets. Candidate order is preserved during testing. Inputs are neither filtered nor truncated; the longest compiled prompt is 6,850 tokens and the mean length is 222.53 tokens.

Accuracy uses argmax predictions on hard targets only. ECE uses the same hard subset and 15 equal-width confidence bins:
\begin{equation}
\mathrm{ECE}=\sum_{b=1}^{15}\frac{|B_b|}{N_h}
\left|\mathrm{acc}(B_b)-\mathrm{conf}(B_b)\right|,\label{eq:ece}
\end{equation}
where confidence is the maximum candidate probability. Distribution NLL and the unnormalized multiclass Brier score use all targets:
\begin{align}
\mathrm{NLL}&=-\frac{1}{N}\sum_{n,i}y_{ni}\log p_{ni},\\
\mathrm{Brier}&=\frac{1}{N}\sum_{n,i}(p_{ni}-y_{ni})^2.
\end{align}
For \score{}, expected-score MAE compares expected level indices under the predicted and target distributions. Ordinal accuracy still uses discrete argmax and is a different metric.

We report micro accuracy over all hard targets, four-domain macro accuracy, and the unweighted average across three professional domains. Medicine accounts for 64.96\% of all decisions and 65.61\% of hard targets, so micro and macro performance answer different questions. Pooled (micro) ECE bins decisions from all domains together and is not an average of domain ECEs. Training-time validation summaries use 10 ECE bins; they are not directly interchangeable with 15-bin benchmark ECE. Model selection always uses held-out data and never CJ-Bench.

\paragraph{Extraction metrics.} A predicted span counts as correct only if its start, end, and type all match a gold span (exact match), and unrepresentable gold spans count as misses. We report micro precision, recall, and F1 pooled over the 232,291 gold spans of the CJ-IE test split, macro F1 averaged over its 35 sources, and macro F1 per family (NER, relation objects, event arguments). Extraction scores are reported at two operating points: the \emph{model-card} point (none shift $s=\log 5$ decides which spans are entities) and the \emph{release default} (decision shift $s_d=0.75$, Section~\ref{sec:span}); the first is the one the training evaluations and the model cards use. Entity ECE bins the probability of each predicted span (10 bins) against whether it is exactly correct. Relation scores are for objects given the subject, and event scores are for arguments given the event type; neither is an end-to-end triple or record score. Extraction-related decisions are scored with hard-target accuracy.

\paragraph{Evaluated artifacts.} \emph{Released models} are the merged BF16 weights evaluated with their fitted temperature; they are what users download, and the main comparison tables report them. \emph{Checkpoints} are unmerged adapters evaluated at $T=1$; trajectories use them. The unified models' CJ-Bench evaluations were run at $T=1$ on their merged weights; their fitted-temperature results are computed from the stored log probabilities (Section~\ref{sec:analysis-protocol}). Each Q8\_0 model is converted directly from its BF16 release and keeps its temperature.

\subsection{Analyses of Stored Predictions}
\label{sec:analysis-protocol}
Every full CJ-Bench run stores per-decision log probabilities together with the domain, source, question type, target, and parent identity of each decision. The analyses below re-read these files without running any model. Recomputing every reported metric from them reproduces the stored summaries for all eleven analyzed runs (three decision-only releases, their three final checkpoints, and five unified-run evaluations) to within $10^{-9}$.

\emph{Clustered intervals.} Decisions derived from one parent record are not independent. We use a stratified cluster bootstrap~\citep{efron1993bootstrap,field2007bootstrapping}: in each of 2,000 replicates, parents are resampled with replacement within each domain, and every decision inherits its parent's multiplicity. The same draws are applied to all models, so differences between models are paired. We report 95\% percentile intervals. ECE is biased upward under resampling, so we interpret ECE intervals only for paired differences.

\emph{Temperature-only ablation.} Merging changes logits slightly (Section~\ref{sec:quant}), so released and checkpoint results differ in more than temperature. To isolate temperature, we divide the stored $T=1$ log probabilities of each final checkpoint by its fitted $T$ and renormalize. This is exact for fixed logits and leaves every argmax unchanged. For the unified models, which were evaluated only at $T=1$, the same operation gives their fitted-temperature results.

\emph{Selective prediction.} Hard decisions are ranked by maximum probability. We report accuracy at fixed coverage, the area under the risk--coverage curve (AURC), and the excess AURC over an oracle ranking with the same accuracy (E-AURC)~\citep{geifman2017selective,geifman2019biasreduced}.

\emph{Paired transitions and sources.} For two models we count hard targets that change from wrong to right and from right to wrong. For every source with at least 200 hard targets, we compute a paired cluster bootstrap within that source; these 55 tests are not corrected for multiplicity.

\emph{Span utilities.} For each CJ-IE test row (and dev row), a one-time inference pass stores the gold spans and, for every candidate span that is not clearly none (best type within 8 utility units of none), three numbers: the none utility, the best type's utility, and the log-sum-exp of all type utilities. These determine the decoded output for any none shift $s$ and probability threshold exactly. We verified that re-decoding at the model-card shift reproduces the stored micro F1 to within 0.06 points for four runs, and to within 0.53 points for the CJ-IE-only run, which was evaluated unmerged with a different batch composition (the largest per-source difference is 0.5 points for the four runs and 1.5 for that run). Intervals for span metrics resample the 78,936 parent groups within each source, with the same draws for every model. The dev-selected shift maximizes macro F1 on the CJ-IE dev split. The release default $s_d=0.75$ was fixed by the repository's own dev sweep (Section~\ref{sec:span}) before these analyses, and we only re-decode at it; the test split is not used for any choice.

\subsection{Baselines and Evidence Provenance}
Baseline accuracies and ECE values are imported from Tables 1 and 2 of Chinese-Jev~\citep{wang2026chinesejev}; we do not rerun those models. Chinese-Jev-G denotes its general checkpoint and Chinese-Jev-S denotes three domain specialists. Jev is a hosted model. The imported Qwen3.5-2B baseline is not our decision-fine-tuned 2B model. We report no external baseline for extraction: public NER scores use their own type sets and annotation rules, and the CJ-IE test split is a standardized internal set, so a number placed next to a leaderboard would not be comparable.

Local tables are generated from stored evaluation files, with source paths and SHA256 hashes recorded in the accompanying artifact. The manuscript uses a November 1, 2026 evidence snapshot. The unified models' predictions and training histories were copied read-only from the training server, and the span utilities were produced by inference-only passes of the released unified weights (the unified 4B run finished training on October 30, 2026) and of the two earlier continuation runs' final checkpoints. We did not repeat any training. Imported baseline latency is omitted from the accuracy tables because hardware and, for Jev, network scope differ from our measurements.

\section[Results: Decision Models]{Results: Decision-Only Models}
\label{sec:results}
This section studies the three decision-only models; the unified models follow in Section~\ref{sec:unified-results}, and their rows in Table~\ref{tab:main} are discussed there. We first compare the released members with the reported baselines (Sections~\ref{sec:main}--5.2), then ask which differences are statistically resolved (Section~\ref{sec:uncertainty}), how they develop during training (Section~5.4), and where the larger model gains and loses (Section~5.5). The last two subsections separate calibration from ranking quality (Sections~\ref{sec:calibration} and~\ref{sec:selective}).

\subsection{Accuracy and Calibration by Domain}
\label{sec:main}
\begin{table}[t]
\caption{\textbf{Main results on the full CJ-Bench.} One released checkpoint per size covers all four domains; the last three rows are the unified models (Section~\ref{sec:unified-decision}).}
\label{tab:main}
\centering
{\small
\setlength{\tabcolsep}{3.1pt}
\begin{tabular}{lrrrrrrrrrr}
\arrayrulecolor{cdInk}\toprule
\rowcolor{cdMist}& \multicolumn{2}{c}{General} & \multicolumn{2}{c}{Medical} & \multicolumn{2}{c}{Legal} & \multicolumn{2}{c}{Finance} & \multicolumn{2}{c}{Prof.\ mean} \\
\rowcolor{cdMist}Model & ACC$\uparrow$ & ECE$\downarrow$ & ACC$\uparrow$ & ECE$\downarrow$ & ACC$\uparrow$ & ECE$\downarrow$ & ACC$\uparrow$ & ECE$\downarrow$ & ACC$\uparrow$ & ECE$\downarrow$ \\
\arrayrulecolor{cdLine}\midrule
Qwen3.5-2B & 32.95 & 43.68 & 53.88 & 28.71 & 31.53 & 44.82 & 38.11 & 49.12 & 41.17 & 40.88 \\
Open-Jev & 52.33 & 10.54 & 73.08 & 2.88 & 44.07 & 23.15 & 48.11 & 10.19 & 55.09 & 12.07 \\
SemIF & 44.21 & 27.27 & 62.17 & 10.39 & 43.07 & 19.99 & 47.89 & 18.45 & 51.04 & 16.28 \\
Laya Multilingual & 40.22 & 23.04 & 24.55 & 43.05 & 34.72 & 31.49 & 38.31 & 26.18 & 32.53 & 33.57 \\
Jev (hosted API) & 68.35 & 11.45 & 84.29 & 5.70 & 68.35 & 8.03 & 78.39 & 2.98 & 77.01 & 5.57 \\
Chinese-Jev G + S & 69.20 & 3.78 & 87.65 & 7.25 & 60.10 & 17.45 & 64.72 & 11.55 & 70.82 & 12.08 \\
\arrayrulecolor{cdLine}\midrule
\rowcolor{cdTealLight}\szA\,\textbf{Ours 0.8B} & 72.26 & 1.68 & 88.85 & 1.12 & 67.84 & 4.05 & 74.00 & 2.45 & 76.90 & 2.54 \\
\rowcolor{cdTealLight}\szB\,\textbf{Ours 2B} & 73.96 & 1.93 & 89.50 & \best{0.24} & 68.74 & 3.37 & 76.50 & 2.13 & 78.25 & 1.92 \\
\rowcolor{cdTealLight}\szC\,\textbf{Ours 4B} & 75.69 & 2.91 & 91.19 & 0.25 & 77.21 & 2.02 & 83.61 & 1.65 & 84.00 & \best{1.30} \\
\rowcolor{cdIndigoLight}\uniA\,\textbf{Unified 0.8B} & 73.86 & \best{1.02} & 88.61 & 1.02 & 67.88 & 2.29 & 74.31 & 2.37 & 76.93 & 1.90 \\
\rowcolor{cdIndigoLight}\uniB\,\textbf{Unified 2B} & 75.43 & 1.67 & 90.03 & 0.49 & 73.42 & 2.58 & 77.61 & 1.76 & 80.35 & 1.61 \\
\rowcolor{cdIndigoLight}\uniC\,\textbf{Unified 4B} & \best{77.30} & 1.56 & \best{91.64} & 0.90 & \best{82.44} & \best{1.69} & \best{84.00} & \best{1.44} & \best{86.03} & 1.34 \\
\arrayrulecolor{cdInk}\bottomrule
\end{tabular}
\par\smallskip\begin{minipage}{\linewidth}\tablenote{Accuracy and ECE in \%. Baseline rows are reported values from Chinese-Jev~\citep{wang2026chinesejev}, not reproduced here; Chinese-Jev G + S uses its general model (G) for the general domain and three separately trained specialists (S) for the professional domains. Each of our rows is one released checkpoint at its fitted temperature serving all four domains; unified models also extract spans. The unified rows apply the fitted temperature to their stored $T=1$ log probabilities (exact; see Section~\ref{sec:analysis-protocol}). Bold: best in column.}\end{minipage}
}

\end{table}
Table~\ref{tab:main} compares the released members with the reported baselines. All three members exceed the reported general-domain results. The 4B model reaches 75.69\% general accuracy, 6.49 points above Chinese-Jev General and 7.34 points above Jev; 2B and 0.8B reach 73.96\% and 72.26\%. In the professional domains, the 4B model obtains 91.19\% medical, 77.21\% legal, and 83.61\% financial accuracy. It exceeds Jev in each domain, by 6.90, 8.86, and 5.22 points, and the three Chinese-Jev specialists by 3.54, 17.11, and 18.89 points. Its 84.00\% professional average is 6.99 points above Jev and 13.18 above Chinese-Jev-S.

The smaller members are closer to Jev. The 2B average (78.25\%) is 1.24 points above Jev but trails it by 1.89 points in finance, and the 0.8B average (76.90\%) is 0.11 points below Jev. Every released member has lower ECE than every reported baseline in each of the four domains. General-domain ECE is 1.68\%, 1.93\%, and 2.91\% for 0.8B, 2B, and 4B; it rises with size, a pattern examined in Section~\ref{sec:calibration}.

\begin{caveat}[Reported baselines]
Baseline numbers are values reported by Chinese-Jev~\citep{wang2026chinesejev} from its own evaluation; we did not rerun them, so these comparisons carry no interval. Chinese-Jev uses about 322M-parameter encoders and four separately trained models, while each of our rows is one larger causal model. Intervals for our own models follow in Section~\ref{sec:uncertainty}.
\end{caveat}

\begin{figure}[t]
\centering
\includegraphics[width=\textwidth]{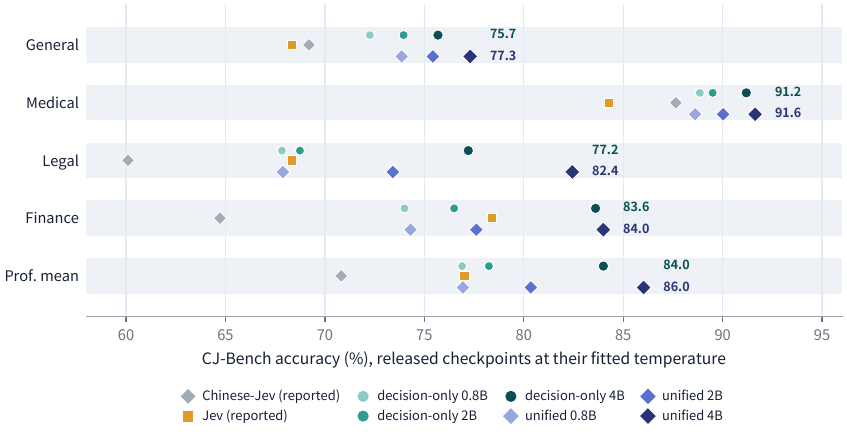}
\caption{\textbf{Accuracy by domain.} The decision-only models (circles) and the unified models (diamonds) against the reported baselines, at each model's fitted temperature. Among the decision-only models the 4B model leads in every domain; the labels give the accuracy of the decision-only 4B (teal) and unified 4B (indigo) models, and the unified 4B model is higher in every domain. The unified 2B model sits between the decision-only 2B and 4B models.}
\label{fig:domains}
\end{figure}

\subsection{Aggregate and Typed Results}
\begin{table}[t]
\caption{\textbf{Aggregate results of the six released files.} BF16 rows are the merged releases; each Q8\_0 row is its direct conversion.}
\label{tab:aggregate}
\centering
{\small
\setlength{\tabcolsep}{4.5pt}
\begin{tabular}{llrrrrrr}
\arrayrulecolor{cdInk}\toprule
\rowcolor{cdMist}Size & Format & Micro ACC & Macro ACC & ECE & NLL & Brier & Score MAE \\
\arrayrulecolor{cdLine}\midrule
\rowcolor{cdTealLight}\szA\,0.8B & BF16 & 83.11 & 75.74 & 0.27 & 0.4313 & 0.2238 & 0.5844 \\
\szA\,0.8B & Q8\_0 & 83.12 & 75.81 & 0.27 & 0.4313 & 0.2238 & 0.5840 \\
\rowcolor{cdTealLight}\szB\,2B & BF16 & 84.11 & 77.18 & 0.64 & 0.4165 & 0.2133 & 0.5642 \\
\szB\,2B & Q8\_0 & 84.11 & 77.33 & 0.65 & 0.4163 & 0.2132 & 0.5637 \\
\rowcolor{cdTealLight}\szC\,4B & BF16 & 85.97 & 81.92 & 1.00 & 0.3706 & 0.1909 & 0.5253 \\
\szC\,4B & Q8\_0 & 85.99 & 81.95 & 0.96 & 0.3705 & 0.1907 & 0.5253 \\
\arrayrulecolor{cdInk}\bottomrule
\end{tabular}
\par\smallskip\begin{minipage}{\linewidth}\tablenote{ACC and ECE in \%. Macro ACC weights the four domains equally; micro ACC weights hard targets, 65.6\% of which are medical. Score MAE covers the 38,746 ordinal questions. Each Q8\_0 file is converted from the BF16 model above it and keeps its temperature.}\end{minipage}
}

\end{table}
Table~\ref{tab:aggregate} reports aggregate metrics for the six released files. From 0.8B to 2B to 4B, micro accuracy rises from 83.11\% to 84.11\% to 85.97\%, and four-domain macro accuracy from 75.74\% to 77.18\% to 81.92\%. The larger rise in macro accuracy reflects gains concentrated in the small legal and financial subsets. Distribution NLL falls from 0.431 to 0.417 to 0.371, Brier from 0.224 to 0.213 to 0.191, and expected-score MAE from 0.584 to 0.564 to 0.525. Pooled ECE moves the other way, from 0.27\% to 0.64\% to 1.00\%. Each Q8\_0 model stays within 0.02 points of its BF16 source on micro accuracy and within 0.0003 on NLL.

\begin{table}[t]
\caption{\textbf{Released models by question type.} Counts give the total and hard-target denominators.}
\label{tab:types}
\centering
{\small
\setlength{\tabcolsep}{4pt}
\begin{tabular}{llrrrrrrrrr}
\arrayrulecolor{cdInk}\toprule
\rowcolor{cdMist}& & \multicolumn{3}{c}{ACC (\%) $\uparrow$} & \multicolumn{3}{c}{ECE (\%) $\downarrow$} & \multicolumn{3}{c}{NLL $\downarrow$} \\
\rowcolor{cdMist}Type & Total / hard & \szA\,0.8B & \szB\,2B & \szC\,4B & \szA\,0.8B & \szB\,2B & \szC\,4B & \szA\,0.8B & \szB\,2B & \szC\,4B \\
\arrayrulecolor{cdLine}\midrule
\texttt{noul} & 134,577 / 134,577 & 94.96 & 95.11 & \best{95.47} & 0.19 & 0.34 & 0.43 & 0.117 & 0.115 & 0.107 \\
\texttt{choice} & 134,577 / 134,577 & 78.32 & 80.12 & \best{83.25} & 0.58 & 0.90 & 1.26 & 0.587 & 0.559 & 0.475 \\
\texttt{score} & 38,746 / 35,668 & 56.45 & 57.67 & \best{60.43} & 1.67 & 1.65 & 2.18 & 0.984 & 0.969 & 0.922 \\
\arrayrulecolor{cdInk}\bottomrule
\end{tabular}
\par\smallskip\begin{minipage}{\linewidth}\tablenote{Score MAE is 0.584, 0.564, and 0.525 level indices. All 3,078 soft targets are general-domain \score{} decisions.}\end{minipage}
}

\end{table}
Question-type results in Table~\ref{tab:types} show that Boolean judgments remain far easier than categorical and ordinal decisions. Accuracy improves with size for every type, most for \choice{} (78.32\% to 83.25\%). \score{} remains the weakest type at 60.43\% for 4B. Among \score{} hard targets, 27.3\% are wrong for all three sizes, against 3.1\% for \noul{}. The domain-by-type breakdown in Appendix Table~\ref{tab:domain-type} shows that 4B is the most accurate member in all twelve cells. The 2B model trails 0.8B in two cells: legal \noul{} (82.07\% vs.\ 85.60\%) and financial \score{} (80.33\% vs.\ 82.83\%). Because candidate counts, source tasks, and domain mixtures differ by type, type gaps should not be read as controlled effects of interface design alone.

\subsection{Statistical Uncertainty and Size Effects}
\label{sec:uncertainty}
\begin{figure}[t]
\centering
\includegraphics[width=0.92\textwidth]{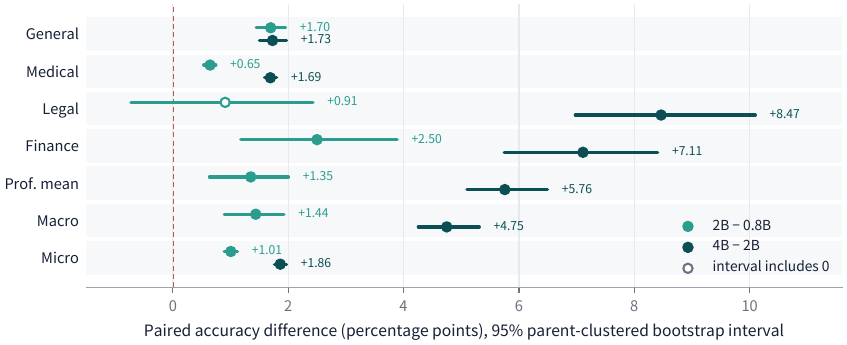}
\caption{\textbf{Paired size effects with 95\% parent-clustered bootstrap intervals} (2,000 replicates, stratified by domain, same draws for every model). Filled points: the interval excludes zero. Every 2B$\to$4B effect is resolved; the 0.8B$\to$2B effect in law is not. Appendix Table~\ref{tab:bootstrap} lists the values.}
\label{fig:forest}
\end{figure}
Figure~\ref{fig:forest} attaches parent-clustered 95\% intervals to the differences between released models. Clustering matters. In the general domain, the clustered standard error of accuracy is 0.59--0.65 points, 4.2--4.6 times the 0.14 points obtained by treating 96,922 decisions as independent. For law the ratio is about 1.2--1.4, because most legal parents contribute few decisions. Interval half-widths for a single model are about 1.1 points for general, 0.2 for medical, and 1.3--2.0 for the legal and financial subsets.

Paired differences resolve most, but not all, size effects. From 2B to 4B, every domain and every aggregate improves with an interval excluding zero. The gains are largest in law (+8.47 points, 95\% CI [+6.98, +10.10]) and finance (+7.11, [+5.75, +8.41]), the two domains with 24,000 training decisions each. From 0.8B to 2B, general, medical, financial, and all aggregate accuracies improve significantly, but the legal difference (+0.91, [$-$0.73, +2.43]) is not resolved. On micro accuracy, the 2B-to-4B step (+1.86) is nearly twice the 0.8B-to-2B step (+1.01). Among paired ECE differences, pooled ECE increases significantly at each step (+0.37 and +0.36 points), general-domain ECE increases from 2B to 4B (+0.99), and medical ECE decreases from 0.8B to 2B ($-$0.88). Legal and financial ECE differences between adjacent sizes are not resolved.

\begin{finding}
Moving from 2B to 4B, with data, objective, and minimum batch held fixed, buys resolved gains in every domain, and the largest gains are in law and finance, the two domains with the least training data. These intervals capture test-set sampling for fixed trained models; they do not capture variation across training seeds.
\end{finding}

\subsection{Training Trajectories}
\begin{figure}[t]
\centering
\includegraphics[width=\textwidth]{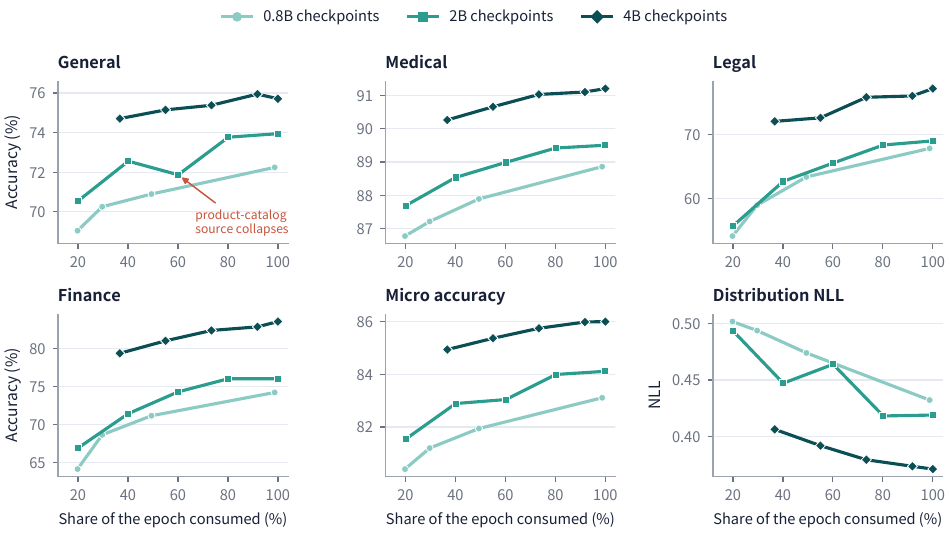}
\caption{\textbf{Full-benchmark trajectories of all 14 evaluated checkpoints}, plotted against the share of the shared 13,048,000-decision epoch consumed rather than optimizer steps. All points are unmerged adapters at $T=1$; lines connect measured checkpoints from one run per size and are not replicates.}
\label{fig:trajectory}
\end{figure}
Figure~\ref{fig:trajectory} and Appendix Table~\ref{tab:trajectory} align the three runs by consumed data. At about 20\% of the epoch, 2B is ahead of 0.8B in all four domains, for example 70.55\% vs.\ 69.04\% general accuracy. The 4B run is ahead of the finished 2B model at its first fully benchmarked checkpoint, which has seen 36.7\% of the epoch. It leads 2B in every domain: 74.71\% vs.\ 73.95\% general, 90.26\% vs.\ 89.50\% medical, 72.12\% vs.\ 69.05\% legal, and 79.36\% vs.\ 76.03\% finance. Legal and financial accuracy rise the most within every run: by 13.79 and 10.06 points for 0.8B, 13.37 and 9.08 for 2B, and 5.16 and 4.17 for 4B over its shorter evaluated span.

All runs flatten late. The last evaluated interval adds 0.12 micro-accuracy points for 2B (80\% to 100\% of the epoch) and 0.02 for 4B (92\% to 100\%). Validation NLL decreases across the evaluated checkpoints of every run, but test metrics do not. The 2B general accuracy drops from 72.56\% to 71.87\% at 60\% of the epoch, before recovering. Over the same interval, accuracy on the 3,151-decision \texttt{mep3m\_product\_catalog} source falls from 75.06\% to 30.05\%, while its NLL rises from 0.99 to 4.35. The same source oscillates in the 4B run (77.85\%, 62.36\%, 73.31\%, 70.01\%, 64.58\%). Pooled ECE is also non-monotonic in every run. The final 4B step slightly lowers general accuracy relative to step 5,000 (75.71\% vs.\ 75.95\%); the released checkpoint nevertheless remains the one selected by validation NLL.

\subsection{Where the Larger Model Helps}
\begin{table}[t]
\centering
\begin{minipage}{0.7\textwidth}
\caption{\textbf{Paired top-1 transitions} between released models of adjacent sizes on the full benchmark.}
\label{tab:transitions}
\centering
{\small
\setlength{\tabcolsep}{4.5pt}
\begin{tabular}{llrrrr}
\arrayrulecolor{cdInk}\toprule
\rowcolor{cdMist}Step & Group & Top-1 agree (\%) & W$\to$R & R$\to$W & Net \\
\arrayrulecolor{cdLine}\midrule
0.8B $\to$ 2B & All & 89.64 & 13,673 & 10,608 & \textbf{+3,065} \\
 & General & 83.66 & 6,759 & 5,114 & \textbf{+1,645} \\
 & Medical & 93.13 & 6,124 & 4,833 & \textbf{+1,291} \\
 & Legal & 75.28 & 428 & 389 & \textbf{+39} \\
 & Finance & 79.31 & 362 & 272 & \textbf{+90} \\
\arrayrulecolor{cdLine}\midrule
2B $\to$ 4B & All & 90.74 & 14,317 & 8,640 & \textbf{+5,677} \\
 & General & 85.65 & 6,325 & 4,650 & \textbf{+1,675} \\
 & Medical & 93.79 & 6,929 & 3,547 & \textbf{+3,382} \\
 & Legal & 75.65 & 627 & 263 & \textbf{+364} \\
 & Finance & 80.89 & 436 & 180 & \textbf{+256} \\
\arrayrulecolor{cdInk}\bottomrule
\end{tabular}
\par\smallskip\begin{minipage}{0.98\linewidth}\tablenote{W$\to$R: hard targets fixed by the larger model; R$\to$W: new errors. 9.68\% of hard targets are wrong for all three sizes; 90.32\% are right for at least one.}\end{minipage}
}

\end{minipage}
\end{table}
Table~\ref{tab:transitions} decomposes accuracy changes into paired transitions. Size changes alter many individual predictions in both directions: 0.8B and 2B disagree on the top-1 candidate in 10.4\% of decisions, and 2B and 4B in 9.3\%. From 2B to 4B, 14,317 hard targets are fixed and 8,640 are broken, a net gain of 5,677. Net gains are 8.5\% of legal and 7.1\% of financial hard targets, compared with 1.7\% for both general and medical. The ratio of fixed to broken targets is 2.4 for law and finance, against 1.3 for 0.8B to 2B overall. Across all three sizes, 9.68\% of hard targets are never answered correctly and 90.32\% are answered correctly by at least one member.

\begin{table}[t]
\caption{\textbf{Source-level changes from the 2B to the 4B release}: the six largest gains and every source whose 95\% paired interval is entirely negative.}
\label{tab:sources}
\centering
{\small
\setlength{\tabcolsep}{4.5pt}
\begin{tabular}{llrrrrrl}
\arrayrulecolor{cdInk}\toprule
\rowcolor{cdMist}Domain & Source & Hard & Parents & \szB\,2B & \szC\,4B & $\Delta$ (pp) & 95\% CI \\
\arrayrulecolor{cdLine}\midrule
Legal & \texttt{explicit\_legal\_rules} & 1,498 & 1,049 & 53.27 & 75.57 & \textcolor{cdTealDark}{\textbf{+22.30}} & [+19.43, +25.20] \\
Medical & \texttt{empec} & 2,062 & 2,061 & 58.39 & 73.76 & \textcolor{cdTealDark}{\textbf{+15.37}} & [+13.14, +17.64] \\
Medical & \texttt{medqa\_taiwan} & 1,316 & 1,316 & 66.87 & 82.22 & \textcolor{cdTealDark}{\textbf{+15.35}} & [+12.77, +18.01] \\
Finance & \texttt{explicit\_finance\_rules} & 750 & 750 & 78.27 & 91.87 & \textcolor{cdTealDark}{\textbf{+13.60}} & [+10.67, +16.53] \\
General & \texttt{cmrc2019} & 3,048 & 299 & 73.36 & 85.33 & \textcolor{cdTealDark}{\textbf{+11.98}} & [+10.21, +13.78] \\
Medical & \texttt{medqa\_mainland} & 2,438 & 2,411 & 77.36 & 88.97 & \textcolor{cdTealDark}{\textbf{+11.61}} & [+10.01, +13.27] \\
\arrayrulecolor{cdLine}\midrule
General & \texttt{mep3m\_product\_catalog} & 3,151 & 3,151 & 70.96 & 64.58 & \textcolor{cdCoral}{\textbf{-6.38}} & [-7.62, -5.11] \\
\arrayrulecolor{cdInk}\bottomrule
\end{tabular}
\par\smallskip\begin{minipage}{\linewidth}\tablenote{55 sources have at least 200 hard targets; 30 improve and 1 decline with intervals excluding zero (parent-clustered paired bootstrap within each source, no multiplicity correction).}\end{minipage}
}

\end{table}
Source-level tests in Table~\ref{tab:sources} cover the 55 sources with at least 200 hard targets. From 2B to 4B, 30 sources improve with intervals excluding zero and one declines. The largest gains come from rule-application and examination-style sources: explicit legal rules (+22.3 points), the EMPEC and Taiwan/mainland MedQA medical examinations (+15.4, +15.4, +11.6), explicit finance rules (+13.6), and CMRC 2019 reading comprehension (+12.0). The single regression is the product-catalog source ($-$6.38 points; source ECE 22.6\%), the same source that is unstable during both 2B and 4B training. From 0.8B to 2B, 19 sources improve and six decline significantly. The declines are smaller, between 0.9 and 3.8 points, in GMD12, explicit finance rules, LCR-CN, CBLUE-QIC, WebMedQA, and LCMDC. Without a multiplicity correction, roughly three of 55 nominal 95\% tests would be expected to flip by chance, so individual small regressions should be treated as leads rather than findings.

\begin{finding}
The 4B gains are broad (30 of 55 sources) and concentrated in rule application and professional examinations, while one product-catalog source regresses and stays poorly calibrated. Source-level checks are needed before deploying any member on a single task family.
\end{finding}

\subsection{Calibration and Temperature}
\label{sec:calibration}
\begin{table}[t]
\caption{\textbf{Temperature-only ablation on fixed logits}: each final checkpoint at $T=1$ versus the same logits divided by its validation-fitted temperature.}
\label{tab:temperature}
\centering
{\small
\setlength{\tabcolsep}{4pt}
\begin{tabular}{lrrrrrrr}
\arrayrulecolor{cdInk}\toprule
\rowcolor{cdMist}Size & $T$ & Pooled ECE & General & Medical & Legal & Finance & Pooled NLL \\
\arrayrulecolor{cdLine}\midrule
\szA\,0.8B & 1.1081 & 1.07 \better 0.28 & 3.03 \better 1.69 & 0.17 \worse 1.13 & 5.70 \better 4.03 & 1.50 \worse 3.06 & 0.4323 \better{} 0.4312 \\
\szB\,2B & 1.0749 & 1.49 \better 0.64 & 3.01 \better 1.93 & 0.73 \better 0.24 & 3.78 \better 3.28 & 1.98 \better 1.86 & 0.4191 \better{} 0.4164 \\
\szC\,4B & 1.0248 & 1.26 \better 0.98 & 3.25 \better 2.89 & 0.35 \better 0.21 & 1.65 \better 1.33 & 1.40 \worse 1.77 & 0.3714 \better{} 0.3706 \\
\arrayrulecolor{cdInk}\bottomrule
\end{tabular}
\par\smallskip\begin{minipage}{\linewidth}\tablenote{Same logits before and after division by the validation-fitted $T$, so predictions and accuracy are identical within each row. ECE in \%. \better{}: improves; \worse{}: worsens.}\end{minipage}
}

\end{table}
\begin{figure}[t]
\centering
\includegraphics[width=\textwidth]{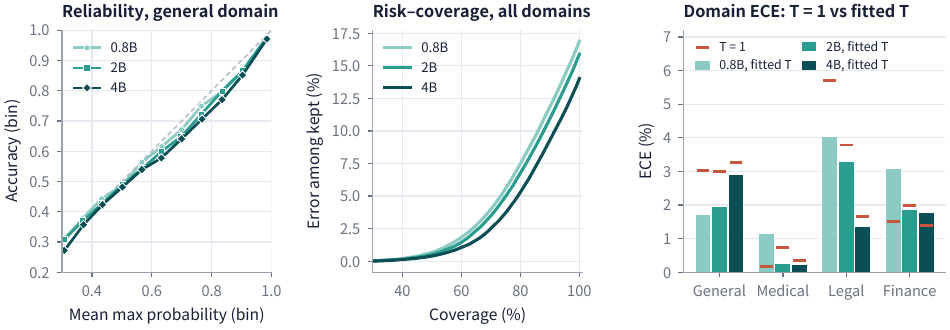}
\caption{\textbf{Calibration and ranking.} Left: general-domain reliability of the released models (bins with fewer than 200 decisions omitted). Middle: risk--coverage curves over all hard targets. Right: domain ECE of the final checkpoints at $T=1$ (red marks) and after dividing the same logits by the fitted temperature (bars).}
\label{fig:calibration}
\end{figure}
The fitted temperature falls with size, from 1.108 to 1.075 to 1.025: on the calibration partition, the larger members are already closer to calibrated at $T=1$. Table~\ref{tab:temperature} isolates the effect of temperature on fixed logits. In every member it lowers pooled ECE (1.07\% to 0.28\%, 1.49\% to 0.64\%, and 1.26\% to 0.98\%) and pooled NLL. Its domain effects are mixed. For 0.8B, it improves general and legal ECE but raises medical ECE from 0.17\% to 1.13\% and financial ECE from 1.50\% to 3.06\%. For 4B, financial ECE rises from 1.40\% to 1.77\%. Only for 2B does every domain improve.

The sign of the gap between mean confidence and accuracy explains why a pooled metric hides this (Figure~\ref{fig:calibration}, right). At $T=1$ the 0.8B checkpoint is overconfident by 3.0 points on general decisions and calibrated on average on medical ones (0.0). After one global softening, general overconfidence falls to 1.4 points, but medical and financial decisions become underconfident by 1.1 and 2.0 points. Errors of opposite sign then cancel inside the shared confidence bins, so a pooled ECE of 0.28\% coexists with 1.1--4.0\% ECE in every domain. The 4B financial subset is already slightly underconfident at $T=1$ ($-$0.4 points), so softening it further raises its ECE. Temperature also slightly worsens expected-score MAE for every member (by 0.001--0.006), consistent with softening pulling expected levels toward the middle of the scale.

Merging contributes separately. On fixed 4B logits, temperature lowers legal ECE from 1.65\% to 1.33\%, yet the released merged model has 2.02\% legal ECE at the same temperature. The difference comes from the merged artifact rather than the temperature, on a subset of 4,300 decisions whose ECE interval spans roughly 1.7--3.5\%. The general-domain reliability curves show the larger models below the diagonal between 0.6 and 0.85 confidence. On fixed 4B logits at the fitted temperature, general-domain ECE is 2.89\% and mean confidence exceeds accuracy by 2.89 points, so this error is almost entirely overconfidence.

\begin{finding}
A single global temperature is a pooled fix: it can trade one domain's overconfidence against another's underconfidence. Report per-domain ECE alongside pooled ECE, and fit per-domain temperatures on validation data where domains are known at serving time.
\end{finding}

\subsection{Selective Prediction}
\label{sec:selective}
\begin{table}[t]
\caption{\textbf{Selective prediction} with the released models, ranking hard decisions by maximum candidate probability.}
\label{tab:selective}
\centering
{\small
\setlength{\tabcolsep}{4pt}
\begin{tabular}{lrrrrrrrrrr}
\arrayrulecolor{cdInk}\toprule
\rowcolor{cdMist}& \multicolumn{4}{c}{Accuracy (\%) at coverage} & \multicolumn{2}{c}{All domains (\%)} & \multicolumn{4}{c}{AURC by domain (\%)} \\
\rowcolor{cdMist}Size & 50\% & 80\% & 90\% & 100\% & AURC & E-AURC & General & Medical & Legal & Finance \\
\arrayrulecolor{cdLine}\midrule
\szA\,0.8B & 99.32 & 92.53 & 88.17 & 83.11 & 3.40 & 1.89 & 9.79 & 1.51 & 13.28 & 9.29 \\
\szB\,2B & 99.47 & 93.24 & 89.07 & 84.11 & 3.08 & 1.75 & 8.59 & 1.42 & 13.81 & 8.56 \\
\szC\,4B & 99.61 & 94.72 & 90.71 & 85.97 & 2.52 & 1.48 & 7.70 & 1.07 & 7.59 & 4.33 \\
\arrayrulecolor{cdInk}\bottomrule
\end{tabular}
\par\smallskip\begin{minipage}{\linewidth}\tablenote{Hard decisions ranked by maximum candidate probability. AURC: area under the risk--coverage curve; E-AURC subtracts the AURC of an oracle ranking with the same accuracy. Lower is better for both.}\end{minipage}
}

\end{table}
Calibration error and ranking quality are different properties. Table~\ref{tab:selective} and the middle panel of Figure~\ref{fig:calibration} evaluate whether maximum probability ranks errors. Abstaining on the least confident 20\% of hard decisions raises accuracy to 92.53\%, 93.24\%, and 94.72\% for 0.8B, 2B, and 4B; at 50\% coverage all three exceed 99.3\%. AURC falls from 3.40\% to 3.08\% to 2.52\%. E-AURC removes the part of AURC explained by accuracy alone, and it also falls (1.89\% to 1.75\% to 1.48\%), so the larger members rank their own errors better and are not merely more accurate. The gap is largest in law, where AURC drops from 13.81\% for 2B to 7.59\% for 4B. Medical decisions are easiest to rank (AURC 1.07--1.51\%). Law has the highest AURC for 0.8B and 2B; for 4B the general domain does (7.70\% vs.\ 7.59\%).

\begin{finding}
Despite its higher ECE, the 4B model is the best of the three for threshold-based deferral. Deferral thresholds for an application should still be set on its own labeled data.
\end{finding}

\renewcommand{\sectioncolor}{cdIndigo}
\section{Results: Unified Models}
\label{sec:unified-results}
The unified models raise four questions. Do they keep, or lose, the decision ability of decision-only models (Section~\ref{sec:unified-decision})? How well do they extract spans (Section~\ref{sec:extraction})? What does keeping decisions cost the extraction readout, compared with training on extraction alone (Section~\ref{sec:retention-cost})? And how much of that cost is a decoding choice, and where do the errors fall (Section~\ref{sec:decoding})? All differences below are paired and use the parent-clustered intervals of Section~\ref{sec:analysis-protocol}, which describe test-set sampling for fixed models and not training variability.

\subsection{Decisions Are Retained}
\label{sec:unified-decision}
\begin{figure}[t]
\centering
\includegraphics[width=\textwidth]{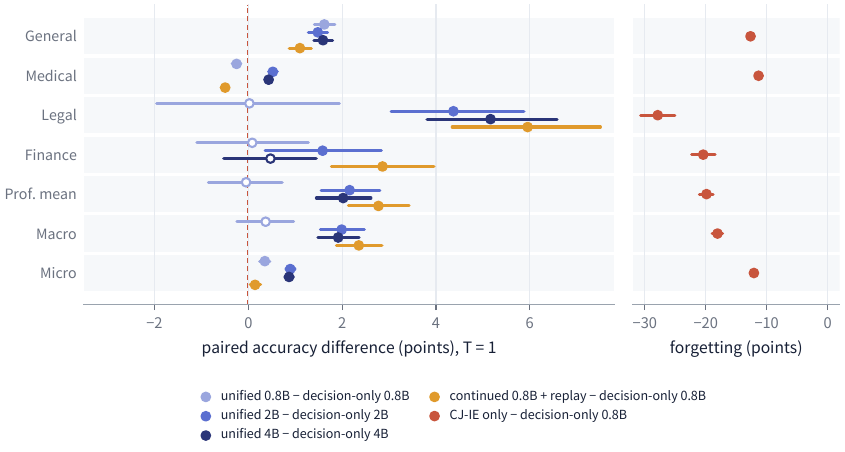}
\caption{\textbf{Paired accuracy differences on CJ-Bench} (final checkpoints at $T=1$; 95\% parent-clustered intervals; filled: the interval excludes zero). Left: the three unified models and the replay-and-KL run against the decision-only model of their size. Right, on its own scale: continuing the 0.8B decision model on CJ-IE alone. Appendix Table~\ref{tab:unified-paired} lists the values.}
\label{fig:forest-unified}
\end{figure}
Figure~\ref{fig:forest-unified} compares each unified model with the decision-only model of the same size. The unified 0.8B model reaches 83.46\% against 83.11\%, a difference of $+0.35$ points (95\% CI [$+0.24$, $+0.46$]). The gain is in the general domain ($+1.62$ [$+1.41$, $+1.84$]); medicine is slightly lower ($-0.25$ [$-0.34$, $-0.16$]), and law, finance, and the professional mean are unresolved ($+0.02$ [$-1.95$, $+1.94$], $+0.08$ [$-1.10$, $+1.28$], and $-0.05$ [$-0.85$, $+0.72$]). The unified 2B model improves in every domain: general $+1.48$ [$+1.28$, $+1.69$], medical $+0.52$ [$+0.43$, $+0.62$], legal $+4.37$ [$+3.04$, $+5.88$], finance $+1.58$ [$+0.36$, $+2.83$], and micro accuracy $+0.89$ [$+0.81$, $+0.99$]. The unified 4B model reaches 86.86\% against 85.99\% for the decision-only 4B checkpoint ($+0.87$ [$+0.78$, $+0.95$]) and improves general ($+1.59$ [$+1.40$, $+1.80$]), medical ($+0.43$ [$+0.35$, $+0.51$]), and legal accuracy ($+5.16$ [$+3.81$, $+6.58$]); finance ($+0.47$ [$-0.52$, $+1.45$]) is unresolved, while the professional mean gains $+2.02$ [$+1.45$, $+2.61$] and macro accuracy $+1.92$ [$+1.48$, $+2.36$]. The unified 4B model is the most accurate released model in every domain and $1.85$ points [$1.75$, $1.96$] above the unified 2B model, with gains of $9.02$ points in law and $6.39$ in finance. The unified 2B model stays $0.99$ points [$0.87$, $1.10$] below the decision-only 4B model, and $3.65$ points [$3.00$, $4.32$] below it on the professional mean (Figure~\ref{fig:domains}).

\begin{caveat}[One source carries much of the gain]
The 3,151-decision product-catalog source (1.0\% of hard targets) is answered correctly 98.25\%, 98.13\%, and 98.64\% of the time by the unified 0.8B, 2B, and 4B models, against 54.27\%, 70.99\%, and 64.58\% for the decision-only models of those sizes; its instability during decision-only training was noted in Section~\ref{sec:results}. With this source set aside, the unified 0.8B model is $-0.10$ points [$-0.22$, $+0.01$] from the decision-only 0.8B model, and its general-domain gain shrinks to $+0.20$ [$-0.06$, $+0.48$]; the unified 2B and 4B models keep $+0.62$ [$+0.53$, $+0.72$] and $+0.52$ [$+0.44$, $+0.61$]. The CJ-IE mixture contains product-title NER and typing tasks that may supply the missing signal, but we ran no ablation and do not know.
\end{caveat}
Source-level tests show the same picture. Against the decision-only 2B model, the unified 2B model improves 23 of 55 sources with intervals excluding zero and declines in one (T2Ranking, $-0.67$ points). Against the decision-only 4B model, the unified 4B model improves 20 and declines in one (CMedQA2, $-0.05$), with the largest gains in the product catalog ($+34.1$), LCR-CN ($+9.4$), and the explicit legal rules ($+7.7$); against the unified 2B model it improves 31 sources and declines in none. Against the decision-only 0.8B model, the unified 0.8B model improves nine and declines in ten, led by LCR-CN ($-6.3$), the explicit finance rules ($-2.8$), and MedQA-mainland ($-2.6$). Top-1 answers differ for 8.8\% (0.8B), 8.3\% (2B), and 6.6\% (4B) of decisions, with 10,400, 11,044, and 9,433 hard targets fixed and 9,326, 8,317, and 6,785 broken, a churn comparable to the difference between sizes.

\begin{table}[t]
\caption{\textbf{Temperature on the unified models' fixed logits.} Each final checkpoint at $T=1$ against the same logits divided by its fitted temperature; ECE in \%.}
\label{tab:unified-temperature}
\centering
{\footnotesize
\setlength{\tabcolsep}{3.6pt}
\begin{tabular}{lrrrrrrr}
\arrayrulecolor{cdInk}\toprule
\rowcolor{cdMist}Model & $T$ & Pooled ECE & General & Medical & Legal & Finance & Pooled NLL \\
\arrayrulecolor{cdLine}\midrule
\uniA\,0.8B & 1.1326 & 1.12 \better 0.43 & 2.84 \better 1.02 & 0.36 \worse 1.02 & 2.32 \better 2.29 & 1.13 \worse 2.37 & 0.4243 \better{} 0.4226 \\
\uniB\,2B & 1.0617 & 0.92 \better 0.26 & 2.46 \better 1.67 & 0.25 \worse 0.49 & 3.29 \better 2.58 & 2.24 \better 1.76 & 0.3881 \better{} 0.3871 \\
\uniC\,4B & 1.0986 & 0.97 \better 0.14 & 2.84 \better 1.56 & 0.16 \worse 0.90 & 2.45 \better 1.69 & 1.51 \better 1.44 & 0.3483 \better{} 0.3468 \\
\arrayrulecolor{cdInk}\bottomrule
\end{tabular}
\par\smallskip\begin{minipage}{\linewidth}\tablenote{Same logits before and after division by the validation-fitted $T$; ECE in \%. \better{}: improves; \worse{}: worsens.}\end{minipage}
}

\end{table}
The unified models are calibrated by the same recipe as the decision-only models. Their fitted temperatures are 1.133, 1.062, and 1.099, and pooled ECE falls from 1.12\% to 0.43\% (0.8B), from 0.92\% to 0.26\% (2B), and from 0.97\% to 0.14\% (4B), with pooled NLL also lower (Table~\ref{tab:unified-temperature}); the unified 4B model has the lowest pooled ECE of the six released decision models. As for the decision-only models (Section~\ref{sec:calibration}), the pooled gain hides domain effects of both signs: medical ECE rises (0.36\% to 1.02\% at 0.8B, 0.25\% to 0.49\% at 2B, 0.16\% to 0.90\% at 4B) and financial ECE rises at 0.8B (1.13\% to 2.37\%). Ranking by confidence is also at least as good as for the decision-only models: E-AURC is 1.77\%, 1.58\%, and 1.35\% against 1.89\%, 1.75\%, and 1.48\% (Appendix Table~\ref{tab:unified-selective}).

\begin{ufinding}
Training one weight file on both tasks did not cost decision accuracy at any size, and its measured benefit is small: at 0.8B it is confined to one source, at 2B and 4B it is about $+0.6$ and $+0.5$ points without that source. The pairs differ in more than the extraction task (at 0.8B the decision-only model was trained with CE alone and the unified model with CE+RLCD, plus token budgets, update counts, and the selection protocol), so these are recipe comparisons, not estimates of an effect of multitask training.
\end{ufinding}

\subsection{Typed Span Extraction on CJ-IE}
\label{sec:extraction}
\begin{table}[t]
\caption{\textbf{Extraction on the CJ-IE test split.} Exact-match micro and macro scores of the five joint-training runs over 232,291 gold spans in 35 sources, at the model-card operating point (left) and at the release default (right); Table~\ref{tab:recipe-results} gives their CJ-Bench accuracy.}
\label{tab:extraction}
\centering
{\small
\setlength{\tabcolsep}{2.7pt}
\begin{tabular}{lrrrrrrrrrrr}
\arrayrulecolor{cdInk}\toprule
\rowcolor{cdMist}& \multicolumn{3}{c}{Micro at $s=\log 5$ (\%)} & \multicolumn{4}{c}{Macro-over-sources F1 (\%)} & \multicolumn{3}{c}{Micro at $s_d=0.75$ (\%)} & Entity \\
\rowcolor{cdMist}Run & P & R & F1 & All 35 & NER & Rel. & Event & P & R & F1 & ECE (\%) \\
\arrayrulecolor{cdLine}\midrule
\runA\,CJ-IE only & 79.1 & 69.2 & \textbf{73.9} & 74.9 & 75.9 & 64.8 & 69.1 & 72.1 & 74.6 & \textbf{73.3} & 1.7 \\
\runB\,replay + KL & 75.4 & 57.3 & \textbf{65.1} & 65.0 & 66.6 & 48.7 & 57.8 & 68.2 & 65.5 & \textbf{66.8} & 1.2 \\
\uniA\,unified 0.8B & 73.3 & 55.5 & \textbf{63.1} & 62.5 & 64.0 & 47.2 & 54.5 & 66.7 & 62.8 & \textbf{64.7} & 4.1 \\
\uniB\,unified 2B & 75.4 & 58.6 & \textbf{66.0} & 66.3 & 67.4 & 55.4 & 59.2 & 68.6 & 66.0 & \textbf{67.3} & 1.8 \\
\uniC\,unified 4B & 78.9 & 61.1 & \textbf{68.9} & 69.8 & 71.0 & 56.9 & 64.0 & 72.5 & 68.7 & \textbf{70.6} & 3.2 \\
\arrayrulecolor{cdInk}\bottomrule
\end{tabular}
\par\smallskip\begin{minipage}{\linewidth}\tablenote{Exact-match scores from the training runs at the operating point of the model cards (none shift $\log 5=1.61$, which keeps probabilities calibrated), and re-decoded from the stored span utilities with the release default of 2026-10-31 (decision shift $s_d=0.75$ decides which spans are entities, probabilities still use $\log 5$). Entity ECE uses 10 bins over predicted spans at $\log 5$ (re-decoded).}\end{minipage}
}

\end{table}
On the 232,291 gold spans of the CJ-IE test split, the unified 0.8B model reaches micro F1 of 63.1 (precision 73.3, recall 55.5), the unified 2B model 66.0 (75.4, 58.6), and the unified 4B model 68.9 (78.9, 61.1) at the model-card operating point (Table~\ref{tab:extraction}). Each step up in size is resolved: 2B over 0.8B by $+2.84$ points [$+2.65$, $+3.05$] (macro-over-sources F1 $+3.84$ [$+3.55$, $+4.15$]) and 4B over 2B by $+2.88$ [$+2.69$, $+3.05$] (macro $+3.42$ [$+3.13$, $+3.71$]). At the release default, the three models reach 64.7, 67.3, and 70.6, gains of 1.6, 1.4, and 1.8 points. All models are conservative at the model-card point, with precision above recall by 17.8, 16.8, and 17.8 points; at the release default the margin is 3.8, 2.6, and 3.8. By family, macro F1 is 64.0, 67.4, and 71.0 for NER, 47.2, 55.4, and 56.9 for relation objects, and 54.5, 59.2, and 64.0 for event arguments. All 35 sources improve from 0.8B to 4B, and 33 of the 35 improve from 2B to 4B (sign test $p=4\times10^{-8}$); the largest gains over 2B are Weibo ($+11.0$), CLUENER ($+8.2$), CAIL2022 ($+7.6$), Wanchuang ($+7.1$), and MultiCoNER22 ($+7.0$), and the two that do not improve are CMeIE ($-1.2$) and IMCS21 ($-0.5$).

\begin{figure}[t]
\centering
\includegraphics[width=\textwidth]{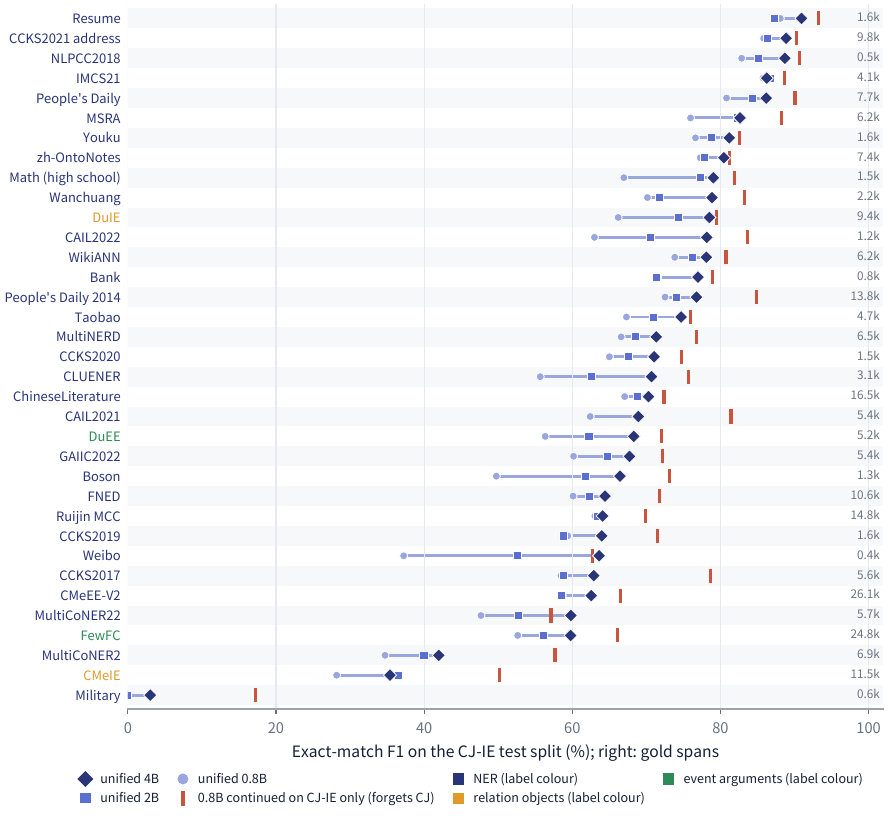}
\caption{\textbf{Extraction F1 for every source of the CJ-IE test split,} sorted by the unified 4B score. Red ticks mark the 0.8B model continued on CJ-IE alone, the reference that forgets CJ-Bench (Section~\ref{sec:retention-cost}); it is above the unified 0.8B and 2B models on every source and above the unified 4B model on 33 of 35. Label colours: indigo NER, amber relation objects, green event arguments.}
\label{fig:sources}
\end{figure}
Figure~\ref{fig:sources} shows the spread. The unified 4B model is strongest where entity classes are few and boundaries are conventional: resumes (90.9), the 16 address elements of CCKS2021 (88.9), NLPCC2018 (88.7), medical dialogue (IMCS21, 86.3), and People's Daily (86.2). It is weakest on a source whose four types share one name and definition (Military, 3.0, recall 1.5, against 0.0 for the smaller models), on medical relation objects (CMeIE, 35.4, recall 24.4), on the 33-type MultiCoNER2 (42.0), and on social-media text (Weibo, 63.6, precision 77.1, recall 54.1). Of the 127 types with at least 400 gold spans, 41 reach F1 of 80 or more (35 at 2B), 60 lie between 50 and 80, and 26 are below 50; no type scores zero, where four did at 2B. The best are the address elements province (99.2), city (98.0), and house number (97.1).

\begin{table}[t]
\caption{\textbf{Extraction-related decisions on the CJ-IE test split} (hard-target accuracy, \%).}
\label{tab:ie-decision}
\centering
{\footnotesize
\setlength{\tabcolsep}{3.4pt}
\begin{tabular}{llrrrrrr}
\arrayrulecolor{cdInk}\toprule
\rowcolor{cdMist}Task & Question & Items & \runA\,CJ-IE only & \runB\,replay+KL & \uniA\,uni. 0.8B & \uniB\,uni. 2B & \uniC\,uni. 4B \\
\arrayrulecolor{cdLine}\midrule
\texttt{ner\_typing} & type of a marked mention & 20,312 & 93.9 & 93.2 & 92.7 & 93.6 & 94.4 \\
\texttt{re\_pair} & relation of an entity pair & 40,417 & 89.8 & 89.1 & 88.8 & 89.6 & 90.5 \\
\texttt{ee\_type} & event type X described? & 13,604 & 94.9 & 94.1 & 92.8 & 93.7 & 92.7 \\
\texttt{ee\_category} & category of the event & 5,863 & 83.7 & 80.9 & 84.9 & 83.7 & 85.4 \\
\texttt{doc\_field} & mention for an event field & 2,756 & 96.8 & 96.6 & 97.1 & 97.5 & 97.8 \\
\texttt{doc\_event} & event X disclosed? & 1,942 & 99.6 & 99.7 & 99.7 & 99.7 & 99.8 \\
\arrayrulecolor{cdLine}\midrule
\textbf{all} &  & \textbf{84,894} & \textbf{91.6} & \textbf{90.8} & \textbf{90.6} & \textbf{91.3} & \textbf{91.9} \\
\arrayrulecolor{cdInk}\bottomrule
\end{tabular}
\par\smallskip\begin{minipage}{\linewidth}\tablenote{Hard-target accuracy (\%) on the CJ-IE test split (594 soft targets excluded from accuracy). Task views are described in Appendix~\ref{app:cjie}.}\end{minipage}
}

\end{table}
The letter readout handles the extraction-related decisions of CJ-IE (Table~\ref{tab:ie-decision}). The unified 4B model scores 91.9\% over 84,894 items, against 91.3\% for 2B and 90.6\% for 0.8B, and is above the CJ-IE-only run (91.6\%). Entity typing (94.4\% at 4B), relation classification of a given pair (90.5\%), and the document-level tasks (97.8\% and 99.8\%) improve with size; event-type classification is the one task where 4B does not improve (92.7\% against 93.7\% for 2B and 92.8\% for 0.8B), and event-category classification is not monotonic (84.9\%, 83.7\%, 85.4\%). Span probabilities are reasonably calibrated at the model-card operating point: entity ECE is 4.1\% (0.8B), 1.8\% (2B), and 3.2\% (4B). The top bin is overconfident for the 0.8B model (confidence 0.96 against 0.91 accuracy) and the 2B model (0.96 against 0.94), and slightly underconfident for the 4B model (0.95 against 0.96).

\begin{ufinding}
A 0.8B, 2B, or 4B model extracts typed spans from one weight file at 63 to 69 F1 (65 to 71 at the release default), with strong results on conventional NER and weak ones on medical relations, social media, and sources whose types lack definitions. Every step up in size improves F1 by about three points and nearly every source. These are scores on an internal, standardized split, not leaderboard numbers (Section~\ref{sec:setup}), and no zero-shot claim is made.
\end{ufinding}

\subsection{The Price of Keeping Decisions}
\label{sec:retention-cost}
\begin{table}[t]
\caption{\textbf{Retention against extraction.} The decision-only models (top) have no span head. Below, the five joint-training runs of Table~\ref{tab:recipes}: CJ-Bench accuracy at $T=1$ and CJ-IE test scores at the model-card operating point.}
\label{tab:recipe-results}
\centering
{\small
\setlength{\tabcolsep}{3.4pt}
\begin{tabular}{lrrrrrrrr}
\arrayrulecolor{cdInk}\toprule
\rowcolor{cdMist}& \multicolumn{3}{c}{CJ-Bench ACC (\%), $T=1$} & \multicolumn{5}{c}{CJ-IE test (\%)} \\
\rowcolor{cdMist}Run & All & Legal & Finance & Dec.\ ACC & P & R & F1 & Macro F1 \\
\arrayrulecolor{cdLine}\midrule
\szA\,decision-only 0.8B & 83.11 & 67.86 & 74.22 & -- & -- & -- & -- & -- \\
\szB\,decision-only 2B & 84.11 & 69.05 & 76.03 & -- & -- & -- & -- & -- \\
\szC\,decision-only 4B & 85.99 & 77.28 & 83.53 & -- & -- & -- & -- & -- \\
\arrayrulecolor{cdLine}\midrule
\runA\,CJ-IE only & 71.02 & 39.98 & 53.81 & 91.6 & 79.1 & 69.2 & 73.9 & 74.9 \\
\runB\,replay + KL & 83.25 & 73.81 & 77.08 & 90.8 & 75.4 & 57.3 & 65.1 & 65.0 \\
\uniA\,unified 0.8B & 83.46 & 67.88 & 74.31 & 90.6 & 73.3 & 55.5 & 63.1 & 62.5 \\
\uniB\,unified 2B & 85.00 & 73.42 & 77.61 & 91.3 & 75.4 & 58.6 & 66.0 & 66.3 \\
\uniC\,unified 4B & 86.86 & 82.44 & 84.00 & 91.9 & 78.9 & 61.1 & 68.9 & 69.8 \\
\arrayrulecolor{cdInk}\bottomrule
\end{tabular}
\par\smallskip\begin{minipage}{\linewidth}\tablenote{CJ-IE test: ``Dec.\ ACC'' is accuracy on the 84,894 extraction-related decisions; P, R and F1 are exact-match micro scores over 232,291 gold spans in 35 sources; macro F1 averages the 35 per-source F1 values. Extraction values are those stored by each training run.}\end{minipage}
}

\end{table}
\begin{figure*}[t]
\centering
\includegraphics[width=\textwidth]{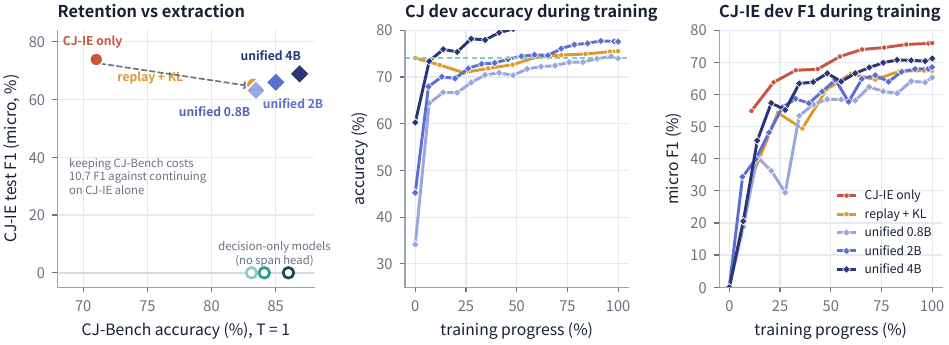}
\caption{\textbf{Retention against extraction.} Left: CJ-Bench accuracy against CJ-IE test F1 for the five runs; the decision-only models (open circles) have no span head. Middle and right: CJ dev accuracy and CJ-IE dev F1 during training (dev data only; the dashed line marks the decision model before training, and the CJ-IE-only run has no CJ dev curve).}
\label{fig:tradeoff}
\end{figure*}
Continuing the decision 0.8B model on CJ-IE alone, with no CJ data, extracts best of the five runs (micro F1 73.9; Figure~\ref{fig:tradeoff}) and forgets the decisions it started with. CJ-Bench accuracy falls from 83.11\% to 71.02\%, a change of $-12.09$ points [$-12.32$, $-11.86$]. Law falls $27.88$ points [$-30.68$, $-25.08$] to 39.98\% and finance $20.42$ points [$-22.32$, $-18.48$]; 53 of the 55 sources decline significantly (only the product-catalog source rises, by 21.2 points); pooled ECE rises from 1.07\% to 2.84\%, and legal ECE by 18.6 points (Figure~\ref{fig:forest-unified}, right). Judged on the extraction dev split alone, this is the best of the five runs.

Replay with a KL anchor and RLCD removes the forgetting: the replay-and-KL run reaches 83.25\% on CJ-Bench, $+0.14$ points [$+0.04$, $+0.26$] from the decision-only model, with law $+5.95$ [$+4.35$, $+7.51$] and finance $+2.86$ [$+1.77$, $+3.96$] and medicine $-0.49$ [$-0.58$, $-0.41$]. The unified 0.8B model, trained from the raw backbone on all of CJ-Dataset, reaches 83.46\% but gives back $5.93$ points [$4.55$, $7.41$] in law and $2.78$ [$1.60$, $3.96$] in finance relative to the replay-and-KL run. That run has seen each legal and finance decision twice, once while training the decision model and once as replay; we suspect this, but the runs do not test it. The unified 2B and 4B models, trained in the same way from larger raw backbones, are $4.37$ and $5.16$ points above the decision-only models of their size in law.

The extraction side pays. Relative to the CJ-IE-only run, micro F1 is lower by 8.8 points for replay-and-KL, 10.7 for the unified 0.8B model, 7.9 for the unified 2B model, and 5.0 for the unified 4B model. The gap shrinks with size, but the reference is a 0.8B model: no extraction-only continuation exists at 2B or 4B, so the 4B number compares a larger model that keeps its decisions with a smaller one that does not. Each difference is resolved in the re-decoded paired comparison (73.3 against 63.1: $+10.2$ [$+10.0$, $+10.5$]; against 2B $+7.4$ [$+7.2$, $+7.6$]; against 4B $+4.5$ [$+4.3$, $+4.7$]). The gap is not confined to a few sources: the CJ-IE-only run beats the unified 0.8B and 2B models on all 35 sources (sign test $p=6\times10^{-11}$) and the unified 4B model on 33 (MultiCoNER22 and Weibo are the exceptions, $p=4\times10^{-8}$). It is mostly recall (69.2 against 55.5, 58.6, and 61.1) rather than precision (79.1 against 73.3, 75.4, and 78.9), and at 4B it is recall alone. It is also much smaller for the extraction-related decisions, where the unified 4B model (91.9\%) is above the CJ-IE-only run (91.6\%). At the release default the same paired comparisons give gaps of 8.6, 6.0, and 2.7 points ([$+8.4$, $+8.8$], [$+5.8$, $+6.2$], and [$+2.5$, $+2.9$]).

\begin{caveat}[What this comparison does not isolate]
The five runs differ in starting point (decision model or raw backbone), in CJ data (none, 348,000 decisions, or all 13.05 million), in objective (CE or CE+RLCD), and in how updates are composed (Table~\ref{tab:recipes}). Spans are 65\% of the micro-batches of the CJ-IE-only run and about 8\% of those of the unified runs, so an update averages 21 span rows in the first and 40 in the unified runs, and the runs take 27,514 updates (CJ-IE-only) and 14,504 to 15,304 (unified). The replay-and-KL run, where spans are 54\% of the micro-batches, still loses 8.8 points, so span share alone does not explain the cost. The dev micro F1 of the unified 0.8B and 2B models was still rising at the end of the epoch (they gained 3.0 and 3.7 points since step 10,000), while the 4B model changed by only 0.4 points after step 12,000 (70.8 to 71.2), so a longer schedule might shrink the gap at the smaller sizes; we ran no second epoch and no single-factor ablation.
\end{caveat}

\begin{ufinding}
Keeping CJ-Bench accuracy and extracting spans with one weight file cost 5 to 11 extraction-F1 points relative to an extraction-only continuation that forgets CJ-Bench by 12 points, and the cost falls as the unified model grows (10.7, 7.9, and 5.0 points at 0.8B, 2B, and 4B). A model card that reports the unified F1 without this reference understates what better recipes might recover, and a larger unified model recovers part of it by size alone.
\end{ufinding}

\subsection{Decoding Choices and Where Errors Fall}
\label{sec:decoding}
\begin{figure*}[t]
\centering
\includegraphics[width=\textwidth]{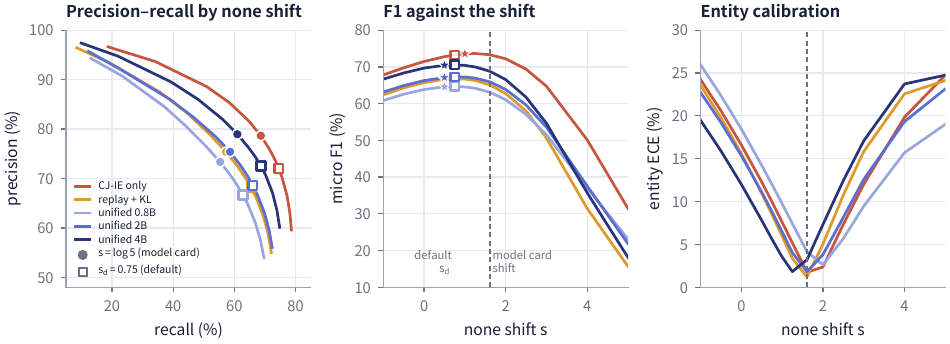}
\caption{\textbf{The none shift.} Left: precision against recall as the none shift $s$ varies (dots: model-card shift 1.61; squares: release default $s_d=0.75$). Middle: micro F1 against $s$ (stars: shift selected on the dev split; squares: release default). Right: entity ECE against $s$. The CJ-IE-only curve lies above the others at every operating point.}
\label{fig:operating}
\end{figure*}
\begin{table}[t]
\caption{\textbf{Decoding variants} (micro F1 \%, re-decoded from stored span utilities).}
\label{tab:decoding}
\centering
{\small
\setlength{\tabcolsep}{3.2pt}
\begin{tabular}{lrrrrrr}
\arrayrulecolor{cdInk}\toprule
\rowcolor{cdMist}Run & Model card ($s=1.61$) & Default ($s_d=0.75$) & Dev $s^{*}$ ($s^{*}$) & Nested & $p\geq0.5$ & Best on test ($s$) \\
\arrayrulecolor{cdLine}\midrule
\runA\,CJ-IE only & 73.3 {\color{cdGray}\scriptsize[73.1, 73.5]} & 73.3 {\color{cdGray}\scriptsize[73.1, 73.5]} & 73.6 (1.0) & 73.5 & 72.9 & 73.6 (1.25) \\
\runB\,replay + KL & 65.1 {\color{cdGray}\scriptsize[64.9, 65.4]} & 66.8 {\color{cdGray}\scriptsize[66.6, 67.0]} & 66.6 (0.5) & 65.3 & 64.0 & 66.8 (0.75) \\
\uniA\,unified 0.8B & 63.1 {\color{cdGray}\scriptsize[62.8, 63.4]} & 64.7 {\color{cdGray}\scriptsize[64.4, 65.0]} & 64.6 (0.5) & 63.2 & 62.2 & 64.7 (0.75) \\
\uniB\,unified 2B & 65.9 {\color{cdGray}\scriptsize[65.7, 66.2]} & 67.3 {\color{cdGray}\scriptsize[67.1, 67.5]} & 67.1 (0.5) & 66.1 & 65.0 & 67.3 (0.75) \\
\uniC\,unified 4B & 68.8 {\color{cdGray}\scriptsize[68.6, 69.1]} & 70.6 {\color{cdGray}\scriptsize[70.4, 70.8]} & 70.5 (0.5) & 69.0 & 67.6 & 70.6 (0.75) \\
\arrayrulecolor{cdInk}\bottomrule
\end{tabular}
\par\smallskip\begin{minipage}{\linewidth}\tablenote{Micro F1 (\%), re-decoded from stored span utilities. Brackets: 95\% stratified cluster-bootstrap intervals. ``Model card'' is the operating point of the training-time evaluation, ``Default'' the release configuration since 2026-10-31 (probabilities still at $\log 5$). ``Dev $s^{*}$'' selects the none shift on the dev split (macro F1); ``Nested'' keeps overlapping spans; the last column is an oracle that peeks at the test split.}\end{minipage}
}

\end{table}
Lowering the none shift trades precision for recall. The model-card shift $\log 5=1.61$ undoes the 0.2 weight on none candidates during training and leaves all five runs precision-heavy. The release default $s_d=0.75$, which decides the entities while the probabilities keep $\log 5$, raises test F1 by 1.7, 1.6, 1.4, and 1.8 points for the replay-and-KL, unified 0.8B, unified 2B, and unified 4B runs (to 66.8, 64.7, 67.3, and 70.6; Table~\ref{tab:decoding}) and leaves the CJ-IE-only run unchanged (73.3), whose best shift is higher. Selecting the shift on the dev split by macro F1 gives $s^{*}=0.5$ for the four runs that keep CJ-Bench accuracy and $1.0$ for the CJ-IE-only run, and results within 0.3 points of the default (the dev grid has no 0.75); an oracle shift chosen on the test split would add at most 0.2 points to a dev-selected result. A probability threshold of 0.5 lowers F1 by 0.4 to 1.3 points, and keeping overlapping spans changes it by at most 0.2. The cost of retention is therefore not an operating-point artifact: at the dev-selected shifts the CJ-IE-only run still leads the unified 0.8B model by $+8.97$ points [$+8.76$, $+9.19$], the unified 2B model by $+6.44$ [$+6.26$, $+6.61$], and the unified 4B model by $+3.10$ [$+2.92$, $+3.27$] (paired). In the precision--recall plane (Figure~\ref{fig:operating}, left), the CJ-IE-only curve lies above the other four at every recall; at 55\% recall its precision is 86.8\%, against 73.6\%, 77.7\%, and 82.6\% for the unified 0.8B, 2B, and 4B models.

The shift also moves calibration. Entity ECE is smallest between $s=1.25$ and $s=2$ for every run (1.2\% to 4.1\% at $\log 5$ across runs) and grows quickly on either side; at the F1-optimal shift 0.5 it is 14.2\% for the unified 0.8B, 11.1\% for the 2B, and 7.8\% for the 4B model. Deciding with $s_d$ while reporting probabilities at $\log 5$ keeps the probabilities usable, but not perfectly calibrated: entity ECE at the release default is 4.0\%, 2.3\%, and 4.2\% for the unified models, against 4.1\%, 1.8\%, and 3.2\% at the model-card point, with a small rise for the 2B and 4B models and a small fall for 0.8B. An application that thresholds on span probability should keep the model-card probabilities and set its own threshold, or fit a mapping on held-out data.

\begin{figure}[t]
\centering
\includegraphics[width=\textwidth]{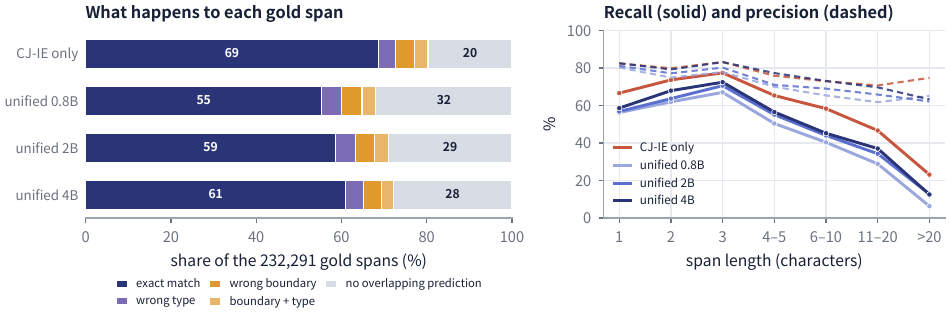}
\caption{\textbf{Where the errors fall} at the model-card operating point. Left: what happens to each of the 232,291 gold spans (numbers in \%). Right: recall (solid) and precision (dashed) by span length in characters; spans longer than 20 units cannot be represented.}
\label{fig:errors}
\end{figure}
The error taxonomy (Figure~\ref{fig:errors}, left) locates the cost. The unified 0.8B model matches 55.4\% of the gold spans exactly, the unified 2B model 58.6\%, the unified 4B model 61.0\%, and the CJ-IE-only run 68.7\%. The shares of wrong-type spans (4.0--4.6\%), wrong-boundary spans (4.2--4.8\%), and both (2.8--3.2\%) are nearly equal across the four. What differs is the share of gold spans overlapped by no prediction at all: 31.9\%, 29.0\%, 27.9\%, and 19.7\%. That accounts for 92\% of the exact-match gap between the CJ-IE-only run and the unified 0.8B and 2B models, and all of it for the 4B model, so retention costs spans the model never proposes, not spans it mislabels or mis-bounds. Among the predictions of the unified 0.8B model that are wrong, 44\% overlap no gold span, 25\% have the right type with the wrong boundary, 21\% have the right span with the wrong type, and 10\% have both wrong; the unified 4B model's wrong predictions split 40\%, 26\%, 23\%, and 10\%. Recall falls with length (Figure~\ref{fig:errors}, right): for the 4B model it is 59--72\% for spans of one to three characters, 45\% for six to ten, 37\% for eleven to twenty, and 13\% beyond, while precision stays between 63\% and 83\%. At the release default the share of gold spans with no overlapping prediction falls to 19.7\%, 17.0\%, and 15.7\% for the unified 0.8B, 2B, and 4B models (10.3\% for the CJ-IE-only run) and the share of spurious predictions among all predictions rises from 8.5--11.7\% to 12.5--16.1\% across the five runs.
\renewcommand{\sectioncolor}{cdTeal}

\section[Deployment]{Deployment and Quantization}
\label{sec:deployment}
\subsection{Serving Interface}
The serving implementation accepts context and a collection of named questions, validates their candidate schemas, and returns probabilities, derived decisions, and the sharpness statistic in Eq.~\eqref{eq:confidence}. A request supports up to 64 questions, each compiled separately. The GPU server supports dynamic batching; the local GGUF backend serializes access to its mutable inference context and fully resets the hybrid recurrent state between decisions, so no prefix cache is reused across questions. Thus one HTTP request can contain multiple decisions without implying joint reasoning or a shared model pass over all of them; the exception is a unified model's request that also carries entity types, whose short questions share one pass with the span row (Section~\ref{sec:unified-serving}). Both servers expose the candidate tournament of Section~\ref{sec:tournament} as a separate endpoint for \choice{} questions with up to 1,024 candidates; a field that fits into one question is answered identically to the ordinary endpoint.

The Transformers and llama.cpp~\citep{llamacpp} backends use the same candidate-only normalization. The GGUF implementation checks tokenization consistency and preserves the input-length and question-type rules. For each size we evaluate the merged BF16 release and a Q8\_0 file converted directly from it; neither is a separate decision-training run.

\subsection{Serving a Unified Model}
\label{sec:unified-serving}
A unified model is loaded once and answers both request shapes. The server exposes \texttt{POST /v1/predict}, which routes by request shape (questions: decisions; types: extraction; both: both answers about the same input), \texttt{POST /v1/decide} (the decision interface above), \texttt{POST /v1/extract} (a text, a mapping from type name to definition, and an optional probability threshold; it returns spans with character offsets and probabilities), and \texttt{GET /v1/info} (model name, temperature, whether extraction is available). It also keeps the other endpoints of the decision server (batch, compile, tournament), so one process replaces the decision-only service. Decision Studio reaches it through one environment variable and adds an extraction mode, an extraction case gallery, an event pipeline, and a triple-extraction scenario. Because both readouts live in one safetensors file, a unified 2B model needs 3.79\,GB of weights for both abilities against 3.76\,GB for the decision-only 2B model alone, and a unified 4B model 8.44\,GB against 8.41\,GB.

\begin{table}[t]
\caption{\textbf{Latency of the unified models} for decisions, extraction, and requests with both readouts.}
\label{tab:serving}
\centering
{\footnotesize
\setlength{\tabcolsep}{4.0pt}
\begin{tabular}{lrrrr}
\arrayrulecolor{cdInk}\toprule
\rowcolor{cdMist}Request & Tokens & \uniA\,0.8B & \uniB\,2B & \uniC\,4B \\
\arrayrulecolor{cdLine}\midrule
Decision, one question & -- & 44 & 45 & 60 \\
Decision, three questions (one pass each) & -- & 133 & 135 & 179 \\
Extraction, 57 characters, 3 types & 218 & 53 & 53 & 69 \\
Extraction, 44 characters, 17 types & 627 & 54 & 55 & 70 \\
Extraction, 318 characters, 3 types & 617 & 61 & 63 & 79 \\
\arrayrulecolor{cdLine}\midrule
One question and extraction, two calls & -- & 97 & 98 & 129 \\
One question and extraction, one request & -- & \textbf{49} & \textbf{49} & \textbf{65} \\
Three questions and extraction, two calls & -- & 185 & 187 & 248 \\
Three questions and extraction, one request & -- & \textbf{49} & \textbf{50} & \textbf{67} \\
\arrayrulecolor{cdInk}\bottomrule
\end{tabular}
\par\smallskip\begin{minipage}{\linewidth}\tablenote{Median latency in ms (40 warm repetitions, one request at a time) of the released unified models in BF16 on one idle A800; the 95th percentile is within 7\% of the median everywhere. Tokens: length of the extraction input (text twice plus the type list), 2B tokenizer. Peak allocated memory: 1.8, 4.2, and 9.1 GB.}\end{minipage}
}

\end{table}
Extraction is slower than a decision at the same size because the input is longer and the span head scores every candidate span. We timed the three released unified models on one idle A800 in a Python call that includes tokenization, batching, GPU work, and decoding but not HTTP (Table~\ref{tab:serving}). A decision with one question takes 44\,ms at 0.8B and 2B and 60\,ms at 4B, and an extraction 53 to 63\,ms (69 to 79\,ms at 4B), growing with the input from 218 to 627 tokens. The two smaller models are not compute-bound at these lengths, so their latencies are nearly equal, as for the decision-only models (Section~\ref{sec:deployment-cost}). A request that carries both readouts and is short enough is answered in one backbone pass (Section~\ref{sec:joint}): it takes 49, 49, and 65\,ms with one question and 49, 50, and 67\,ms with three, against 97, 98, and 129\,ms and 185, 187, and 248\,ms for the same work as separate calls, which score the questions one at a time. The one-pass path is not bit-identical to the two-pass path, because left padding changes BF16 rounding. In a check of 210 CJ-IE test rows with three questions each (630 decisions), the largest probability difference was 0.027 at 0.8B, 0.035 at 2B, and 0.028 at 4B; 3, 5, and 7 decisions changed their top candidate (the questions were not about the rows, so many probabilities were near the boundary), and 8, 9, and 4 rows returned a different entity set, spans near the decision threshold. Peak allocated memory in the timing run was 1.8, 4.2, and 9.1\,GB. The servers are shared, and these numbers are not comparable with the decision latencies of Figure~\ref{fig:cost} beyond their order of magnitude. Version 4 reported 91 to 99.5\,ms for 0.8B extraction, measured on an earlier run with an earlier implementation.

\subsection{Full-Benchmark Quantization Results}
\label{sec:quant}
\begin{table}[t]
\caption{\textbf{Q8\_0 against its BF16 source}, paired over all 307,900 decisions at the same temperature.}
\label{tab:quantization}
\centering
{\footnotesize
\setlength{\tabcolsep}{4.2pt}
\begin{tabular}{lcrrrrrrr}
\arrayrulecolor{cdInk}\toprule
\rowcolor{cdMist}Size & Weights (GB) & $\Delta$ACC (pp) & Top-1 (\%) & R$\to$W / W$\to$R & Mean $|\Delta p|$ & p95 max $|\Delta p|$ & Max $|\Delta p|$ & $\Delta$ECE (pp) \\
\arrayrulecolor{cdLine}\midrule
\szA\,0.8B & 1.50 $\to$ 0.81 & +0.0180 & 98.92 & 1,185 / 1,240 & 0.0024 & 0.0236 & 0.2269 & -0.001 \\
\szB\,2B & 3.76 $\to$ 2.01 & -0.0013 & 99.00 & 1,105 / 1,101 & 0.0023 & 0.0237 & 0.2752 & +0.017 \\
\szC\,4B & 8.41 $\to$ 4.48 & +0.0177 & 99.19 & 948 / 1,002 & 0.0019 & 0.0205 & 0.1679 & -0.040 \\
\arrayrulecolor{cdInk}\bottomrule
\end{tabular}
\par\smallskip\begin{minipage}{\linewidth}\tablenote{Q8\_0 (llama.cpp CUDA) versus its BF16 source (Transformers) on all 307,900 decisions at the same temperature. R$\to$W: hard targets right in BF16 and wrong in Q8\_0; W$\to$R: the reverse. $|\Delta p|$ is the absolute candidate-probability difference.}\end{minipage}
}

\end{table}
Table~\ref{tab:quantization} compares each Q8\_0 model with its BF16 source on all 307,900 decisions at the same temperature. Q8\_0 roughly halves the weight files (to 0.81, 2.01, and 4.48\,GB). Micro accuracy changes by at most 0.018 points and ECE by at most 0.04 points. Top-1 agreement is 98.92\%, 99.00\%, and 99.19\% for 0.8B, 2B, and 4B. Predictions are not identical: 948 to 1,240 hard targets change correctness in each direction, and the largest per-decision probability difference reaches 0.17--0.28. Agreement is lowest on \score{} questions (96.6--97.1\%), where adjacent levels often have similar probabilities. Small domain shifts should not be read as capability changes. For example, the 2B Q8\_0 model has 0.15 points higher macro accuracy, which comes from 28 additional correct legal decisions out of 4,300.

Quantization is not the only source of such changes. Temperature cannot change an argmax, yet each merged BF16 release agrees with its own unmerged adapter on the top-1 candidate for only 99.32\%, 99.34\%, and 99.55\% of decisions, because merging LoRA into BF16 weights changes the logits. Merging thus changes 0.45--0.68\% of top-1 predictions, against 0.81--1.08\% for quantization. A 1,200-decision control recorded with the releases separates backend from precision. For 4B, switching only the backend (Transformers BF16 to BF16 GGUF) and switching only the precision (BF16 GGUF to Q8\_0) each preserve 99.17\% of top-1 predictions. For 2B, BF16 GGUF and Q8\_0 agree on 99.08\%, and Q8\_0 and Transformers BF16 on 98.75\%.

\begin{finding}
Q8\_0 is safe in aggregate at every size, but merging, backend, and precision each flip a fraction of a percent of individual predictions. Applications with decision thresholds should validate the exact artifact they deploy.
\end{finding}

\subsection{Measured Inference Cost}
\label{sec:deployment-cost}
\begin{figure}[t]
\centering
\includegraphics[width=\textwidth]{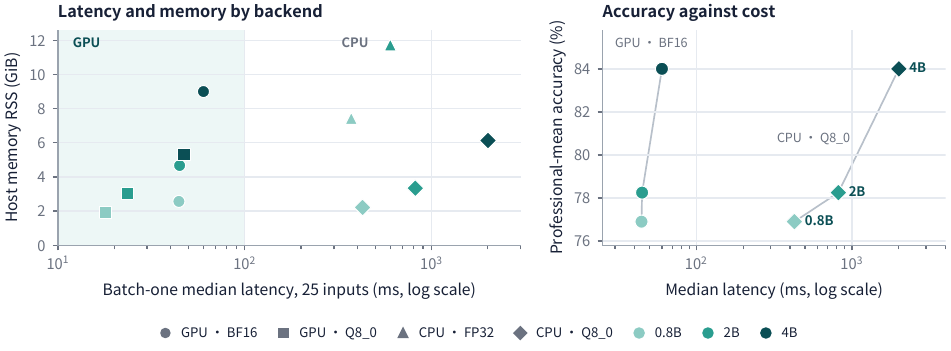}
\caption{\textbf{Inference cost under one protocol}: the same 25 benchmark inputs, three warm repeats each, batch size one, on a shared server (A800 80GB PCIe; Xeon Gold 5318Y with 16 threads). Left: median latency against host memory for every measured backend. Right: professional-mean accuracy against median latency for GPU BF16 (circles) and CPU Q8\_0 (diamonds). Appendix Table~\ref{tab:efficiency} lists all values.}
\label{fig:cost}
\end{figure}
Figure~\ref{fig:cost} applies one protocol to every released model. On the A800, BF16 batch-one median latency is 44.4\,ms for 0.8B, 44.8\,ms for 2B, and 60.1\,ms for 4B. At these input lengths the two smaller models are not compute-bound, so their latencies are nearly equal. Q8\_0 on the GPU through llama.cpp is faster at the median (17.9, 23.6, and 47.3\,ms) but has a heavier tail on long inputs, with 95th percentiles of 193--569\,ms against 55--199\,ms for Transformers BF16. On 16 CPU threads, Q8\_0 median latency is 0.43, 0.82, and 2.00\,s and resident memory is 2.2, 3.3, and 6.1\,GiB; long inputs take up to tens of seconds. For 0.8B and 2B, Transformers FP32 on the CPU is faster than Q8\_0 at the median but needs 3.4--3.5 times the memory. These figures support CPU deployment for short inputs and offline batches, not a universal quantization speedup.

Separate measurements give context but are not comparable to the figure. The full-benchmark evaluations, with batches of up to 65,536 padded tokens, peak at 6.7, 10.6, and 18.2\,GB of GPU memory for the 0.8B, 2B, and 4B adapters. Earlier 60-decision medians on the A800 that include input preparation were 70.1, 69.0, and 90.3\,ms for the unmerged adapters, and 44.7 and 59.8\,ms for the merged 2B and 4B models. On a Windows desktop (Intel Core i7-9700, eight threads), the documented short-input Q8\_0 median is 1.35\,s for 2B and 2.55\,s for 4B. None of these timings include HTTP or network delay, and the servers were shared with other jobs.

\subsection{Application Workbench}
Decision Studio provides a workbench for ticket triage, input/output guardrails, live-message moderation, and agent routing, and can be served by the GPU or CPU server of any released size. These examples combine model probabilities with explicit application rules. A ticket application can ask independently about category, urgency, and refund intent, as in Figure~\ref{fig:teaser}a; a routing application can require higher confidence for consequential actions. Such thresholds are application policies rather than learned guarantees.

The workbench is an integration artifact, not an additional labeled benchmark. Its screenshots and example timings are not included in quantitative accuracy claims. One documented qualitative failure is that spoiler detection can react to the word ``spoiler'' in a request asking others not to reveal spoilers. Controlled robustness, prompt-injection, and out-of-distribution evaluations remain open.

\section{Limitations}
\label{sec:limits}
\paragraph{Scope of the empirical contribution.}
The study evaluates three decision-only runs and three unified runs, one per configuration, on one decision benchmark family and one internal extraction split. The method adapts established components and is best viewed as an empirical model-and-systems report. We do not claim a new decision architecture, a new dataset beyond CJ-IE, or a general state-of-the-art result across Chinese understanding or extraction tasks.

\paragraph{Uncontrolled comparisons.}
Imported baselines have not been reproduced locally, and differences in model capacity, training objectives, data scheduling, and evaluation implementation may contribute to observed gaps. All our members are larger than the approximately 322M-parameter Chinese-Jev models, so the comparison is between complete systems rather than an architecture at matched size. Within the family, three sets of comparisons are only partly controlled. In the decision-only series, the 0.8B--2B step changes size, objective, and batch together, and the 2B--4B step is close to controlled but still changes the token budget and therefore the number and composition of updates; the three sizes do not establish a scaling law. In the unified series, the three sizes share one recipe and one objective, but the 2B run uses half the token budget and the 4B run gradient checkpointing; each unified--decision-only pair also differs in more than the extraction task. The five joint-training runs differ in starting point, data, objective, and batch composition at once, and no extraction-only continuation exists at 2B or 4B, so the shrinking retention cost across sizes is measured against a 0.8B reference (Section~\ref{sec:retention-cost}). No CE-only run exists at 2B or 4B, so the contribution of RLCD is unmeasured; the only CE versus CE+RLCD contrast, decision-only versus unified 0.8B, also differs in CJ-IE data. Removing autoregressive generation does not establish a latency advantage over compact encoders, and hardware differences preclude a speedup ratio against imported timings.

\paragraph{Uncertainty and dependence.}
The clustered intervals cover test-set sampling for fixed models, not variation across training seeds, data orders, or GPU counts; each configuration was trained once, and the 0.8B CJ-IE-only run was evaluated unmerged with a different batch composition than the others. The source-level tests are uncorrected for multiplicity. The same test set has been evaluated in full for 14 checkpoints, six decision-only released files, and five unified-run evaluations. Selection never used test labels, but repeated inspection still risks adapting design choices to CJ-Bench, and a fresh holdout would strengthen confirmation.

\paragraph{Extraction scope.}
The CJ-IE test split is an internal, standardized, sampled split. Its type names and definitions are the ones seen in training, so the scores are best-case for type definitions and cannot be compared with published leaderboards. We evaluated no unseen type systematically; a single one-line type (``product'') worked in one example (Appendix~\ref{app:examples}), and sources whose types lack definitions score zero. Relation and event scores are conditional on the subject and the event type, and are not end-to-end triple or record scores. Recall is low (55--61\% at the model-card operating point, 63--69\% at the release default): most missed spans receive no proposal, long spans are hard, and spans longer than 20 units cannot be represented (0.59\% of test spans). A specialist trained on one corpus can be much better on that corpus: an earlier feasibility study with a 0.8B decision model, 4,000 CLUENER sentences, and the same head reached F1 78.7 on the official CLUENER dev split, which is the CLUENER test split of CJ-IE (1,343 sentences), against 55.7, 62.5, and 70.7 for the unified 0.8B, 2B, and 4B models; the type definitions, training data, and recipe differ, so this is a caution rather than a controlled measurement. The extraction readout is available only through PyTorch.

\paragraph{Calibration and distribution shift.}
ECE depends on confidence binning, sample composition, and the evaluation distribution. Section~\ref{sec:calibration} shows that pooled ECE can hide miscalibration of opposite sign in different domains, that a global temperature can worsen individual domains, and that merging alone can move ECE on a small domain by more than the temperature does; the unified models show the same pattern. The 4B decision-only model has the highest general-domain and pooled ECE of the decision-only family, and its product-catalog source has 22.6\% source-level ECE. Span probabilities depend on the none shift: the shift that maximizes F1 gives entity ECE of 8\% to 14\% for the unified models, and the release default decides with one shift and reports probabilities at another, which keeps entity ECE at 2.3\% to 4.2\%. The API sharpness field is not a calibrated probability of correctness. Classwise calibration and shifted-domain tests are not reported, and high-confidence outputs can still be wrong.

\paragraph{Data and task semantics.}
We inherit the upstream split construction and labels. Format and checksum audits and parent-separated calibration do not establish absence of overlap with backbone pretraining, and CJ-Dataset's finance domain includes a relation corpus (FinRE) that CJ-IE also uses, so an overlap between the two collections is possible and was not audited. A medical matching label, for example, measures the original matching task rather than independently verified medical correctness. The same distinction applies to legal and financial labels. Negative relation pairs in CJ-IE are entity pairs that the annotation allows but does not mark, which need not be absent relations. Probabilities should not be the sole basis for consequential professional decisions.

\paragraph{Missing ablations and robustness studies.}
The available experiments do not isolate soft versus hard targets, candidate permutation, mixing strategy, LoRA rank, RLCD, the share of span batches, the echo layout at scale, or the product-catalog effect of the unified recipe. Ordered-score performance remains the weakest at every size. Stability under candidate reordering, paraphrases, distractor additions, adversarial inputs, and abstention requirements requires direct evaluation. The candidate tournament has no labeled evaluation; its accuracy for more than 16 candidates is unknown, and its finalist-only probabilities are not comparable with single-question probabilities. An explicit ``none of the above'' option may be needed when the provided candidate set is incomplete; the present softmax otherwise allocates all mass among supplied candidates.

\paragraph{Release and licensing.}
The project code is MIT-licensed, with the adapted RLCD component under Apache-2.0 and preserved third-party notices. Released weights keep the Apache-2.0 license of their Qwen3.5 backbones. Data permissions are source-specific and include non-commercial restrictions for some sources, and some CJ-IE corpora have unclear terms; this project does not grant additional rights to redistribute or commercially reuse those sources, and CJ-IE itself is not released with the models. A model release does not by itself resolve every downstream data-use requirement.

\section{Conclusion}
\model{} adapts causal language backbones into a common single-pass interface for Chinese probabilistic decisions and, in its unified members, typed span extraction from the same weight file. The 4B models reach 85.97\% (decision-only) and 86.86\% (unified) on CJ-Bench, above every reported baseline in each domain, with gains over the smaller models that are largest where training data are scarcest. The unified models keep decision accuracy while extracting spans (68.9 micro F1 at 4B), but their decision gains over decision-only models are small. The central finding about the unified recipe is a price that falls with size: against a 0.8B model continued on extraction data alone, which forgets 12 points of CJ-Bench, keeping decision ability costs 10.7, 7.9, and 5.0 F1 points at 0.8B, 2B, and 4B, mostly as spans the model never proposes. Global temperature scaling, merging, and quantization change calibration and individual predictions in ways that aggregates hide. Single-factor ablations of the recipe, repeated seeds, a longer schedule at the smaller sizes, extraction-only continuations at 2B and 4B, systematic evaluation on unseen type systems, per-domain calibration, and robustness to candidate order and paraphrase are the next steps needed to strengthen these conclusions.

\section*{Acknowledgments}
We thank the creators of Qwen3.5, CJ-Dataset and CJ-Bench, Chinese-Jev, CalOn, \texttt{hmm}, B2NERD, the Chinese NER, relation, and event corpora used in CJ-IE, and llama.cpp for their publicly available models, data, and software. The project documentation and third-party notices record the implementation lineage. No additional funding or collaboration is asserted in this report.

\FloatBarrier
\bibliographystyle{plainnat}
{\small\bibliography{references}}
\clearpage
\appendix
\addcontentsline{toc}{section}{\protect\numberline{A--G}Appendices}
\addtocontents{toc}{\protect\setcounter{tocdepth}{-1}}
\renewcommand{\sectioncolor}{cdAmber}
\section{Prompt and Interface Details}
\label{app:prompt}
The user-message bodies below are English translations for readability. The actual compiler uses Chinese labels and applies the Qwen chat template with \texttt{enable\_thinking=False} and an assistant generation prefix. It does not append a target answer to the model input. A \emph{decision} prompt reads:
\begin{lstlisting}
Context to judge:
{state}

Question:
{instructions}

Candidates:
A: {key_1} -- {description_1}
B: {key_2} -- {description_2}
...
Return only the corresponding candidate letter.
\end{lstlisting}
Structured contexts or descriptions are serialized as compact JSON with Unicode preserved. \choice{} candidate insertion order defines its inference-time mapping; training-time permutation updates both the rendered options and target vector. \score{} levels are indexed from zero, while its internal letter indices start with A as for other types.

An \emph{extraction} prompt (unified models only) reads:
\begin{lstlisting}
Find the entities of the following types in the text.
Entity types:
- {name_1}: {definition_1}
- {name_2}: {definition_2}
...
Text:
{text, natural tokenization}
Character by character:
{text again, one unit per position}
\end{lstlisting}
The type list is shuffled during training so that a type vector cannot depend on its position. Each unit of the second copy (one CJK character or symbol, one ASCII word or number, or one whitespace character) is tokenized separately, and span candidates are read from that copy. Each type contributes the mean and last-token states of its name-and-definition tokens to a type vector.

An illustrative English-language decision request is shown below. It documents the schema, not a new measured model prediction.
\begin{lstlisting}
{
  "state": "The customer has requested an invoice three times.",
  "questions": {
    "department": {
      "type": "choice",
      "instructions": "Which team should handle this?",
      "criteria": {
        "billing": "Payments, invoices, refunds",
        "technical": "Faults and system issues",
        "sales": "Purchase and upgrade inquiries"
      }
    }
  }
}
\end{lstlisting}
For a categorical question, the response returns the probability map, the winning key, and the sharpness statistic. \noul{} adds the true probability; \score{} adds the expected level index and a legend. Reported output-token usage is zero because the service does not generate or decode an answer token. A tournament request carries one \choice{} question with optional \texttt{group\_size}, \texttt{keep}, and \texttt{seed}; its response adds every elimination round with the group probabilities and the advancing keys. An extraction request and response have the following shape:
\begin{lstlisting}
{"text": "Wang Xiaoming bought a fridge in Hangzhou.",
 "types": {"person": "Name of a person",
           "location": "Name of a place"},
 "threshold": 0.3}
-> {"task": "extraction",
    "entities": [{"text": "Wang Xiaoming", "label": "person",
                  "start": 0, "end": 13, "score": 0.99}, ...]}
\end{lstlisting}
Offsets are character indices, half-open. A request that carries both \texttt{questions} and \texttt{types} receives both answers about the same input (\texttt{"task": "decision+extraction"}, with \texttt{answers} and \texttt{entities}); a request that carries neither is rejected.

\section{Worked Extraction Examples}
\label{app:examples}
\begin{figure}[t]
\centering
\includegraphics[width=\textwidth]{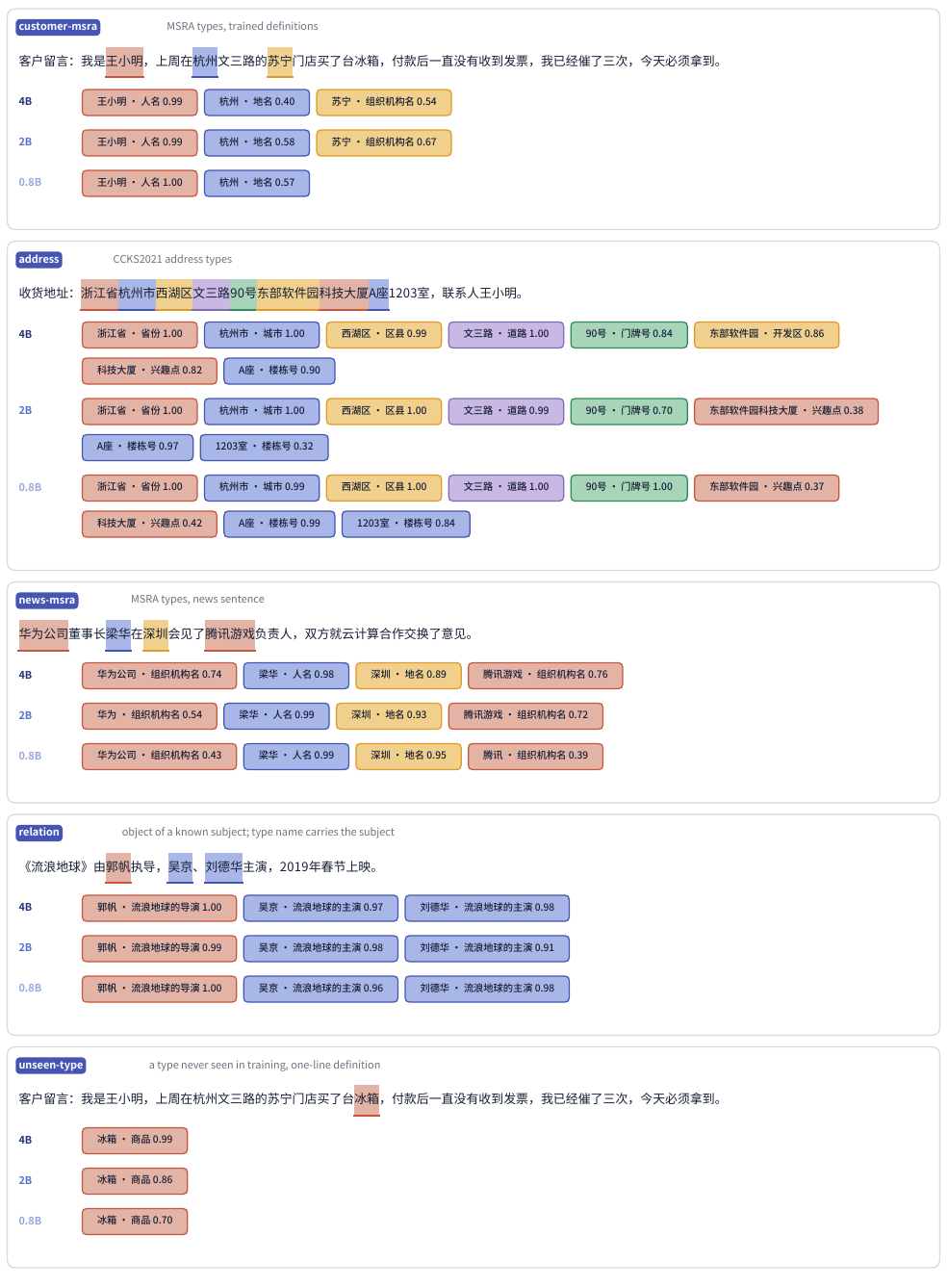}
\caption{\textbf{Real outputs of the released unified models} on five hand-written requests (GPU, BF16, release defaults, no probability threshold). Highlighted spans in the text are those of the 4B model; the chips list every span of each model with its probability. The types of the first three requests are copied from the CJ-IE training files, so they carry the trained definitions (MSRA, CCKS2021 address). The relation request puts the subject into the type name. The last request uses one type that never appears in training, with a one-line definition; it worked here, but one example is not evidence of a capability.}
\label{fig:gallery}
\end{figure}
Figure~\ref{fig:gallery} shows what the span readout returns. Several observations are visible without any claim of generality. The 0.8B model proposes no span for the organization in the first request, which the 2B and 4B models find with probabilities 0.67 and 0.54. In the news sentence, the 0.8B model marks only part of the second organization (0.39) and the 2B model only part of the first (0.54), while the 4B model returns both with full boundaries (0.74 and 0.76). In the address request, all three models agree on the five clean elements. The room number is found by the 0.8B model (0.84) and, at a low probability, by the 2B model (0.32), but not by the 4B model; the compound place name is split into two spans by the 0.8B and 4B models and returned as one by the 2B model. The relation request returns the same director and the same two actors at every size, with probabilities between 0.91 and 1.00, and the one-line type of the last request is found at every size (0.70, 0.86, and 0.99). In these examples the least certain spans have probabilities below about 0.7, which is where a threshold would act.

\section{CJ-IE Details}
\label{app:cjie}
\paragraph{Task views.} Every source record produces the views that its annotations support, always within one split. NER rows list the corpus's types and definitions. For a marked mention, \texttt{ner\_typing} asks which type it has (at most 16 types). Relation rows (\texttt{re\_object}) contain one subject each, with the predicate and subject in the type name and the object as the gold span; \texttt{re\_pair} asks for the relation of a given ordered pair, where negatives are ordered pairs allowed by the types but unannotated, at half the number of positives. Event rows (\texttt{ee\_argument}) contain one event type each, with the trigger and the roles as types; 25\% of sentences get an extra row with an absent event type and no spans. \texttt{ee\_type} asks whether a sentence describes an event type, and \texttt{ee\_category} which category a one-event sentence belongs to. The ChFinAnn announcements yield \texttt{doc\_event} (is an equity event disclosed) and \texttt{doc\_field} (which mention fills a field, or none) when all records of an event agree on the value. Quality flags mark gold, silver (WikiANN, MultiNERD, automatic), and distant (ChFinAnn) sources.

\paragraph{Leakage and duplicates.} The largest removals are 4,955 MSRA training records that belong to People's Daily dev or test, 2,506 CMeIE and 916 CMeEE-V2 records that share medical passages, and 59 training windows whose text equals a held-out window. After construction, no normalized text is shared between any two splits. Duplicates across sources inside training are kept, since the same sentence under two annotation rules illustrates why definitions matter; sources wholly contained in another (CCIR2021, finance-sina, most of CCKS2018) are not included. A relation corpus whose labels are bag-level distant supervision (IPRE), English corpora, and a traditional-Chinese fact-checking set are excluded.

\paragraph{Sources.} Table~\ref{tab:sources-ie} lists the 35 span sources of the test split with the exact-match scores of the five runs and the entity ECE of the unified 4B model. The test split of the People's Daily NER corpus has no training counterpart because its training text is contained in MSRA.

\begin{table}[!ht]
\caption{\textbf{All 35 span sources of the CJ-IE test split.} Exact-match precision, recall, and F1 of the unified 4B model, F1 of the other four runs, and the 4B entity ECE.}
\label{tab:sources-ie}
\centering
{\scriptsize
\setlength{\tabcolsep}{2.4pt}
\begin{tabular}{llrrrrrrrrr}
\arrayrulecolor{cdInk}\toprule
\rowcolor{cdMist}& & & \multicolumn{3}{c}{Unified 4B (\%)} & \multicolumn{4}{c}{F1 of other runs (\%)} & \\
\rowcolor{cdMist}Source & Family & Gold & P & R & F1 & Unif.\ 2B & Unif.\ 0.8B & Repl.+KL & CJ-IE only & ECE (\%) \\
\arrayrulecolor{cdLine}\midrule
Resume & NER & 1,630 & 93.7 & 88.3 & \textbf{90.9} & 87.3 & 88.1 & 89.5 & 93.3 & 6.2 \\
CCKS2021 address & NER & 9,828 & 91.2 & 86.6 & \textbf{88.9} & 86.4 & 85.8 & 86.7 & 90.2 & 0.8 \\
NLPCC2018 & NER & 505 & 90.5 & 86.9 & \textbf{88.7} & 85.2 & 82.9 & 85.3 & 90.7 & 3.9 \\
IMCS21 & NER & 4,108 & 90.3 & 82.5 & \textbf{86.3} & 86.7 & 85.7 & 85.6 & 88.7 & 4.1 \\
People's Daily & NER & 7,707 & 93.8 & 79.7 & \textbf{86.2} & 84.3 & 80.8 & 81.9 & 90.1 & 9.4 \\
MSRA & NER & 6,160 & 93.9 & 73.8 & \textbf{82.6} & 82.2 & 76.0 & 78.5 & 88.3 & 9.6 \\
Youku & NER & 1,573 & 89.4 & 74.3 & \textbf{81.2} & 78.8 & 76.6 & 78.4 & 82.5 & 4.6 \\
zh-OntoNotes & NER & 7,359 & 83.6 & 77.6 & \textbf{80.5} & 77.9 & 77.3 & 76.2 & 81.2 & 2.0 \\
Math (high school) & NER & 1,514 & 83.3 & 75.2 & \textbf{79.0} & 77.2 & 67.0 & 73.9 & 81.9 & 3.8 \\
Wanchuang & NER & 2,222 & 85.1 & 73.4 & \textbf{78.9} & 71.8 & 70.1 & 71.1 & 83.2 & 9.1 \\
DuIE & relation & 9,433 & 79.6 & 77.4 & \textbf{78.5} & 74.3 & 66.2 & 71.5 & 79.4 & 3.3 \\
CAIL2022 & NER & 1,159 & 87.2 & 70.8 & \textbf{78.2} & 70.5 & 63.0 & 69.9 & 83.6 & 9.3 \\
WikiANN & NER & 6,234 & 79.0 & 77.2 & \textbf{78.1} & 76.2 & 73.8 & 75.2 & 80.7 & 4.7 \\
Bank & NER & 772 & 80.2 & 74.0 & \textbf{77.0} & 71.3 & 71.5 & 73.5 & 78.9 & 3.8 \\
People's Daily 2014 & NER & 13,760 & 82.3 & 71.9 & \textbf{76.8} & 74.1 & 72.5 & 74.7 & 84.9 & 1.3 \\
Taobao & NER & 4,677 & 81.4 & 69.0 & \textbf{74.7} & 70.9 & 67.3 & 71.3 & 76.0 & 2.2 \\
MultiNERD & NER & 6,545 & 73.9 & 68.9 & \textbf{71.3} & 68.5 & 66.6 & 67.9 & 76.8 & 7.0 \\
CCKS2020 & NER & 1,496 & 80.2 & 63.8 & \textbf{71.0} & 67.6 & 65.0 & 67.0 & 74.8 & 7.4 \\
CLUENER & NER & 3,072 & 77.1 & 65.3 & \textbf{70.7} & 62.5 & 55.7 & 64.3 & 75.6 & 3.9 \\
ChineseLiterature & NER & 16,455 & 80.2 & 62.5 & \textbf{70.3} & 68.8 & 67.1 & 67.7 & 72.4 & 3.5 \\
CAIL2021 & NER & 5,427 & 81.7 & 59.6 & \textbf{68.9} & 68.9 & 62.4 & 65.5 & 81.4 & 12.0 \\
DuEE & event & 5,250 & 73.5 & 63.8 & \textbf{68.3} & 62.2 & 56.3 & 58.2 & 72.1 & 1.2 \\
GAIIC2022 & NER & 5,356 & 75.5 & 61.4 & \textbf{67.7} & 64.7 & 60.2 & 62.8 & 72.2 & 2.7 \\
Boson & NER & 1,315 & 80.3 & 56.7 & \textbf{66.4} & 61.7 & 49.7 & 56.1 & 73.2 & 10.7 \\
FNED & NER & 10,611 & 78.5 & 54.6 & \textbf{64.4} & 62.3 & 60.1 & 61.8 & 71.8 & 6.4 \\
Ruijin MCC & NER & 14,774 & 80.3 & 53.3 & \textbf{64.1} & 63.4 & 63.1 & 63.3 & 69.9 & 4.8 \\
CCKS2019 & NER & 1,638 & 76.2 & 55.1 & \textbf{64.0} & 58.8 & 59.4 & 57.9 & 71.5 & 4.6 \\
Weibo & NER & 410 & 77.1 & 54.1 & \textbf{63.6} & 52.6 & 37.2 & 44.8 & 62.7 & 13.5 \\
CCKS2017 & NER & 5,611 & 77.5 & 52.9 & \textbf{62.9} & 58.8 & 58.5 & 62.1 & 78.6 & 7.2 \\
CMeEE-V2 & NER & 26,081 & 77.5 & 52.4 & \textbf{62.6} & 58.5 & 58.6 & 58.7 & 66.5 & 3.8 \\
MultiCoNER22 & NER & 5,707 & 69.3 & 52.6 & \textbf{59.8} & 52.8 & 47.7 & 49.0 & 57.1 & 4.0 \\
FewFC & event & 24,834 & 68.8 & 52.8 & \textbf{59.8} & 56.1 & 52.6 & 57.3 & 66.1 & 5.1 \\
MultiCoNER2 & NER & 6,899 & 53.3 & 34.6 & \textbf{42.0} & 40.0 & 34.7 & 40.0 & 57.7 & 12.5 \\
CMeIE & relation & 11,522 & 64.1 & 24.4 & \textbf{35.4} & 36.5 & 28.2 & 26.0 & 50.1 & 6.6 \\
Military & NER & 647 & 58.8 & 1.5 & \textbf{3.0} & 0.0 & 0.0 & 3.0 & 17.3 & 26.2 \\
\arrayrulecolor{cdInk}\bottomrule
\end{tabular}
\par\smallskip\begin{minipage}{\linewidth}\tablenote{Exact-match scores on the CJ-IE test split at the model-card operating point (none shift $\log 5$), sorted by the unified 4B F1. ECE: entity-level (10 bins) of the unified 4B model. The Military corpus has four types with identical names and definitions.}\end{minipage}
}

\end{table}

\section{Additional Benchmark Breakdown}
\label{app:breakdown}
\begin{table}[!ht]
\caption{\textbf{Domain-by-type hard-target accuracy (\%)} of the released decision-only models.}
\label{tab:domain-type}
\centering
{\small
\setlength{\tabcolsep}{4.5pt}
\begin{tabular}{lrrrrrrrrr}
\arrayrulecolor{cdInk}\toprule
\rowcolor{cdMist}& \multicolumn{3}{c}{\noul{}} & \multicolumn{3}{c}{\choice{}} & \multicolumn{3}{c}{\score{}} \\
\rowcolor{cdMist}Domain & \szA\,0.8B & \szB\,2B & \szC\,4B & \szA\,0.8B & \szB\,2B & \szC\,4B & \szA\,0.8B & \szB\,2B & \szC\,4B \\
\arrayrulecolor{cdLine}\midrule
General & 87.59 & 88.01 & \best{88.50} & 72.93 & 76.16 & \best{78.60} & 54.64 & 56.05 & \best{58.35} \\
Medical & 97.82 & 97.87 & \best{98.04} & 80.55 & 81.81 & \best{85.04} & 66.50 & 66.19 & \best{67.49} \\
Legal & 85.60 & 82.07 & \best{88.87} & 64.07 & 67.20 & \best{74.13} & 51.69 & 55.15 & \best{67.31} \\
Finance & 76.33 & 81.75 & \best{85.67} & 62.83 & 67.42 & \best{77.00} & 82.83 & 80.33 & \best{88.17} \\
\arrayrulecolor{cdInk}\bottomrule
\end{tabular}
}

\end{table}
\begin{table}[!ht]
\caption{\textbf{Every full-benchmark checkpoint evaluation} of the three decision-only training runs.}
\label{tab:trajectory}
\centering
{\footnotesize
\setlength{\tabcolsep}{3.8pt}
\begin{tabular}{lrrrrrrrrrrr}
\arrayrulecolor{cdInk}\toprule
\rowcolor{cdMist}Size & Step & Prog.\,\% & Val.\ NLL & General & Medical & Legal & Finance & Prof. & Micro & Prof.\ ECE & Micro ECE \\
\arrayrulecolor{cdLine}\midrule
\szA\,0.8B & 2,000 & 19.8 & 0.739 & 69.04 & 86.78 & 54.07 & 64.17 & 68.34 & 80.41 & 2.73 & 1.63 \\
\szA\,0.8B & 3,000 & 29.7 & 0.691 & 70.26 & 87.22 & 58.93 & 68.67 & 71.61 & 81.21 & 1.75 & 1.25 \\
\szA\,0.8B & 5,000 & 49.4 & 0.676 & 70.90 & 87.89 & 63.37 & 71.17 & 74.14 & 81.95 & 1.49 & 0.84 \\
\rowcolor{cdTealLight}\szA\,0.8B & 10,000 & 98.7 & 0.600 & 72.24 & 88.86 & 67.86 & 74.22 & 76.98 & 83.11 & 2.46 & 1.07 \\
\arrayrulecolor{cdLine}\midrule
\szB\,2B & 1,000 & 20.1 & 0.694 & 70.55 & 87.69 & 55.67 & 66.94 & 70.10 & 81.55 & 3.71 & 2.03 \\
\szB\,2B & 2,000 & 40.1 & 0.630 & 72.56 & 88.54 & 62.67 & 71.44 & 74.22 & 82.89 & 2.28 & 1.27 \\
\szB\,2B & 3,000 & 60.1 & 0.603 & 71.87 & 88.99 & 65.56 & 74.31 & 76.28 & 83.04 & 1.63 & 1.63 \\
\szB\,2B & 4,000 & 80.2 & 0.592 & 73.77 & 89.42 & 68.40 & 76.03 & 77.95 & 83.99 & 1.74 & 0.85 \\
\rowcolor{cdTealLight}\szB\,2B & 4,994 & 100.0 & 0.586 & 73.95 & 89.50 & 69.05 & 76.03 & 78.19 & 84.11 & 2.17 & 1.49 \\
\arrayrulecolor{cdLine}\midrule
\szC\,4B & 2,000 & 36.7 & 0.536 & 74.71 & 90.26 & 72.12 & 79.36 & 80.58 & 84.93 & 1.73 & 1.85 \\
\szC\,4B & 3,000 & 55.0 & 0.512 & 75.15 & 90.66 & 72.67 & 81.00 & 81.44 & 85.36 & 1.05 & 1.19 \\
\szC\,4B & 4,000 & 73.4 & 0.477 & 75.38 & 91.03 & 75.91 & 82.36 & 83.10 & 85.74 & 1.63 & 1.31 \\
\szC\,4B & 5,000 & 91.8 & 0.458 & 75.95 & 91.10 & 76.12 & 82.83 & 83.35 & 85.97 & 1.64 & 1.73 \\
\rowcolor{cdTealLight}\szC\,4B & 5,446 & 100.0 & 0.458 & 75.71 & 91.20 & 77.28 & 83.53 & 84.00 & 85.99 & 1.13 & 1.26 \\
\arrayrulecolor{cdInk}\bottomrule
\end{tabular}
\par\smallskip\begin{minipage}{\linewidth}\tablenote{Unmerged adapters at $T=1$; accuracy and ECE in \%. Progress is the share of the 13,048,000 training decisions consumed. Highlighted rows are the validation-selected checkpoints that were merged and released.}\end{minipage}
}

\end{table}
\begin{table}[!ht]
\caption{\textbf{Accuracy with 95\% parent-clustered bootstrap intervals} and paired differences between adjacent decision-only sizes (percentage points); the values behind Figure~\ref{fig:forest}.}
\label{tab:bootstrap}
\centering
{\footnotesize
\setlength{\tabcolsep}{4.5pt}
\begin{tabular}{lrrrrr}
\arrayrulecolor{cdInk}\toprule
\rowcolor{cdMist}Group & \szA\,0.8B & \szB\,2B & \szC\,4B & 2B $-$ 0.8B & 4B $-$ 2B \\
\arrayrulecolor{cdLine}\midrule
General & 72.26 {\color{cdGray}$\pm$1.17} & 73.96 {\color{cdGray}$\pm$1.06} & 75.69 {\color{cdGray}$\pm$1.15} & \sigmark+1.70 {\color{cdGray}[+1.45, +1.95]} & \sigmark+1.73 {\color{cdGray}[+1.50, +1.97]} \\
Medical & 88.85 {\color{cdGray}$\pm$0.20} & 89.50 {\color{cdGray}$\pm$0.19} & 91.19 {\color{cdGray}$\pm$0.17} & \sigmark+0.65 {\color{cdGray}[+0.54, +0.75]} & \sigmark+1.69 {\color{cdGray}[+1.59, +1.80]} \\
Legal & 67.84 {\color{cdGray}$\pm$1.96} & 68.74 {\color{cdGray}$\pm$1.61} & 77.21 {\color{cdGray}$\pm$1.64} & \nsmark+0.91 {\color{cdGray}[-0.73, +2.43]} & \sigmark+8.47 {\color{cdGray}[+6.98, +10.10]} \\
Finance & 74.00 {\color{cdGray}$\pm$1.48} & 76.50 {\color{cdGray}$\pm$1.51} & 83.61 {\color{cdGray}$\pm$1.26} & \sigmark+2.50 {\color{cdGray}[+1.18, +3.89]} & \sigmark+7.11 {\color{cdGray}[+5.75, +8.41]} \\
Prof.\ mean & 76.90 {\color{cdGray}$\pm$0.82} & 78.25 {\color{cdGray}$\pm$0.74} & 84.00 {\color{cdGray}$\pm$0.70} & \sigmark+1.35 {\color{cdGray}[+0.64, +2.00]} & \sigmark+5.76 {\color{cdGray}[+5.10, +6.49]} \\
Macro & 75.74 {\color{cdGray}$\pm$0.67} & 77.18 {\color{cdGray}$\pm$0.61} & 81.92 {\color{cdGray}$\pm$0.60} & \sigmark+1.44 {\color{cdGray}[+0.89, +1.93]} & \sigmark+4.75 {\color{cdGray}[+4.26, +5.31]} \\
Micro & 83.11 {\color{cdGray}$\pm$0.61} & 84.11 {\color{cdGray}$\pm$0.56} & 85.97 {\color{cdGray}$\pm$0.58} & \sigmark+1.01 {\color{cdGray}[+0.89, +1.12]} & \sigmark+1.86 {\color{cdGray}[+1.76, +1.97]} \\
\arrayrulecolor{cdInk}\bottomrule
\end{tabular}
\par\smallskip\begin{minipage}{\linewidth}\tablenote{2,000 parent-clustered replicates, stratified by domain; the same draws are applied to every model. $\pm$ is the half-width of the 95\% percentile interval. \sigmark{}: interval excludes zero; \nsmark{}: it does not. Parent units: 56,064 general, 76,270 medical, 2,048 legal, 3,274 finance.}\end{minipage}
}

\end{table}
\begin{table}[!ht]
\caption{\textbf{Unified against decision-only models}, paired differences with 95\% parent-clustered intervals; the values behind Figure~\ref{fig:forest-unified}.}
\label{tab:unified-paired}
\centering
{\footnotesize
\setlength{\tabcolsep}{2.7pt}
\begin{tabular}{lrlrlrl}
\arrayrulecolor{cdInk}\toprule
\rowcolor{cdMist}& \multicolumn{2}{c}{\uniA\,0.8B unified} & \multicolumn{2}{c}{\uniB\,2B unified} & \multicolumn{2}{c}{\uniC\,4B unified} \\
\rowcolor{cdMist}Group & ACC & $\Delta$ vs decision-only [95\% CI] & ACC & $\Delta$ vs decision-only [95\% CI] & ACC & $\Delta$ vs decision-only [95\% CI] \\
\arrayrulecolor{cdLine}\midrule
General & 73.86 & \sigmark+1.62 {\color{cdGray}[+1.41, +1.84]} & 75.43 & \sigmark+1.48 {\color{cdGray}[+1.28, +1.69]} & 77.30 & \sigmark+1.59 {\color{cdGray}[+1.40, +1.80]} \\
Medical & 88.61 & \sigmark-0.25 {\color{cdGray}[-0.34, -0.16]} & 90.03 & \sigmark+0.52 {\color{cdGray}[+0.43, +0.62]} & 91.64 & \sigmark+0.43 {\color{cdGray}[+0.35, +0.51]} \\
Legal & 67.88 & \nsmark+0.02 {\color{cdGray}[-1.95, +1.94]} & 73.42 & \sigmark+4.37 {\color{cdGray}[+3.04, +5.88]} & 82.44 & \sigmark+5.16 {\color{cdGray}[+3.81, +6.58]} \\
Finance & 74.31 & \nsmark+0.08 {\color{cdGray}[-1.10, +1.28]} & 77.61 & \sigmark+1.58 {\color{cdGray}[+0.36, +2.83]} & 84.00 & \nsmark+0.47 {\color{cdGray}[-0.52, +1.45]} \\
Prof.\ mean & 76.93 & \nsmark-0.05 {\color{cdGray}[-0.85, +0.72]} & 80.35 & \sigmark+2.16 {\color{cdGray}[+1.55, +2.80]} & 86.03 & \sigmark+2.02 {\color{cdGray}[+1.45, +2.61]} \\
Macro & 76.17 & \nsmark+0.37 {\color{cdGray}[-0.24, +0.96]} & 79.12 & \sigmark+1.99 {\color{cdGray}[+1.54, +2.47]} & 83.85 & \sigmark+1.92 {\color{cdGray}[+1.48, +2.36]} \\
Micro & 83.46 & \sigmark+0.35 {\color{cdGray}[+0.24, +0.46]} & 85.00 & \sigmark+0.89 {\color{cdGray}[+0.81, +0.99]} & 86.86 & \sigmark+0.87 {\color{cdGray}[+0.78, +0.95]} \\
\arrayrulecolor{cdInk}\bottomrule
\end{tabular}
\par\smallskip\begin{minipage}{\linewidth}\tablenote{Accuracy (\%) at $T=1$ of the final unmerged checkpoints and its paired difference to the decision-only model of the same size; 2,000 parent-clustered paired bootstrap replicates (same draws for every model). \sigmark{}: interval excludes zero; \nsmark{}: it does not. The two members of a pair differ in more than the extraction task (Section~\ref{sec:unified-decision}).}\end{minipage}
}

\end{table}
\begin{table}[!ht]
\caption{\textbf{Selective prediction of the unified models.} Compare Table~\ref{tab:selective} for the decision-only models.}
\label{tab:unified-selective}
\centering
{\footnotesize
\setlength{\tabcolsep}{3.4pt}
\begin{tabular}{lrrrrrrrrrr}
\arrayrulecolor{cdInk}\toprule
\rowcolor{cdMist}& \multicolumn{4}{c}{Accuracy (\%) at coverage} & \multicolumn{2}{c}{All (\%)} & \multicolumn{4}{c}{AURC by domain (\%)} \\
\rowcolor{cdMist}Model & 50\% & 80\% & 90\% & 100\% & AURC & E-AURC & General & Medical & Legal & Finance \\
\arrayrulecolor{cdLine}\midrule
\uniA\,0.8B & 99.47 & 92.92 & 88.55 & 83.46 & 3.22 & 1.77 & 8.40 & 1.57 & 13.97 & 9.57 \\
\uniB\,2B & 99.57 & 94.07 & 89.94 & 85.00 & 2.77 & 1.58 & 7.63 & 1.28 & 9.72 & 7.05 \\
\uniC\,4B & 99.66 & 95.42 & 91.59 & 86.86 & 2.26 & 1.35 & 6.74 & 0.99 & 4.57 & 4.20 \\
\arrayrulecolor{cdInk}\bottomrule
\end{tabular}
\par\smallskip\begin{minipage}{\linewidth}\tablenote{Hard decisions ranked by maximum candidate probability at the fitted temperature.}\end{minipage}
}

\end{table}
\begin{table}[!ht]
\caption{\textbf{Batch-one latency and host memory} of the released decision models under one protocol; the values behind Figure~\ref{fig:cost}.}
\label{tab:efficiency}
\centering
{\small
\setlength{\tabcolsep}{4.5pt}
\begin{tabular}{llrrrr}
\arrayrulecolor{cdInk}\toprule
\rowcolor{cdMist}Size & Backend & p50 short (ms) & p50 all (ms) & p95 all (ms) & RSS (GiB) \\
\arrayrulecolor{cdLine}\midrule
\szA\,0.8B & GPU \textperiodcentered{} Transformers BF16 & 44.3 & 44.4 & 55 & 2.57 \\
\szA\,0.8B & GPU \textperiodcentered{} llama.cpp Q8\_0 & 15.6 & 17.9 & 193 & 1.90 \\
\szA\,0.8B & CPU \textperiodcentered{} Transformers FP32 & 319.8 & 371.3 & 7,416 & 7.44 \\
\szA\,0.8B & CPU \textperiodcentered{} llama.cpp Q8\_0 & 347.2 & 427.0 & 7,047 & 2.21 \\
\arrayrulecolor{cdLine}\midrule
\szB\,2B & GPU \textperiodcentered{} Transformers BF16 & 44.7 & 44.8 & 90 & 4.67 \\
\szB\,2B & GPU \textperiodcentered{} llama.cpp Q8\_0 & 20.4 & 23.6 & 275 & 3.02 \\
\szB\,2B & CPU \textperiodcentered{} Transformers FP32 & 545.3 & 601.3 & 11,321 & 11.74 \\
\szB\,2B & CPU \textperiodcentered{} llama.cpp Q8\_0 & 664.5 & 818.0 & 12,885 & 3.34 \\
\arrayrulecolor{cdLine}\midrule
\szC\,4B & GPU \textperiodcentered{} Transformers BF16 & 59.9 & 60.1 & 199 & 9.01 \\
\szC\,4B & GPU \textperiodcentered{} llama.cpp Q8\_0 & 39.7 & 47.3 & 569 & 5.32 \\
\szC\,4B & CPU \textperiodcentered{} llama.cpp Q8\_0 & 1,695.1 & 2,004.0 & 31,481 & 6.14 \\
\arrayrulecolor{cdInk}\bottomrule
\end{tabular}
\par\smallskip\begin{minipage}{\linewidth}\tablenote{25 CJ-Bench inputs (92--6,521 tokens; 16 short inputs of 92--255 tokens), three warm repeats each, batch size one, on a shared server (A800 80GB PCIe, Xeon Gold 5318Y, 16 CPU threads). Pretokenized forward pass and candidate probabilities only. RSS is host memory. 4B FP32 on the CPU was not measured.}\end{minipage}
}

\end{table}
Table~\ref{tab:domain-type} reports domain-by-type accuracy of the released models. All 3,078 soft benchmark targets occur in general-domain \score{} decisions; they contribute to distribution NLL, Brier, and expected-score MAE, but not accuracy or ECE. Table~\ref{tab:trajectory} lists every full-benchmark checkpoint evaluation. Exact per-group counts, per-source metrics, and bootstrap summaries are included in the accompanying machine-readable evidence.

The benchmark contains 56,064 parent units in general, 76,270 in medicine, 2,048 in law, and 3,274 in finance. These counts are much smaller than decision counts in the general domain (1.78 decisions per parent) and medicine (2.62), which is why clustered intervals are wider than naive ones there. One legal source, ELAM, places all 400 of its decisions under a single parent, so it contributes one cluster to the bootstrap.

\section{Reproduction Details}
\label{app:reproduce}
All training runs use random seed 42, LoRA rank 32, scaling 64, and dropout 0.05. The target projection names are:
\begin{lstlisting}
q_proj, k_proj, v_proj, o_proj,
in_proj_qkv, in_proj_z, in_proj_a, in_proj_b,
out_proj, gate_proj, up_proj, down_proj
\end{lstlisting}
The training implementation uses AdamW defaults except for the recorded learning rate, BF16 parameters/computation as configured by the backbone, and non-reentrant gradient checkpointing. The learning-rate and RLCD noise schedules are functions of the fraction of decisions consumed, so runs with different update counts follow the same schedule over data. The seed and data order are recorded, but they do not guarantee bitwise reproducibility across GPU counts, libraries, and kernels. The 0.8B decision-only run changed its distributed world size once during training; all other runs used four GPUs throughout.

The nominal 0.8B decision-only command sequence is:
\begin{lstlisting}
python scripts/prepare_data.py
  --data datasets/CJ-Dataset
  --out datasets/CJ-mixed

torchrun --standalone --nproc_per_node=4
  scripts/train.py --out runs/cj-ddp

python scripts/finalize.py
  --run runs/cj-ddp
  --out models/chinese-decision-0.8b
\end{lstlisting}
Line breaks above are typographical; each command is executed as a single shell command. The 2B and 4B runs change the backbone, output directory, minimum batch, token budget, and objective:
\begin{lstlisting}
# 2B
--base models/Qwen3.5-2B --out runs/cj-2b-rlcd-bs2048
--global-batch-size 2048 --token-budget 524288 --rlcd
# 4B
--base models/Qwen3.5-4B --out runs/cj-4b-rlcd-bs2048
--global-batch-size 2048 --token-budget 262144 --rlcd
\end{lstlisting}
All other shared defaults are as in Table~\ref{tab:training}. Changing the maximum-step limit would change the learning-rate and noise schedules, so a short run with a compressed schedule is not a matched early checkpoint of the one-epoch run.

The unified 0.8B model was trained with the following command; the 2B model changes the base, the output directory, and \texttt{--token-budget} (12,288); the 4B model changes the base and the output directory and adds \texttt{--grad-checkpointing} (the token budget stays 24,576); and four GPUs need \texttt{NCCL\_P2P\_DISABLE=1} on our server:
\begin{lstlisting}
python -m torch.distributed.run --standalone
  --nproc_per_node=4 scripts/train_multitask_ddp.py
  --base models/Qwen3.5-0.8B --data datasets/cj-ie
  --out runs/cj-ie-0.8b-from-base
  --cj-train datasets/cj-full/train.cases.jsonl
  --kl-weight 0 --selection joint
  --cj-dev datasets/cj-full/dev.cases.jsonl
  --cj-calibration datasets/cj-full/calibration.cases.jsonl
  --epochs 1 --lr 1e-4 --head-lr 5e-4 --lora-r 32
  --decision-batch 32 --extraction-batch 16
  --token-budget 24576 --accumulate 8 --warmup 100
  --eval-every 1000
\end{lstlisting}
The replay-and-KL run starts from \texttt{models/chinese-decision-0.8b} with \texttt{--replay datasets/cj-replay/train.cases.jsonl --kl-weight 1.0 --max-retention-drop 1.0 --eval-every 500} and a 348,000-decision replay sample; the CJ-IE-only run omits all CJ data, uses cross-entropy only (RLCD off), \texttt{--warmup 500}, and evaluates every 3,000 updates. CJ-IE is built by \texttt{scripts/build\_ie\_data.py} (about four minutes), and the replay sample and the dev and calibration partitions of CJ-Dataset by \texttt{scripts/build\_replay\_data.py}.

For benchmarking, the evaluator snapshots the validation-selected adapter and records dataset checksums, adapter SHA256, compiler version, metric-code SHA256, temperature, candidate-order policy, coverage, and runtime configuration. It never fits temperature on CJ-Bench. The released data revision used in all runs is
\begin{lstlisting}
a3963a61f71e00051a80721e587d720d8926e8e7
\end{lstlisting}
Checkpoint runs use a maximum evaluation batch of 256 decisions and a 65,536 padded-token budget. Release and Q8\_0 evaluations are separate runs with their own batch and backend settings, so their throughput should not be read as a pure temperature, merge, or quantization effect.

The single-input forward helper avoids padding for a lone sequence because the backbone's linear-attention padding behavior differs between singleton and multirow inputs. Multirow inputs use left padding to a multiple of 64 with attention masks. This detail is relevant when comparing numerical results across serving backends and batch configurations.

The analyses of Section~\ref{sec:analysis-protocol} are reproduced from the stored prediction files with
\begin{lstlisting}
python analysis/analyze_predictions.py --bootstrap 2000
python analysis/analyze_unified.py --bootstrap 2000
python analysis/analyze_spans.py --bootstrap 2000
python generate_artifacts.py
\end{lstlisting}
Each script checks every recomputed metric against the stored run summary before writing its JSON file; the bootstrap seeds are fixed (20261021 and 20261027). The span utilities come from \texttt{analysis/dump\_spans.py}, run once per model and split on a GPU. The latency table of Section~\ref{sec:unified-serving} comes from \texttt{analysis/benchmark\_unified.py} and the one-pass check from \texttt{analysis/check\_combined.py}, both run with the current \texttt{decision} package on one idle GPU, and \texttt{analysis/stats\_unified\_text.py} prints the statistics quoted in Section~\ref{sec:unified-results}. The worked examples of Figures~\ref{fig:pipeline} and~\ref{fig:tournament} come from \texttt{analysis/run\_examples.py}, which sends \texttt{examples/request.json} through the three released decision-only Q8\_0 models and \texttt{examples/tournament.json} through the 4B Q8\_0 model on a CPU (llama.cpp, eight threads). The examples of Figures~\ref{fig:teaser} and~\ref{fig:gallery} come from \texttt{analysis/run\_examples\_unified.py}, which runs the three released unified models on a GPU with the repository's own compiler and pipeline and their release defaults.

\section{Artifact and Extensions}
\label{app:artifact}
The Overleaf project contains ordinary LaTeX source, a BibTeX bibliography, vector PDF figures, generated table fragments, and a compact evidence snapshot. It requires no shell escape, custom class, or Python execution on Overleaf; fonts (XCharter, Source Sans Pro, Source Code Pro, Font Awesome) come from standard \TeX{} Live packages. Figure text uses Noto Sans SC, which shares its Latin design with Source Sans and covers the Chinese examples. The architecture figure is the author's raster figure, cropped. The optional local scripts rebuild figures and tables from the included evidence; their outputs are already packaged.

Future family members should use the same full benchmark, report consumed decisions and training objective, distinguish raw from temperature-scaled and merged results, and report clustered intervals. A controlled scale comparison should hold data exposure, objective, token budget, adapter settings, selection protocol, and evaluation implementation fixed, while reporting compute and memory. A controlled RLCD study should compare objectives within the same backbone and training budget. For the unified models, single-factor ablations of the recipe (starting point, RLCD, KL anchor, and the share of span batches) and a longer schedule would show how much of the retention cost is avoidable.

\section{Revision History}
\label{app:changes}
\paragraph{Version 6 (November 1, 2026).} A correction release with no change to any benchmark number. Section~\ref{sec:rlcd} is retitled and now states that every released model except the first-trained decision-only 0.8B model uses RLCD, including the unified 0.8B model; the comparisons and limitations that depend on the objective name the decision-only or unified series explicitly. The training times of the unified runs are taken from their training logs (17.0, 26.1, and 82.6 hours) instead of the model card. The abstract, the conclusion, and the paragraph on uncontrolled comparisons are shortened.

\paragraph{Version 5 (November 1, 2026).} Adds the unified 4B model, trained from the raw Qwen3.5-4B with the recipe of the other unified models (about 83 hours on four A800 GPUs), and everything that changes with it: unified 4B rows in the comparison tables and figures; Section~\ref{sec:unified-results} for three sizes, with a paired comparison of the unified and decision-only 4B models ($+0.87$ points, which, unlike the 0.8B gain, does not come from one source), extraction at three sizes, and a retention cost that falls from 10.7 to 7.9 to 5.0 F1 points against the 0.8B extraction-only reference; and a new first-page summary in which the unified 4B model is the most accurate released model. Two changes to the released models since version 4 are described: the release default for span decoding (a decision shift of 0.75 since October 31, 2026, with probabilities still at $\log 5$; Section~\ref{sec:span}) and requests with both readouts answered in one pass (October 28, 2026; Section~\ref{sec:joint}). Extraction latency is re-measured with the released unified models and the current implementation (Table~\ref{tab:serving}), replacing the version-4 numbers from an earlier run. No decision-only number changes. New evidence: stored CJ-Bench predictions and the training history of the unified 4B run, inference-only span utilities of the unified 4B model on the CJ-IE test and dev splits, and timing and equivalence checks of the three unified models.

\paragraph{Version 4 (October 27, 2026).} Adds the unified 0.8B and 2B models, which carry a typed span readout in the same weight file, and everything needed to evaluate them: the span head and echo layout (Section~\ref{sec:span}), joint training and export (Section~\ref{sec:joint}), the CJ-IE data (Section~\ref{sec:cjie}, Appendix~\ref{app:cjie}), and Section~\ref{sec:unified-results}, which studies decision retention, extraction quality, the cost of retention against an extraction-only continuation, decoding choices, and errors. The title and abstract change accordingly, the main table gains the unified rows, Figure~\ref{fig:teaser} is redrawn with real outputs of both readouts, and the architecture figure is the author's. No decision-only number changes. New evidence: stored CJ-Bench predictions and training histories of four unified-run evaluations, and inference-only span utilities of four runs on the CJ-IE test and dev splits. A unified 4B model is training and is not reported.

\paragraph{Version 3 (October 22, 2026).} A presentation release with no change to any benchmark number. The report moves to a single-column layout with a new type and color system; the main domain results are merged into one table (Table~\ref{tab:main}); every figure is redrawn at the text width; and new figures show the released models at work, the paired size effects (Figure~\ref{fig:forest}), and the cost of each backend (Figure~\ref{fig:cost}). The worked examples are real outputs of the released Q8\_0 models for the repository's example requests.

\paragraph{Version 2 (October 21, 2026).} Version 1 (October 16, 2026) reported one completed 0.8B run and two early checkpoints (20\% and 40\% of the epoch) of the 2B run. Version 2 made the following changes.
\begin{itemize}
\item The 2B run is complete (step 4,994), and a 4B member (step 5,446) is added; all three sizes are released with fitted temperatures and Q8\_0 conversions.
\item Main comparison tables now report the released merged models with their fitted temperatures. Version 1 reported the unmerged 0.8B adapter at $T=1$ (for example, 72.24\% general and 76.98\% professional accuracy); the released 0.8B model scores 72.26\% and 76.90\%. Accuracy differences between the two artifacts come from merging, not temperature. Checkpoint trajectories remain at $T=1$.
\item New analyses from stored predictions: parent-clustered paired bootstrap intervals, a temperature-only ablation on fixed logits, selective prediction, paired transitions between sizes, merge-induced prediction changes, and source-level tests.
\item Quantization results cover all three sizes; latency is reported under one shared protocol across sizes and backends instead of mixing workloads.
\item The candidate tournament for \choice{} questions with more candidates than one prompt holds is described, and the interface limit of 26 candidates is distinguished from the 16 candidates seen in training.
\item Version 1 stated that the 2B run had not surpassed 0.8B on the professional average. That statement described a 40\%-epoch checkpoint and no longer holds for the completed run.
\end{itemize}

\end{document}